\documentclass[final,12pt]{colt2023} 

\usepackage[utf8]{inputenc} 
\usepackage[T1]{fontenc}    
\usepackage{hyperref}       
\usepackage{url}            
\usepackage{booktabs}       
\usepackage{amsfonts}       
\usepackage{nicefrac}       
\usepackage{microtype}      
\usepackage{xcolor}         
\usepackage{makecell}       

\usepackage{times, amsmath, amssymb, graphicx, url}
\usepackage{thmtools, thm-restate}

\usepackage{subcaption, wrapfig}

\usepackage{xcolor}

\usepackage{fancyhdr}
\usepackage{appendix} 

\usepackage{graphicx}

\usepackage{algorithm,algorithmic} 

\usepackage{amsmath,amsfonts,amssymb}
\usepackage{math_letters} 

\newtheorem{theorem}{Theorem}

\newtheorem{lemma}[theorem]{Lemma}

\newtheorem{corollary}[theorem]{Corollary}

\newtheorem{definition}[theorem]{Definition}
\newtheorem{assumption}[theorem]{Assumption}

\newtheorem{remark}[theorem]{Remark}

\newenvironment{mydefinition}[1]{
	
	\begin{minipage}{0.91\textwidth}
		\vspace{2mm}	
		\begin{definition}[#1]
		}{
		\end{definition}
		\vspace{0mm}
	\end{minipage}
	
}

\newenvironment{myassumption}[1]{
	
	\begin{minipage}{0.91\textwidth}
		\vspace{2mm}	
		\begin{assumption}[#1]
		}{
		\end{assumption}
		\vspace{0mm}
	\end{minipage}
	
}

\newenvironment{mylemma}[1]{
	
	\begin{minipage}{0.91\textwidth}
		\vspace{2mm}	
		\begin{lemma}[#1]
		}{
		\end{lemma}
		\vspace{0mm}
	\end{minipage}
	
}

\newenvironment{mytheorem}[1]{
	
	\begin{minipage}{0.91\textwidth}
		\vspace{2mm}	
		\begin{theorem}[#1]
		}{
		\end{theorem}
		\vspace{0mm}
	\end{minipage}
	
}

\newenvironment{mycorollary}[1]{
	
	\begin{minipage}{0.91\textwidth}
		\vspace{2mm}	
		\begin{corollary}[#1]
		}{
		\end{corollary}
		\vspace{0mm}
	\end{minipage}
	
}

\newenvironment{myproof}[1]{
	
		\noindent
		\hrulefill
		
		\noindent{\bf Proof #1: }				
		
		\vspace{-2mm}
		\noindent
		\hrulefill
		
	}
	{$\hfill\square$
		
		\vspace{-2mm}
		\noindent
		\hrulefill
	\bigskip
}

\renewcommand{\ln}{\log}

\graphicspath{{figsLaplace/}}

\title[Bregman Deviations of Generic Exponential Families]{Bregman Deviations of Generic Exponential Families}
\usepackage{times}

\coltauthor{
 \Name{Sayak Ray Chowdhury} \Email{t-sayakr@microsoft.com}\\
 \addr Microsoft Research, India
 \AND
 \Name{Patrick Saux} \Email{patrick.saux@inria.fr}\\
 \addr Univ. Lille, Inria, CNRS, UMR 9189 - CRIStAL, F-59000, Lille, France
 \AND
 \Name{Odalric-Ambrym Maillard} \Email{odalric.maillard@inria.fr}\\
 \addr Univ. Lille, Inria, CNRS, UMR 9189 - CRIStAL, F-59000, Lille, France
 \AND
 \Name{Aditya Gopalan} \Email{aditya@iisc.ac.in}\\
 \addr Indian Institute of Science, Bangalore, India
}

\begin{document}

\maketitle

\begin{abstract}
We revisit the method of mixtures, or Laplace method, to study the concentration phenomenon in generic (possibly multidimensional) exponential families. Using duality properties of the Bregman divergence to construct nonnegative martingales, we establish a generic inequality controlling the deviation between the parameter of the family and a finite sample estimate. This bound is expressed in the local geometry induced by the Bregman pseudo-metric. Moreover, it is time-uniform and involves a quantity extending the classical \textit{information gain} to exponential families, which we call the \textit{Bregman information gain}.
For the practitioner, we instantiate this novel bound to several classical families, e.g., Gaussian (including with unknown variance or multivariate), Bernoulli, Exponential, Weibull, Pareto, Poisson and Chi-square, yielding explicit forms of the confidence sets and the Bregman information gain. We further compare the resulting confidence bounds to state-of-the-art time-uniform alternatives and show this novel method yields competitive results. 
Finally, we apply our result to the design of generalized likelihood ratio tests for change detection, capturing new settings such as variance change in Gaussian families.
\end{abstract}


\section{Introduction}\label{sec:intro}

Concentration inequalities are a powerful set of methods in statistical theory with key applications in machine learning. Often in machine learning applications, a learner estimates some quantity solely based on samples from an unknown distribution and would like to know the magnitude of the estimation error. The typical example is that of the mean $\mu$ of some real-valued random variable $X$, estimated by its empirical mean built from a sample of $n$ independent and identically distributed (i.i.d.) observations. We refer the interested reader to the monographs of \cite{boucheron2013concentration, raginsky2012concentration, zeitouni1998large} for standard results and related topics.

In many situations, one may want to further estimate some vector parameter, as in, e.g., linear bandits \citep{abbasi2011improved} and logistic bandits \citep{faury2020improved}. A closely related problem is to estimate the parameter of a distribution coming from a parametric family
\citep{chowdhury2021reinforcement}.
\emph{Exponential families} are a flexible way to formalize such distributions over a set $\cX$ by describing densities of the form $p_\theta(x)\propto h(x)\exp(\langle \theta,F(x)\rangle)$, for some given feature function $F:\cX\mapsto\Real^d$ and base function $h$ (Section~\ref{sec:bregman}). Most classical distributions fall into this category, e.g., Gaussian (possibly multivariate, with or without known variance), Exponential, Gamma, Chi-square, Weibull, Pareto, Poisson, Bernoulli and Multinomial distributions \citep{amari2016information}.

A number of problems in current-day machine learning involve sequential, active data-sampling strategies \citep{cesa2006prediction}. This includes multi-armed bandits, reinforcement learning, active learning and federated learning, to mention a few application domains. Since the decision to sample a novel observation 
results from the interaction between the learning algorithm and the environment, 
and depends on past observations, 
one needs to design concentration inequalities working with a \emph{random} number of observations typically at a random stopping time \citep{durrett2019probability}. A natural way to handle this difficulty is to derive time-uniform concentration inequalities, producing sequences of confidence sets valid \textit{uniformly} over all number of observations with high probability, as opposed to being valid for a single number of observation.

A popular method in bandit theory is to combine supermartingale techniques with union bound arguments over a geometric time grid, a technique known as time peeling (or stitching) -- see \citet{bubeck2010bandits, cappe2013kullback} for early uses in bandits, as well as \citet{garivierICB13}, or more recently \citet{maillard19HDR}; see also \citet{howard2020time, howard2021time} for a recent, complementary survey of the history of this field, and \citet{pmlr-v139-kuchibhotla21a} for an extension of Bentkus' concentration bounds \citep{bentkus2004hoeffding} for bounded distributions using time peeling.
The method of mixtures, initiated by \citet{robbins1949application, robbins1970statistical} and popularized in \citet{pena2008self} is a powerful alternative to peeling for developing time-uniform confidence sets. 
It has been applied to sub-Gaussian families in \cite{abbasi2011improved}, leading to a variety of applications \citep{chowdhury2017kernelized,durand2017streaming,kirschner2018information}. In \citet{kaufmann2021mixture}, a generalization to handle one-dimensional exponential families is considered, with applications to Gaussian (known variance) and Gamma distributions (known shape). A fairly different `capital process' construction technique has been recently developed for bounded distributions in \citet{shafer2019game}, and popularized further in \citet{waudby2020estimating}.

In this work, we revisit the method of mixtures for parametric exponential families of arbitrary dimension, expressing deviations in the natural (Bregman) divergence of the family.
The setting of exponential families is convenient for integration in a Bayesian setup, thanks to the notion of conjugate prior that enables us to reduce computation of tedious integrals to simple parameter updates. Here, we exploit this property to obtain explicit mixtures of martingales. Exponential families are largely used in modern machine learning, yet concentration tools available to the practitioner are comparatively scarce beyond the Gaussian case. To help close this gap, we obtain both sharp and computationally tractable confidence sets, especially in the small sample regime.

\paragraph{Outline and contributions.}
In Section~\ref{sec:bregman}, we first recall some background material on exponential families and their associated Bregman divergences.
Section~\ref{sub:main-result} states our main result (Theorem~\ref{thm:genericmom}): a \emph{time-uniform} concentration inequality for exponential families. Specifically, we control the Bregman deviations associated with the log-partition function of the family using a novel information-theoretic quantity, the \emph{Bregman information gain}. On a high level, this quantifies the information gain about the parameter of the family after observing i.i.d. samples from it, which is measured in terms of the natural Bregman divergence of the family. To illustrate the utility of this general result, we detail in Section~\ref{sub:classicalfamilies} how Bregman information gain and deviation inequalities specialize for well-known exponential families, resulting in fully explicit confidence sets (see Table~\ref{table:bregman_cs_partial}). To the best of our knowledge, we are the first to derive an explicit \emph{time-uniform} deviation inequality for two-parameter Gaussian (i.e. both mean and variance are unknown), Chi-square, Weibull, Pareto, and Poisson distributions. Our result is an adaptation of the \textit{method of mixtures} technique, and a proof sketch is outlined in Section~\ref{sub:proof}. In Section~\ref{sec:xps}, we numerically evaluate the high-probability confidence sets built from our method for classical families, and achieve state-of-the-art time-uniform bounds. Finally, in Section~\ref{sec:GLR}, we generalize Theorem~\ref{thm:genericmom} to obtain a \emph{doubly time-uniform} concentration inequality for generic exponential families, which could be of independent interest (Theorem~\ref{thm:Doubly-uniform}). We present an application of both results in controlling the false alarm probability of the Generalized Likelihood Ratio (GLR) test, used for change detection in the exponential family model.
\section{Exponential Families and Bregman Divergence}\label{sec:bregman}
In this section, we introduce exponential families and the link between their Kullback-Leibler (KL) and Bregman divergences, as well as useful properties of these divergences.

\paragraph{Exponential families.}

We consider an exponential family of distributions $\{p_\theta\}_{\theta\in\Theta}$ over some set $\cX$,
parameterized in some open set $\Theta\subset\Real^d$, whose density or mass function has the form $p_\theta(x)\!=\!  h(x)\exp( \langle \theta, F(x)\rangle \!-\! \cL(\theta))$. Here,  $F\!:\!\cX\!\to\!\Real^d$ is the feature function,  $h\!:\!\cX\!\to\!\Real_+$ is the base function,
and $\cL$ represents the normalization term (a.k.a. log-partition function, convex w.r.t. $\theta$) given by $\cL(\theta)=\log \int_\cX h(x)\exp( \langle \theta, F(x)\rangle)dx$. 
We denote by  $\Theta_\cD=\big\{ \theta\in\Real^d: \cL(\theta)< \infty\big\}$ the domain of $\cL$ and by $\Theta_I=\big\{ \theta\in\Theta_\cD: \det \nabla^2 \cL (\theta) > 0 \big\}$ the set on which its Hessian is invertible. We assume that $\Theta\subset \Theta_I$, which is tantamount to assuming that the family is minimal, and ensures we only consider non-degenerate distributions ($\nabla \cL$ is one-to-one on its domain). We use notations $\Pr_\theta, \Esp_\theta$ to explicitly refer to the probability and expectation associated to the distribution $p_\theta$.

\paragraph{Bregman divergence of an exponential family.}

A fundamental property of exponential families is the following form for the KL divergence between two distributions with parameters $\theta,\theta'\in\Theta_\cD$:
\beqan
\KL(p_{\theta},p_{\theta'}) = \langle\theta-\theta',\Esp_{\theta}(F(X))\rangle - \cL(\theta) + \cL(\theta')\,.
\eeqan
Here, $\Esp_{\theta}(F(X))$ is called the vector of \textit{expectation parameters} (a.k.a. \textit{dual parameters}), and is equal to
$\nabla \cL(\theta)$. Hence, it holds that $\KL(p_{\theta},p_{\theta'}) = \cB_{\cL}(\theta',\theta)$, where $\cB_{\cL}$ is known as the Bregman divergence \citep{bregman1967relaxation}
with potential function $\cL$, defined by 
\beqan
\cB_{\cL}(\theta',\theta)  \eqdef  \cL(\theta') - \cL(\theta) -\langle\theta'-\theta,\nabla \cL(\theta)\rangle  \,.
\eeqan

\paragraph{Tail and duality properties.}

The canonical Bregman divergence of an exponential family enjoys two fundamental properties. 
The first one links it to the log-moment generating function of the random variable $F(X)$, which makes Bregman divergences well suited to control the tail behavior of random variables appearing in concentration inequalities. The second one highlights duality properties that enables convenient algebraic manipulations. To this end, for any $\lambda \in \Real^d$, we define the function $\cB_{{\cL},\theta}(\lambda)\!=\!\cB_{\cL}(\theta\!+\!\lambda,\theta)\!=\! {\cL}(\theta\!+\!\lambda) \!-\! {\cL}(\theta) \!-\!\langle\lambda,\nabla {\cL}(\theta)\rangle $. Also, we introduce the Legendre-Fenchel dual operator $\star$ associating a function $G$ to its dual $G^\star(x) = \sup_\lambda \langle \lambda,x\rangle - G(\lambda)$. 
\begin{restatable}[Properties of Bregman divergences]{lemma}{lemone}\label{lem:bregdual}
	For all $\theta\!\in\!\Theta_\cD$ and $\lambda\!\in\!\Real^d$ such that $\theta\!+\!\lambda\in\Theta_\cD$,
	\beqan
	\log \Esp_{\theta}\left[ \exp\left( \langle \lambda, F(X)-\Esp_{\theta}[F(X)]\rangle \right)\right] = \cB_{\cL, \theta}(\lambda)\,.
	\eeqan
	Furthermore, if $\nabla\cL$ is one-to-one, the following Bregman duality relations hold for any $\theta,\theta'\in\Theta_\cD:$
	\beqan
	\cB_{\cL}(\theta',\theta)
	= \cB_{\cL,\theta'}^\star(\nabla{\cL}(\theta)-\nabla{\cL}(\theta'))= \cB_{\cL^\star}(\nabla{\cL}(\theta),\nabla{\cL}(\theta')).
	\eeqan
	More generally, the following holds for any $\alpha\in[0,1]:$ 
	\beqan
	\cB_{{\cL},\theta'}^\star\left( \alpha(\nabla{\cL}(\theta)-\nabla{\cL}(\theta'))\right) = \cB_\cL\big(\theta', \theta_\alpha\big),\,\,\text{where}\,\, \theta_\alpha=\nabla {\cL}^{-1}\big(\alpha \nabla {\cL}(\theta) + (1-\alpha) \nabla {L} (\theta')\big).
	\eeqan
\end{restatable}

The second half of this technical lemma is essentially a change of variable formula to move back and forth between two representations of an exponential family: in natural parameters (measured by the Bregman divergence between $\theta'$ and $\theta$) and in expectation parametrization (measured by the dual Bregman divergence between $\nabla \cL(\theta)=\bE_{\theta}[F(X)]$ and $\nabla \cL(\theta')=\bE_{\theta'}[F(X)]$ ). For more background on this, which forms the basis of the \textit{information geometry} field, we refer to \citet[Sections~2.1 and 2.7]{amari2016information}. This result is at the root of the martingale construction behind Theorem~\ref{thm:genericmom} in the next section. For completeness, the proof of this classical lemma is given in Appendix~\ref{app:technical}.
\section{Time-uniform Bregman Concentration}\label{sec:concentration}
In this section, we are interested in controlling the deviation 
between a parameter $\theta \in \Theta$ and its estimate $\theta_n$ built from $n$ observations from distribution $p_{\theta}$.  We naturally measure this deviation in terms of the canonical Bregman divergence
of the family. Further, we would like to control this deviation not only for a single sample number $n$, 
but {\em simultaneously for all $n\in\bN$}.
Namely, we would like to upper bound quantities of the form $\bP\left[\exists n\!\in\!\Nat : \cB_\cL(\theta,\theta_n) \geq \dots\right]$.
Such a control is very useful in contexts when observations are gathered sequentially (either actively or otherwise), and especially when the number of observations $n$ is unknown beforehand. Classical examples include multi-armed bandits \citep{Auer02}, or model-based reinforcement learning \citep{jaksch2010near}.

\subsection{A generic deviation inequality}\label{sub:main-result}
In this section, we consider the problem of controlling  the Bregman-deviations of a parameter estimate of $\theta$.
To this end, we adapt the method of mixtures (a.k.a. Laplace method) from \citep{pena2008self}.
The method is originally designed in the context of Gaussian distributions, where it yields simple closed-form expressions, even though it can be applied more generally.
We state below a generic extension of the method to parametric exponential families and introduce a quantity that measures a form of \emph{information gain} about $\theta$ after observing $n$ samples, but expressed in terms of the natural Bregman divergence.
For this reason, we call this quantity the {\em Bregman information gain}.
\begin{definition}[Bregman information gain] \label{def:info-gain}
Let $X_1,\ldots,X_n \!\sim\! p_{\theta}$ be i.i.d. samples generated from $p_{\theta}$, where $\theta\in\Theta\subset\Theta_I$, and let  $\theta_0\!\in\!\Theta$ be a reference parameter. 
For any constant $c\!>\!0$, the Bregman information gain about $\theta$ after observing $X_1,\ldots,X_n$ from $\theta_0$ is defined as
\begin{align*}
  \gamma_{n,c}(\theta_0)\!=\! \log\! \left(\!\frac{\int_{\Theta} \exp\big(\!-\!c\cB_\cL(\theta',\theta_0)\big)d\theta'}{\int_{\Theta}\!\exp\!\Big(\!-\!(n\!+\!c)\cB_\cL(\theta',\theta_{n,c}(\theta_0))\Big)d\theta'}\!\right)\!,  
\end{align*}
where $\theta_{n,c}(\theta_0)\!=\!(\nabla \cL)^{-1}\!\bigg(\!\frac{\sum_{t=1}^{n}F(X_t)+c\nabla \cL(\theta_0)}{n+c}\!\bigg)$ denotes a parameter estimate of $\theta$.\footnote{$\theta_{n,c}(\theta_0\!)$ is actually a maximum a posteriori estimate under a conjugate prior on $\theta$, depending on the reference point $\theta_0$.}
\end{definition}



\paragraph{Dependence on $\theta_0$ and example.}

The acute reader can note that the considered parameter estimate $\theta_{n,c}(\theta_0)$ and Bregman information gain $\gamma_{n,c}(\theta_0)$ involve a reference parameter $\theta_0$.
It makes sense to have such a local reference point since the Bregman divergence is typically linked to metrics with \emph{local} (non-constant) curvature. Hence, the (information) geometry seen from the perspective of different points $\theta_0$ may be different, unlike in the Gaussian case.
Specifically, for a Gaussian $\cN(\mu,\sigma^2)$ with known variance $\sigma^2$, the Bregman information gain w.r.t. a reference point $\mu_0$ reads
\begin{align*}
    \gamma_{n,c}^{\cN}(\mu_0)\!=\!\frac{1}{2}\!\log \frac{2\pi  \sigma^2}{c}\!-\!\frac{1}{2}\!\log \frac{2\pi  \sigma^2}{n\!+\!c}\! =\!\frac{1}{2}\! \log \frac{n\!+\!c}{c}\,.
\end{align*}
It is independent of the reference parameter $\mu_0$, which is a consequence of the fact that the Bregman divergence in this case is proportional to the squared Euclidean distance; in other words, Gaussian distributions with known variance exhibit invariant geometry. However, for other exponential families, the geometries are inherently different, and hence Bregman information gain
depend explicitly on local reference points (see Section~\ref{sub:classicalfamilies} for details).
For reference, the classical Gaussian information gain, i.e., the mutual information between a prior $\mu\sim\cN(\mu_0, \sigma^2)$ and the average of an i.i.d. sample $X_1, \dots, X_n$ drawn from $\cN(\mu, c\sigma^2)$ is $\frac{1}{2}\log\frac{n\!+\!c}{c}$, which matches the Bregman information gain.


We now present the main result of this paper -- a time uniform confidence bound for $\theta$ connecting the Bregman divergence geometry of the exponential family with its Bregman information gain.
\begin{restatable}[Main result: Laplace method for generic exponential families]{theorem}{thmone}
\label{thm:genericmom}
Fix any $\delta \in (0,1]$ and $n\!\in\!\Nat$.
Under the hypothesis of Definition~\ref{def:info-gain}, consider the confidence set 
\beqan
	\Theta_{n,c}(\delta) = \left\lbrace \theta_0\in\Theta:  (n+c)\cB_\cL \left(\theta_0, \theta_{n,c}(\theta_0)			\right)\leq  \log\frac{1}{\delta} + \gamma_{n,c}(\theta_0)\right\rbrace.
	\eeqan	
	The following time-uniform control holds whenever the Bregman information gain is well-defined:
	\beqan
\Pr_{\theta}\left[ \exists n\in\Nat : \theta \notin \Theta_{n,c}(\delta)\right] \leq  \delta\,.
	\eeqan
\end{restatable}
Note the implicit definition of the confidence set $\Theta_{n,c}(\delta)$. where the parameter of interest $\theta_0$ appears in both arguments of the Bregman divergence, as well as in the Bregman information gain $\gamma_{n,c}(\theta_0)$. Because of this, \textit{computing} this confidence set from the equation in Theorem~\ref{thm:genericmom} may seen non-trivial at first glance.
However, we show in Section~\ref{sub:classicalfamilies} how these sets simplify for many classical families, revealing how the computation can be made efficiently.
Moreover, we observe that these confidence sets are actually tighter than those of prior work, and as such are especially well suited to be used when $n$ is small (for large $n$, most methods produce essentially equivalent sets).
Note that our concentration bound holds uniformly over all $n$. Equivalently \citep[Lemma~3]{howard2020time}, it also holds that $\Pr_{\theta}\left[ \theta \notin \Theta_{\tau, c}(\delta)\right] \leq  \delta$ for any random stopping time $\tau$.

\paragraph{Comparison with prior work.}
Similar to this work, \citet{kaufmann2021mixture} extend the method of mixtures technique to derive time-uniform concentration bounds for exponential families. However, their proof technique is fairly different, relying instead on discrete mixtures and stitching, which only works for single parameter families, and involves case-specific calculations that are difficult to generalize beyond Gaussian (with known variance) and Gamma (with known shape). In contrast, our method applies to generic exponential families, including distributions with more than one parameter such as Gaussian when both mean and variance are unknown. Moreover, their discrete prior construction leads to technical constants seemingly unrelated to the exponential family model. Our prior is naturally induced by the exponential family, leveraging key properties of Bregman divergences, yields more intrinsic quantities (Bregman information gain) and perhaps a more elegant and shorter proof. Hence, our results are not only more general, but also of fundamental interest.

\paragraph{Asymptotic behavior.}
The asymptotic width of $\Theta_{n, c}(\delta)$ depends on the behavior of $\gamma_{n, c}(\theta_0)$ as $n\rightarrow +\infty$. Standard arguments show that $\theta_{n,c}(\theta_0)\rightarrow \theta$ and the Taylor expansion of $\theta'\mapsto \cB_{\cL}(\theta', \theta)$ around $\theta'=\theta$ is $\cB_{\cL}(\theta', \theta) = \frac{1}{2}(\theta'-\theta)^\top \nabla^2 \cL(\theta) (\theta' - \theta) + o\left( \lVert \theta' - \theta \rVert^2 \right)$ (note that $\nabla^2 \cL(\theta)$ is positive definite for $\theta\in\Theta_I$). Laplace's method for integrals (chapter 20 in \cite{LattimoreBanditAlgorithmsBook}, \cite{shun1995laplace}) then gives the following estimate:
\begin{align*}
\int_{\Theta}\!\exp\!\Big(\!-\!(n\!+\!c)\cB_\cL(\theta',\theta_{n,c}(\theta_0))\Big)d\theta'\approx \int_{\bR^d}\!\exp\!\Big(\!-\!\frac{n\!+\!c}{2}(\theta-\theta')^\top \nabla^2 \cL(\theta) (\theta-\theta')\Big)d\theta'\,,
\end{align*}
which is a simple Gaussian integral, and thus $\gamma_{n, c}(\theta_0) = \frac{d}{2}\log(1\!+\!n/c) + \cO(1)$. This asymptotic scaling is worse than the $\log\log n$ rate of the law of iterated logarithm.
However, this is a standard feature of the method of mixtures compared to stitching \citep{maillard19HDR, howard2020time}, which is compensated by its improved nonasymptotic sharpness, as evidenced in Section~\ref{sec:xps}.

\paragraph{Dependence on the parameter $c$.}
The parameter $c$ is usually chosen to  be $1$ in the Gaussian case. However, it is useful to study its influence over the bounds. We provide in Section~\ref{sec:xps} a detailed study of this parameter, revealing first that the confidence bounds are not significantly altered over a large range of values, and then explaining how to pick an optimized value of $c$, e.g., for a specific horizon $n=n_0$. Note that choosing a variable $c$ changing with $n$ is not allowed by the theory, as this would break the martingale property used in our proof (see Section~\ref{sub:proof}). Moreover, it would be incompatible with the time-uniform lower bound $\lvert \Theta_{n, c}\left(\delta\right)\rvert = \Omega(\sqrt{\log\log n / n})$ provided by the law of iterated logarithm. Indeed, the Bregman information gain with $c_n\!\propto\!n$ would be asymptotically $\gamma_{n, c_n}(\theta_0) = \frac{d}{2}\log(1\!+\!\cO(1))+\cO(1)=\cO(1)$, leading to $\lvert \Theta_{n, c}\left(\delta\right)\rvert = \cO(1/\sqrt{n})$ when $n\rightarrow +\infty$.

\paragraph{Application to bandits.}
One can apply the technique developed in proving Theorem~\ref{thm:genericmom} to build confidence sets in standard $K$-armed bandit problems. In such problems, we are given $K$ distributions $\lbrace p_{\theta_i}\rbrace_{i=1}^{K}$ from a generic exponential family (e.g., Gaussian) with parameters $\lbrace \theta_i \rbrace_{i=1}^{K}$, from which we can draw samples, interpreted as rewards we want to maximize. This setting is standard to analyze regret-optimal bandit algorithms \citep{cappe2013kullback, korda2013thompson, baudry2020sub}.
Specifically, at each time $n$, we choose an arm $i_n \in [K]$ based on past observations, and draw a sample $X_n$ from its distribution $p_{\theta_{i_n}}$. The samples are then used to update knowledge about the parameters $\lbrace \theta_{i}\rbrace_{i=1}^{K}$ by building confidence sets for each of them. Let $N_{i}(n)$ denote the number of times that we have chosen action $i$ up to time $n$. Also, let $\theta_{i,N_{i}(n),c}(\theta_0)$ and $\gamma_{i,N_{i}(n),c}(\theta_0)$ denote the parameter estimate and Bregman information gain for arm $i$, respectively (similar to Definition~\ref{def:info-gain} with $n$ replaced by $N_{i}(n)$ 
). We construct the confidence set for arm $i$ at time $n$ as
\beqan
	\Theta_{i,n,c}(\delta) = \bigg\{ \theta_0\in\Theta:  (N_i(n)+c)\cB_\cL \left(\theta_0,\!\theta_{i,N_i(n),c}(\theta_0)			\right)\leq  \log\frac{1}{\delta} + \gamma_{i,N_i(n),c}(\theta_0)\bigg\}.
	\eeqan
Then, similar to Theorem~\ref{thm:genericmom}, it holds that the true parameter $\theta_i$ lies in the set $\Theta_{i,n,c}(\delta)$ for all time-steps $n \in \Nat$ with probability at least $1-\delta$. Finally, we take a union bound over $i \in [K]$ to obtain confidence sets for all arms (with widths inflated by an additive $\log K$ factor). Such construction is standard and is used in \citet{abbasi2011improved} for (sub)-Gaussian families. Possible applications include UCB algorithms for regret minimization 
and designing GLR stopping rules for tracking algorithms in pure exploration \citep{garivier2016optimal} in the context of generic exponential families (see also Section~\ref{sec:GLR} for another application of GLR tests using the parameter estimate $\theta_{n,c}(\theta_0)$).



\begin{remark}
We provide in Appendix~\ref{sub:legendre} a complementary result (Corollary~\ref{cor:momEF}) using a Legendre function $\cL_0$ instead of the regularizing parameter $c$, which eschews the use of a local reference $\theta_0$. However, we argue that such a global regularization is actually less convenient to use except for univariate and multivariate Gaussian distributions that anyway exhibit invariant geometry. Furthermore, in Theorem~\ref{thm:momEFsequence}, we prove a more general result that handles the case of a sequence $(X_t)_{t\in\Nat}$ of random variables that are not independent, and having possibly different distributions from each other, which we apply to build confidence sets in linear bandits (see Appendix~\ref{app:bandits}).
\end{remark}



\subsection{Specification to classical families}\label{sub:classicalfamilies}

In this section, we specify the result of Theorem~\ref{thm:genericmom} to some classical exponential families. Interestingly, the literature on time-uniform concentration bounds outside of random variables that are bounded, gamma with fixed shape or Gaussian with known variance is significantly scarce, even though many more distributions are commonly used in machine learning models. We derive below explicit confidence sets for a range of distributions, which we believe will be of interest for the wider machine learning and statistics community. For instance, consider active learning in bandit problems \citep{carpentier2011upper}, where one targets upper confidence bounds on the variance (rather than the mean); in the Gaussian case, this can be achieved with Chi-square concentration. \citet{hao2019bootstrapping} studies the classical UCB algorithm for bandits under a weaker assumption that sub-Gaussianity, involving the Weibull concentration.
In differential privacy, concentration of Laplace distribution (symmetrized exponential) is often used to study the utility of differentially private mechanisms \citep{dwork2014algorithmic}. Finally, heavy-tailed distributions such as Pareto have recently been of interest to study risk-averse or corruption in bandit problems \citep{holland2021learning, basu2022bandits}.

We now make explicit the Bregman information gains and confidence sets for some illustrative families (more examples and full derivations are provided in Appendix~\ref{app:specific}).




     
\paragraph{Gaussian (unknown mean and variance).} Let $X \sim \cN(\mu,\sigma^2)$. Given samples $X_1,\ldots,X_n$, we define $S_n = \sum_{t=1}^{n}X_t$ and $\hat\mu_n=S_n/n$. Further, for $\mu,\sigma \in \Real \times \Real^{+}$, we define the normalized sum of squares $Z_n(\mu, \sigma)=\frac{1}{\sigma^2}\sum_{t=1}^{n}(X_t-\mu)^2$. Then, for reference parameters $\mu_0\in \Real,\sigma_0\in \Real^{+},c > 0$, the Bregman information gain reads
\begin{align*}
    \gamma_{n,c}^{\cN}(\mu_0, \sigma_0)=\!\frac{3}{2}\log\left(\frac{n}{n\!+\!c} Z_n(\hat \mu_n, \sigma_0) + \frac{c}{n\!+\!c}Z_n(\mu_0, \sigma_0) + c \right)\!+\! f_{n,c}\,,
\end{align*}  
where $f_{n,c}\!=\!\left(\frac{n+c+1}{2}\right)\log\!\left(n\!+\!c\right) \!-\!\left(\frac{c}{2}\!+\!2\right)\log c \!-\! \frac{n}{2}(1\!+\!\log 2) \!+\! \log \Gamma\!\left(\frac{c+3}{2}\right) \!-\! \log\Gamma\!\left(\frac{n+c+3}{2}\right)$. 



\paragraph{Bernoulli.} Let $X \sim \text{Bernoulli}(\mu)$, with unknown mean $\mu \in [0,1]$. Define, for reference parameter $\mu_0 \in \Real$, $c >0$, the estimate $\mu_{n,c}(\mu_0)\!=\!\frac{S_n+c\mu_0}{n+c}$. Then, the Bregman information gain is given by
\begin{align*}
    \gamma_{n,c}^{\text{Bernoulli}}(\mu_0) = c\,\mathbb{H}(\mu_0) \!-\!(n\!+\!c)\mathbb{H}(\mu_{n,c}(\mu_0)) + \log\frac{\mathbf{B}(c\mu_0,c(1-\mu_0))}{\mathbf{B}((n+c)\mu_{n,c}(\mu_0),(n+c)(1-\mu_{n,c}(\mu_0))},
\end{align*}
where $\mathbf{B}(\alpha, \beta)\!=\!\int_0^1u^{\alpha-1}(1\!-\!u)^{\beta-1}du$ is the Beta function and $\mathbb{H}(\cdot)$ the Bernoulli entropy function.

\paragraph{Exponential.}
Let $X \sim \text{Exp}(1/\mu)$, with unknown mean $\mu > 0$. For reference parameters $\mu_0,c > 0$, the Bregman information gain 
reads
\begin{align*}
\gamma_{n,c}^{\text{Exp}}(\mu_0) = \log\left(S_n/\mu_0+c \right) + \log\frac{\Gamma(c)}{\Gamma(n+c)}+(n+c-1)\log(n+c)-c\log c-n~.
\end{align*}
\paragraph{Pareto.} Let $X\sim \text{Pareto}\left(\alpha\right)$, with unknown shape $\alpha>0$. Define $L_n\!=\!\sum_{t=1}^{n}\log X_t$. Then, for reference parameters $\alpha_0,c > 0$, the Bregman information gain 
is given by
\begin{align*}
\gamma_{n,c}^{\text{Pareto}}(\alpha_0) \!=\! \log\left(\alpha_0L_n\!+\!c \right) \!+\!\log\frac{\Gamma(c)}{\Gamma(n+c)}\!+\!(n\!+\!c\!-\!1)\log(n+c)\!-\!c\log c\!-\!n.
\end{align*}
\paragraph{Chi-square.} Let $X\!\sim\! \chi^2(k)$, where $k \!\in\! \Nat$ is unknown. Define $K_n\!=\!\sum_{t=1}^n \log \frac{X_t}{2}$. For reference points $k_0 \in \Nat, c > 0$, introduce an estimate $k_{n,c}(k_0)$ satisfying $\psi_0\!\left(\!\frac{k_{n,c}(k_0)}{2}\!\right)\!=\!\frac{K_n+c \;\psi_0\left(\frac{k_0}{2}\right)}{n+c}$. In this case, the Bregman information gain is given by 
\begin{align*}
\gamma_{n,c}^{\chi^2}(k_0)&=\frac{k_{n,c}(k_0)}{2}\Big(K_n\!+\!c\psi_0\Big(\frac{k_0}{2}\Big)\Big)\!-\!(n\!+\!c)\log \Gamma\Big(\frac{k_{n,c}(k_0)}{2}\Big) \!+\!c\log \Gamma\Big(\frac{k_0}{2}\Big)\!-\!c\;\frac{k_0}{2}\psi_0\Big(\frac{k_0}{2}\Big)\\
&\!+\!\log \frac{J(c, c\psi_0(\frac{k_0}{2}))}{J(n\!+\!c, K_n\!+\!c\psi_0(\frac{k_0}{2}))}\,,\,\text{where}\,\, J(a, b) \!:= \!\sum\limits_{k'=1}^{\infty} \exp\left( \!-\!a \log \Gamma\left(\! \frac{k'}{2}\!\right)\! +\! b\frac{k'}{2}\!\right)\!.
\end{align*}
The function $J(a,b)$ can be estimated using numerical methods (see Section~\ref{sec:xps}). This paves a way to build high-probability confidence sets for Chi-square distribution which has not been adequately captured in prior work. The sums over $k'$ above derive from the martingale construction of Theorem~\ref{thm:genericmom} with discrete mixture (i.e., w.r.t the counting measure). A continuous (i.e., w.r.t the Lebesgue measure) mixture similar to the other families is also possible, as detailed in the appendix in Remark~\ref{rmk:chi2_gamma}.

\paragraph{Explicit confidence sets.}
We now turn to illustrate the confidence sets in the same exponential families as above. They are obtained by specifying the generic form and simplifying the resulting expression. We provide
the confidence sets for two-parameter Gaussian (i.e., unknown mean and variance), Bernoulli, Exponential, Pareto and Chi-square distributions, respectively,
in Table~\ref{table:bregman_cs_partial}. The technical details of the derivation of the specific forms for each illustrative family is postponed to Appendix~\ref{app:specific}, along with other distributions (Gamma, Poisson, Weibull) in Table~\ref{table:bregman_cs_full}.

\begin{small}

\begin{table}
  \caption{Bregman confidence sets given by Theorem~\ref{thm:genericmom} for representative families}
  \label{table:bregman_cs_partial}
  \centering
  \begin{tabular}{lll}
    \toprule
    Distribution     & Parameters     & Confidence Set \\
    \midrule
    Gaussian & \makecell[l]{$\mu\in\bR$\\ $\sigma\in\bR_+$} & 
    \makecell[l]{$\frac{1}{2}Z_n(\mu, \sigma) \!-\!\frac{n+c+3}{2}\log\left(\frac{n}{n\!+\!c} Z_n(\hat \mu_n, \sigma) \!+\! \frac{c}{n\!+\!c}Z_n(\mu, \sigma) \!+\! c \right)$ \\ $\quad\leq \log\frac{1}{\delta} \!-\! \frac{n}{2}\log 2 \!-\! \left(\frac{c}{2}\!+\!2\right)\log c \!+\! \frac{1}{2}\log \left(n\!+\!c\right)\!+\! \log\frac{\Gamma\left(\frac{c+3}{2}\right)}{\Gamma\left(\frac{n+c+3}{2}\right)}$} 
    \label{eqn:gauss-mean--var_full} \\[7.0ex]
    Bernoulli & $\mu\in[0, 1]$  & \makecell[l]{$S_n \log \!\frac{1}{\mu} \!+\! (n\!-\!S_n)\log \!\frac{1}{1-\mu} \!+\! \log \!\frac{\Gamma(S_n+c\mu)\Gamma(n-S_n+c(1-\mu))}{\Gamma(c\mu)\Gamma(c(1-\mu))}$ \\ $\quad\leq \log \!\frac{1}{\delta} \!+\! \log\! \frac{\Gamma(n+c)}{\Gamma(c)}$} \label{eqn:confsetBernoulli_full} \\[6.0ex]
    Exponential & $\mu\in\bR_+$  & \makecell[l]{$\frac{S_n}{\mu}\!-\!(n\!+\!c\!+\!1) \log\left(\frac{S_n}{\mu}+c \right)$ \\ $\quad\leq \log \frac{1}{\delta} \!+\! \log \frac{\Gamma(c)}{\Gamma(n+c)}\!-\! \log (n\!+\!c)  \!-\! c \log c$} \label{eqn:confsetExpo_full} \\[6.0ex]
    Pareto & $\alpha\in\bR$  & \makecell[l]{$\alpha L_n-(n\!+\!c\!+\!1)\log\left( \alpha L_n+c\right)$ \\ $\quad\leq \log \frac{1}{\delta}+ \log \frac{\Gamma(c)}{\Gamma(n+c)}-\log(n+c)-c\log c$} \label{eqn:confsetPareto_full} \\[6.0ex]
    Chi-square & $k\in\Nat$  & \makecell[l]{$n \log \Gamma \left( \frac{k}{2}\right)\!-\!\frac{k}{2}K_n\nonumber \!-\! \log J\left(c, c\psi_0\left( \frac{k}{2}\right)\right)$ \\ $\quad + \log J\left(n\!+\!c, K_n\!+\!c\psi_0\left( \frac{k}{2}\right)\right)\leq \log \frac{1}{\delta}$} \label{eqn:confsetXhi2_full} \\[2.5ex]
    \bottomrule
  \end{tabular}
  
\end{table}

\end{small}

\begin{figure}[t]

    \centering
	\begin{subfigure}[t]{.45\linewidth}
		\includegraphics[width=1\linewidth, height=1.0\linewidth]{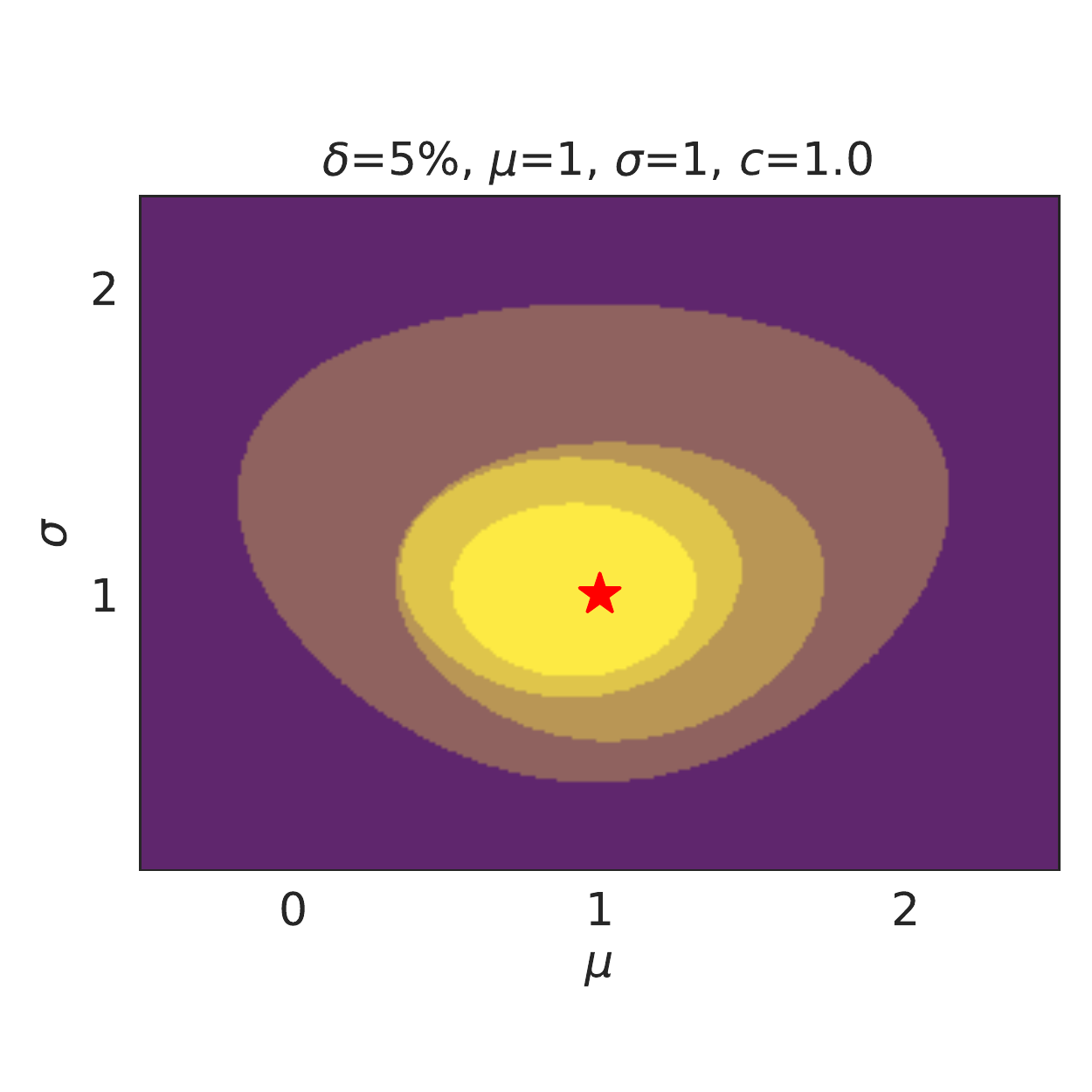}
		\caption{Gaussian (mean and variance)}\label{fig:conf_gaussian2d}
	\end{subfigure}
    \begin{subfigure}[t]{.45\linewidth}
		\includegraphics[width=1\linewidth,height=1\linewidth]{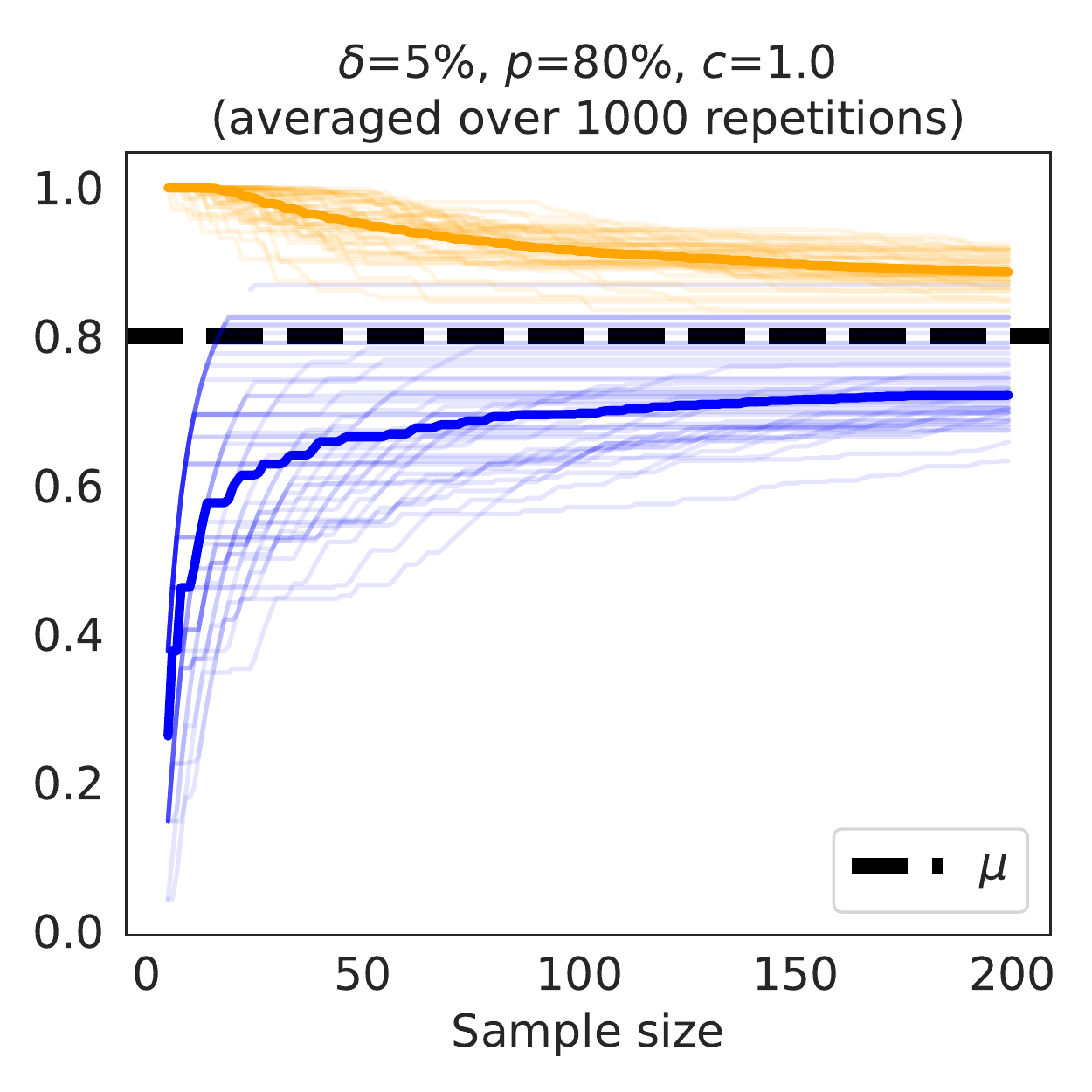}
		\caption{Bernoulli}
	\end{subfigure}\\
    \vspace{3mm}
	\begin{subfigure}[t]{.45\linewidth}
		\includegraphics[width=1\linewidth,height=1\linewidth]{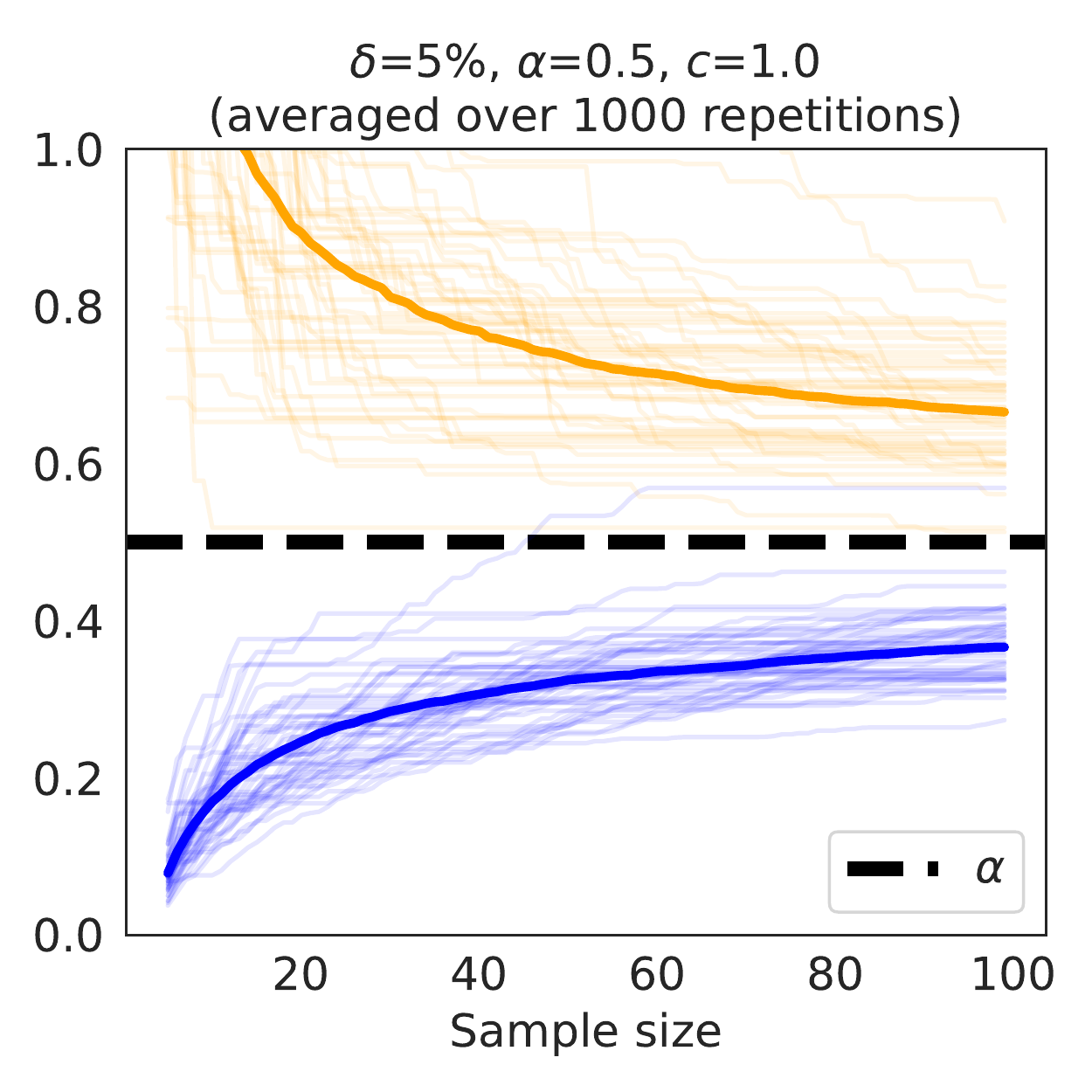}
		\vskip -2mm
		\caption{Pareto}
	\end{subfigure}
	\begin{subfigure}[t]{.45\linewidth}
		\includegraphics[width=1\linewidth,height=1\linewidth]{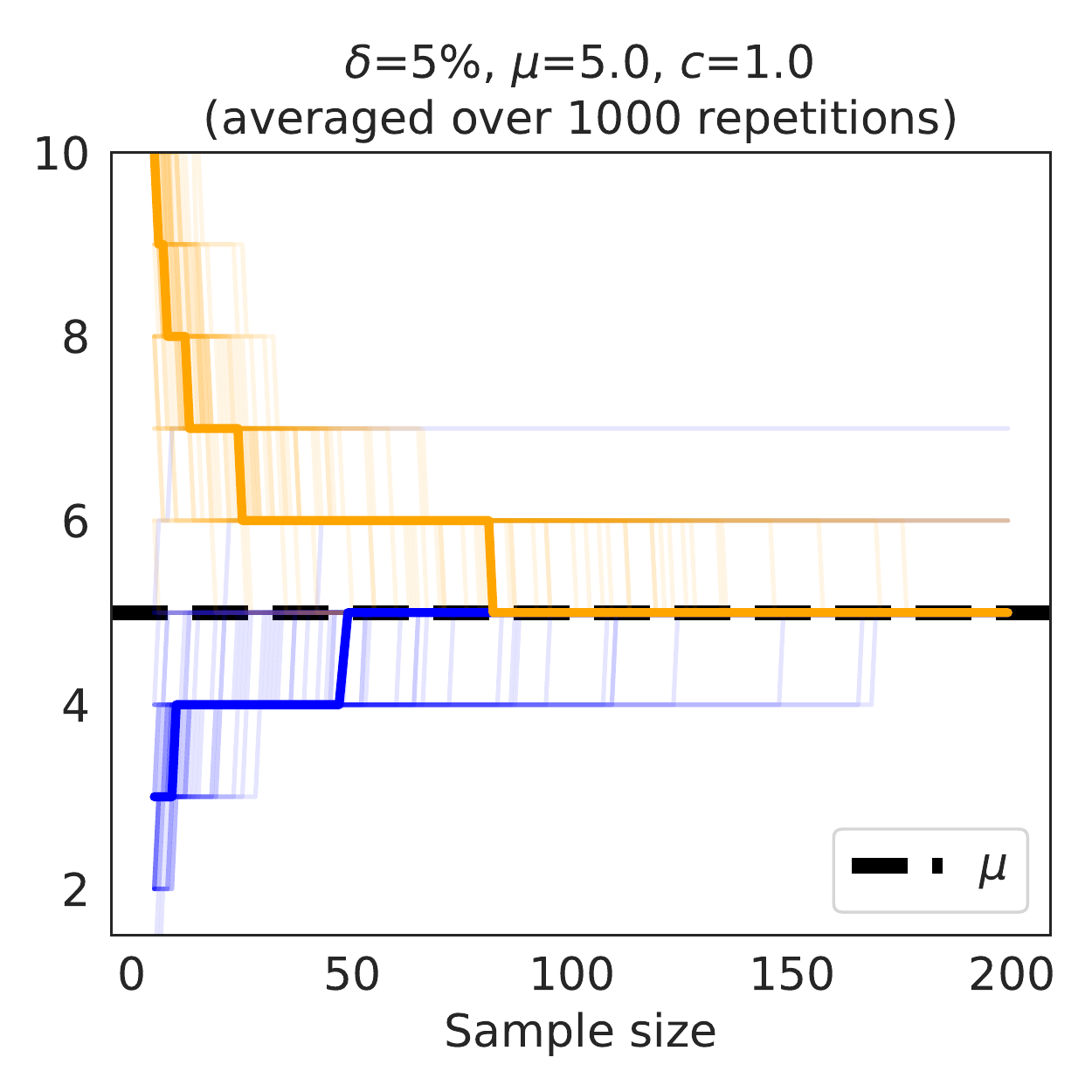}
		\caption{Chi-square}
	\end{subfigure}
    \caption{Example of Gaussian time-uniform joint confidence sets for $(\mu, \sigma)=(1, 1)$ with sample sizes $n\in\left\lbrace 10, 25, 50, 100 \right\rbrace$ observations (smaller confidence sets correspond to larger sample sizes), and examples of time-uniform confidence envelopes for $\text{Bernoulli}(0.8)$, $\text{Pareto}(0.5)$ and $\chi^2(5)$ on several realizations as a function of the number of observations $n$. Thick lines indicate the median curve over 1000 replicates.
    }
	\label{fig:confidencebands_examples}
\end{figure}

We now provide a set of illustrative numerical experiments to display the confidence envelopes resulting from Theorem~\ref{thm:genericmom} in the case of classical exponential families. We plot for a given value of the confidence level $\delta$ and regularization parameter $c$ the convex sets 
$n\mapsto \bigcap\nolimits_{n'\leq n}\Theta_{n',c}(\delta)$ (taking the running intersection is standard in sequential testing and was pioneered in \citet{darling1967iterated}; it ensures that the upper (resp. lower) confidence envelopes are nonincreasing (resp. nondecreasing) as the sample size grows, thus providing tighter bounds). In dimension $d=1$, we report the extremal points of these intervals, which we call the upper and lower envelopes respectively. We refer to Figure~\ref{fig:confidencebands_examples} for Pareto, Chi-square and Gaussian (with unknown $\mu,\sigma$) and Appendix~\ref{app:xps} for many other families. Because we exploit the Bregman geometry, our bounds capture a larger setting than typical mean estimation; for instance, we are able to concentrate around the exponent $\alpha$ of a Pareto distribution even when the distribution is not integrable ($\alpha\!<\!1$). Furthermore, for two-parameter Gaussian, apart form being \emph{anytime}, our confidence sets are convex and bounded in contrast to the one based on Chi-square quantiles with a crude union bound (see Appendix~\ref{app:gauss}).

\subsection{Proof Sketch: Theorem~\ref{thm:genericmom}}\label{sub:proof}
We now sketch the proof of Theorem~\ref{thm:genericmom}, and refer the interested reader to the appendix for details.

\paragraph{Step 1: Martingale construction.}
For any $\lambda\in\Real^d$, we introduce the quantity
	\beqan
	M_n^\lambda &=& \exp\left(\langle \lambda, n(\mu_n\!-\!\mu)\rangle \!-\!n \cB_{\cL,\theta}(\lambda)\right)\,,
	\eeqan
	where $\mu_n \!=\! \frac{1}{n}\sum_{t=1}^{n}\! F(X_t)$ and $\mu\!=\!\Esp_\theta[F(X)]$. By Lemma~\ref{lem:bregdual}, $M_n^\lambda$ is a martingale such that $\Esp[M_n^\lambda]\!=\!1$. We now introduce the distribution $q(\theta|\alpha,\beta)  = \exp(\langle \theta, \alpha\rangle - \beta \cL(\theta)) H(\alpha,\beta)$ where $H(\cdot,\cdot)$ is the normalization function, and define the mixture martingale $M_n = \int M_n^\lambda q(\theta+\lambda|\alpha,\beta)d\lambda$. Note that $M_n$ also satisfies $\Esp[M_n]=1$. 
	
	\paragraph{ Step 2. Choice of parameters and duality properties.}
	Choosing $\alpha=c \nabla \cL(\theta)$ and $\beta = c$ yields
	\beqan
	M_n \!=\!G(\theta,c)\!\int\!\!\exp\!\left(\! \langle \lambda, n(\mu_n\!-\!\mu)\rangle\! -\! (n\!+\!c)\cB_{\cL,\theta}(\lambda)\!\right)d\lambda\,,
	\eeqan
	where $G(\theta,c) = 1/\int \exp\big(-c\cB_{\cL}(\theta',\theta)\big)d\theta'$. Now, we consider the function $\cB_{\cL,\theta}^\star(x) =  \max_\lambda \langle \lambda, x\rangle - \cB_{\cL,\theta}(\lambda)$. Note that its maximal point $\lambda^\star_x$ satisfies $x+ \nabla \cL (\theta) = \nabla \cL(\theta+\lambda^\star_x)$ for every $x$. In particular, for the choice $x = \frac{n}{n+c}(\mu_n-\mu)$, we have $
	  \frac{n}{n+c}\mu_n+ \frac{c}{n+c}\nabla \cL (\theta) = \nabla \cL(\theta+\lambda^\star_x)$,
	which justifies introducing the following regularized estimate
	\beqan
	\theta_{n,c}(\theta) = (\nabla \cL)^{-1}\left(\frac{n}{n+c}\mu_n+ \frac{c}{n+c}\nabla\cL (\theta)\right),
	\eeqan 
	provided that $\nabla \cL$ is one-to-one in $\Theta$.
    
	\paragraph{ Step 3. Martingale rewriting and conclusion.} Note that the duality property also yields $
	\langle \lambda , x\rangle -\cB_{\cL,\theta}(\lambda) -
	\langle \lambda^\star_x , x\rangle +\cB_{\cL,\theta}(\lambda^\star_x) =
	\langle \lambda -\lambda^\star_x, \nabla \cL(\theta+\lambda^\star_x)\rangle + \cL(\theta+\lambda^\star_x)-\cL(\theta+\lambda)$ for each $x$. Using this property, we obtain
	\beqan
	M_n =\exp\big( (n\!+\!c)\cB_{\cL,\theta}^\star(x)\big) \frac{G(\theta,c)}{G( \theta_{n,c}(\theta),n+c)}\,.
	\eeqan
	After applying Markov's inequality, we obtain the following inequality:
	\beqan
	\Pr_{\theta}\bigg[\cB_{\cL,\theta}^\star \bigg(\frac{n}{n\!+\!c}(\mu_n-\mu)\bigg)
	\geq \frac{1}{n\!+\!c} \log\frac{G( \theta_{n,c}(\theta),n\!+\!c)}{\delta\, G(\theta,c)} \bigg] \leq \delta\,.
	\eeqan
	Now, we use duality property of Bregman divergence to obtain $\cB_{\cL,\theta}^\star\!\left(\! \frac{n}{n+c}(\mu_n\!-\!\mu)\right) \!=\! \cB_{\cL}\big(\theta, \theta_{n,c}(\theta)\big)$, which yields the desired form. Finally, by properties of the martingale $M_n$, the deviation bound also holds for any random stopping time $N$, and hence we obtain the time uniform bound by employing a stopping time construction similar to that of \citet{pena2008self,abbasi2011improved}.

\section{Numerical Experiments}\label{sec:xps}

In  this section, we provide illustrative numerical results to show the resulting confidence bands built from Theorem~\ref{thm:genericmom} are promising alternatives to existing competitors (detailed in Appendix~\ref{app:xps}). All the confidence sets presented here are implemented in the open source \textit{concentration-lib} Python package (\url{https://pypi.org/project/concentration-lib/}).

\begin{figure}[t]
    \centering
	\begin{subfigure}[t]{.45\linewidth}
		\includegraphics[width=1\linewidth,height=1\linewidth]{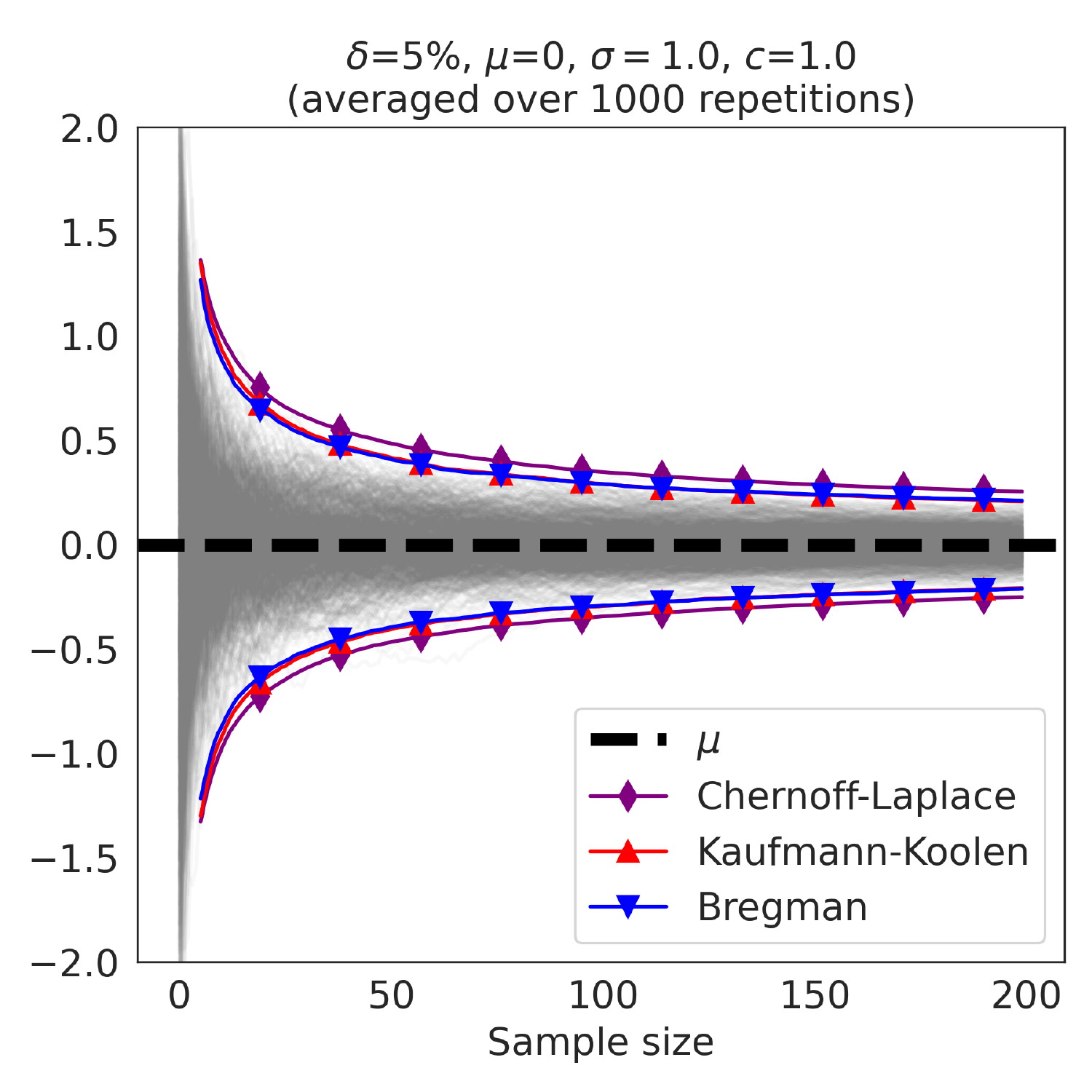}
		\caption{Gaussian}
	\end{subfigure}
	\begin{subfigure}[t]{.45\linewidth}
		\includegraphics[width=1\linewidth,height=1\linewidth]{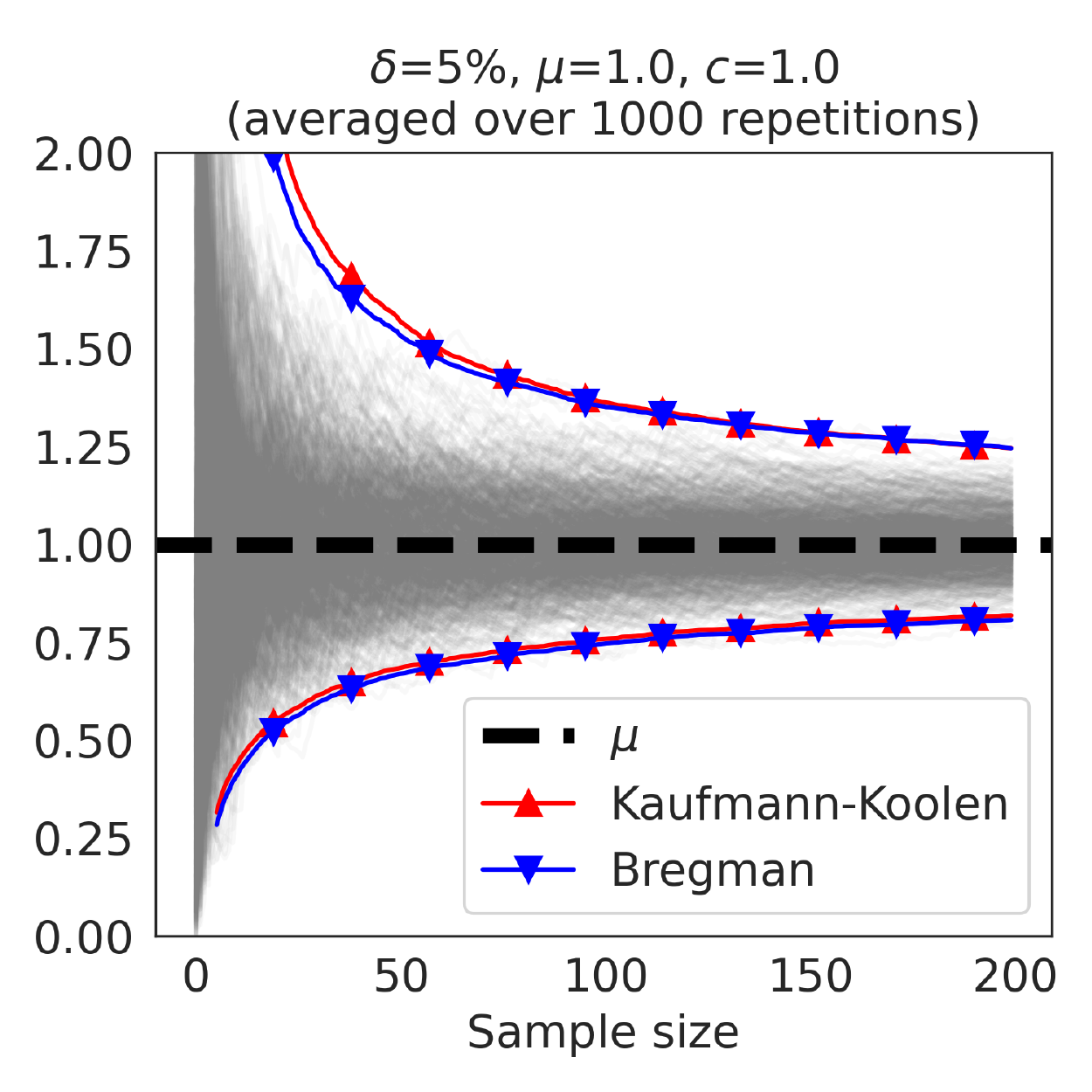}
		\caption{Exponential}
	\end{subfigure}\\
    \vspace{3mm}
     \begin{subfigure}[t]{.45\linewidth}
		\includegraphics[width=1\linewidth,height=1\linewidth]{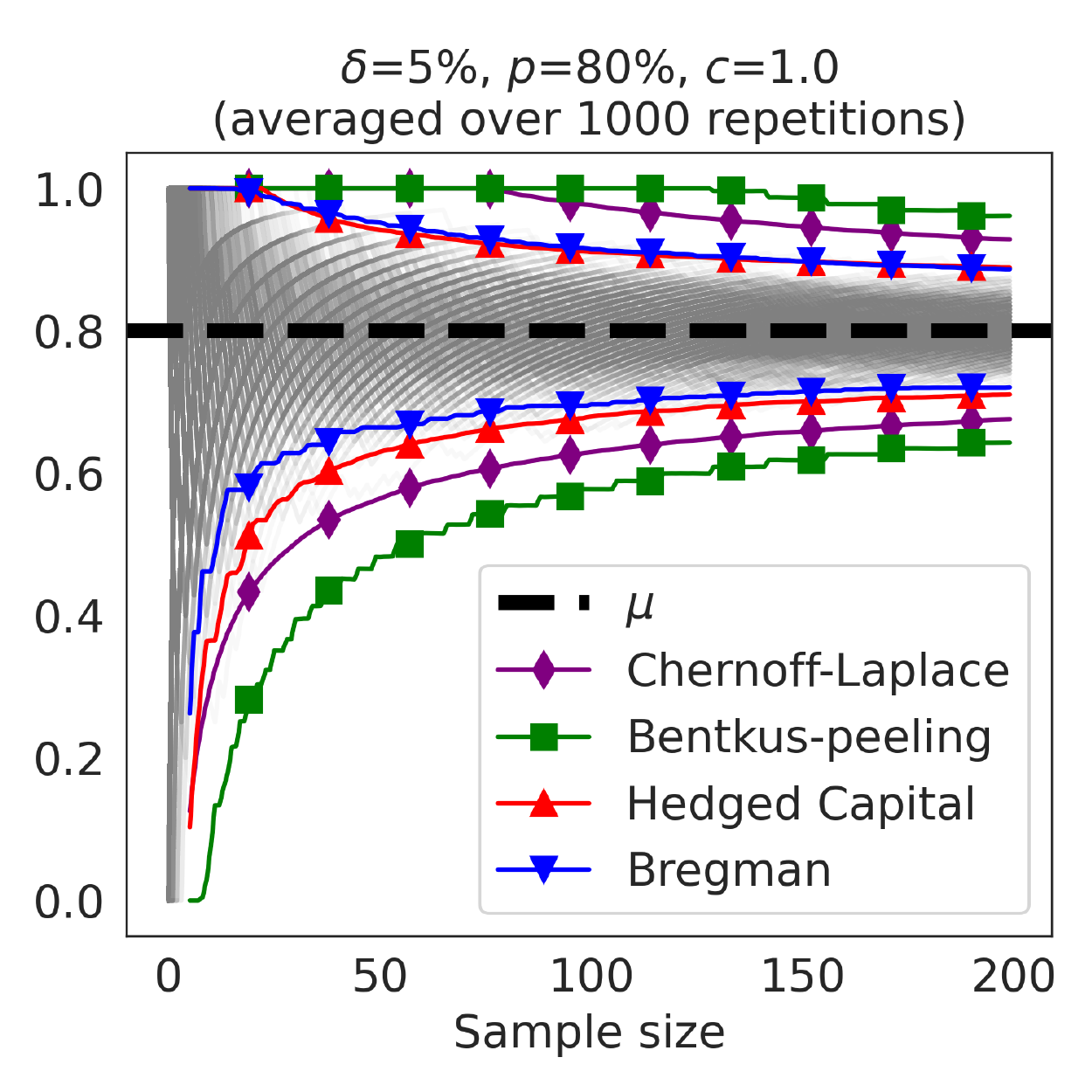}
		\caption{Bernoulli}
	\end{subfigure}
	\begin{subfigure}[t]{.45\linewidth}
	\includegraphics[width=1\linewidth,height=1\linewidth]{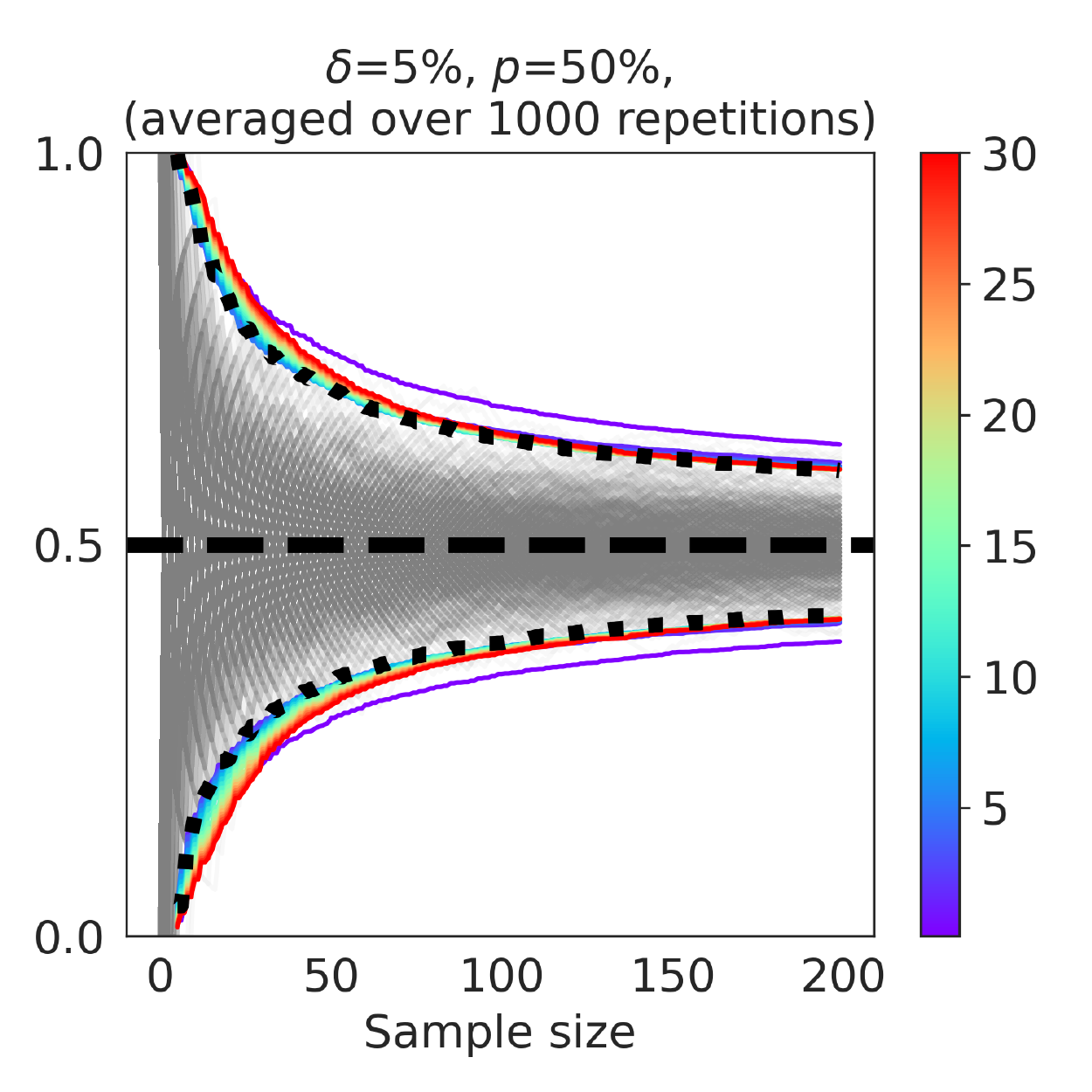}	
    \caption{Bernoulli (tuning $c$)}\label{fig:c_sensi}
	\end{subfigure}
	
    \caption{Comparison of median confidence envelopes around the mean for $\cN(0, 1)$, Bernoulli$(0.8)$ and $\text{Exp}(1)$, as a function of the sample size $n$, over $1000$ independent replicates. Grey lines are trajectories of empirical means $\widehat{\mu}_n$. Bottom right: confidence envelopes for varying regularization $c\!\in\![0.1, 30]$ for Bernoulli$(0.5)$ distribution. Dotted black line: heuristic $c_n \!\approx\! 0.12 n$.
    }
    \label{fig:confidencebands_comparison}
\end{figure}

\paragraph{Comparison.}
We compare our confidence envelopes to state-of-the-art time-uniform bounds in Figure~\ref{fig:confidencebands_comparison} and Appendix~\ref{app:xps}. Interestingly, our upper and lower bounds are not necessarily symmetrical, and in the Bernoulli case, they fit within the distribution support without clipping, contrary to most other methods, thus adapting to the local geometry of the family. Our bounds are comparable to those of \citet{kaufmann2021mixture}, if slightly tighter; however, we emphasize again that our scope is wider and captures more distributions.

\paragraph{Numerical complexity.}

Some of our formulas (Poisson, Chi-square) require the evaluation of integrals for which no closed-form expression exist. However, these can be estimated up to arbitrary precision by numerical routines, see Remarks \ref{rmk:chi2_compute} and \ref{rmk:poisson_compute} for details. Similarly, the use of special functions (digamma, ratio of Gamma) may lead to numerical instabilities or overflows for large $n$; we recommend instead to use the log-Gamma function and an efficient implementation of the log-sum-exp operator. We report results for $n\!\leq\!200$ as we think the small sample regime is where our bounds shine (most reasonable methods produce similar confidence sets for large $n$).
Finally, most confidence sets reported in Figure~\ref{fig:confidencebands_comparison} are implicitly defined (after reorganizing terms) as level sets $\lbrace F(\theta)\leq 0\rbrace$ for some function $F$. For one-dimensional families, we use a root search routine for a fast estimation of the boundaries of such sets; in higher dimension, we evaluate $F$ on a uniform grid (e.g, in Figure~\ref{fig:conf_gaussian2d}, we use a $1024\!\times\!1024$ grid over $[\!-\!2, 4]\!\times\![0.1, 4]$).

\paragraph{Influence of the regularization parameter $c$.}
Here, we fix $\delta\!=\!0.05$ (note that all bounds exhibit the typical $\log\frac{1}{\delta}$ dependency). The parameter $c$ is fixed to $c\!=\!1$ and can be interpreted as a number of virtual prior samples, thus introducing a bias. The choice $c\!=\!1$ is a typical value considered in literature on Gaussian concentration. To see the influence of the parameter $c$ on the bounds, we provide a second set of experiments, in which we compute confidence envelopes for $c$ varying between $0.1$ and $30$ (Figure~\ref{fig:c_sensi}). We find that $c$ has limited influence on the width of the confidence interval.
To get more insight on the tuning of $c$, we perform additional experiments in Appendix~\ref{app:xps}, where we optimize the confidence width w.r.t. to $c$ for varying sample sizes $n$. The resulting choice $c^*_n\!=\!\argmin_{c} \lvert \Theta_{n, c}\left(\delta\right)\rvert$ exhibits a linear trend $c^*_n\!\approx\!0.12 n$, which seems consistent across the tested distributions. Of note, the constant $0.12$ is reminiscent of the optimal tuning of the sub-Gaussian Laplace method \citep{howard2020time}. This is consistent with the observation in Section~\ref{sub:main-result} that the confidence width $\lvert \Theta_{n, c}\left(\delta\right)\rvert$ is equivalent to the Gaussian confidence width when $n\rightarrow +\infty$. We report in Figure~\ref{fig:c_sensi} this heuristic for $c$. However, we restate that sample a size-dependent regularization parameter $c$ is \textit{not} supported by the theory (as it would violate the law of iterated logarithms). Now, this does not prevent from choosing a specific horizon of interest $n_0$ and set $c\!=\!0.12n_0$ to promote sharpness around $n=zn_0$.

\section{Application: Generalized Likelihood Ratio Test in Exponential Families}\label{sec:GLR}

In this section, we apply our result to control the false alarm rate of Generalized Likelihood Ratio (GLR) tests when detecting a change of measure from a sequence of observations.
The GLR test in the exponential family model $\cE$ is defined, for a threshold $\alpha > 0$, as 
\beqan
\tau(\alpha;\cE) = \min\{ t \in \Nat : \max_{s\in[0,t)} G^\cE_{1:s:t}\geq \alpha \},
\eeqan
where the GLR is defined to be $G^\cE_{1:s:t} \!=\! \inf_{\theta'}\sup_{\theta_1,\theta_2} \log \bigg(\!\frac{ \prod_{t'=1}^s p_{\theta_1}(X_{t'})\!\prod_{t'=s+1}^t p_{\theta_2}(X_{t'})  }{ \prod_{t'=1}^t p_{\theta'}(X_{t'})} \!\bigg)$.

Hereafter, we consider that all the observations $(X_{t'})_{t'\in\Nat}$ come from a distribution with same parameter $\theta \in \Theta$. Then, the false alarm rate of the GLR test can be bounded using
\beqan
\Pr_{\theta}(\tau(\alpha;\cE)<\infty) \!=\! \Pr_{\theta}\big( \exists (t,s)\in \Nat^2, s<t :  G^\cE_{1:s:t} \geq \alpha\big).
\eeqan
Observe that solving for the inner supremums in the expression of GLR, we obtain 
\beqan
G^\cE_{1:s:t} &=& \inf_{\theta'}  s \cB_\cL(\theta',\theta_{1:s}) + (t-s)\cB_\cL(\theta',\theta_{s+1:t}),
\eeqan
where we introduce parameter estimates $\theta_{1:s}$ and $\theta_{s+1:t}$ such that $\nabla \cL(\theta_{1:s}) = \frac{1}{s}\sum_{t'=1}^s F(X_{t'})$ and $\nabla \cL(\theta_{s+1:t}) = \frac{1}{t-s}\sum_{t'=s+1}^t F(X_{t'})$, respectively.
Hence, the false alarm rate can be controlled as
\beqan
\Pr_{\theta}(\tau(\alpha;\cE)\!<\!\infty) \!\leq\! 
\Pr_{\theta}\Big(\! \exists s\!\in\! \Nat\!:\!  s \cB_\cL(\theta,\theta_{1:s}) \!\geq\! \alpha_1\!\Big)
\!+\!\Pr_{\theta}\Big(\! \exists (t,s)\!\in\! \Nat^2,\! s\!<\!t \!:\!  (t\!-\!s)\cB_\cL(\theta,\theta_{s+1:t}) \!\geq\! \alpha_2\!\Big)
\eeqan
for appropriate terms $\alpha_1,\alpha_2$ such that $\alpha=\alpha_1+\alpha_2$.
The control of the first term 
comes naturally from our time-uniform deviation result (Theorem~\ref{thm:genericmom}) using a regularized version of the estimate $\theta_{1:s}$. 
For the second term, we need to study the concentration of $\cB_\cL(\theta,\theta_{s+1:t})$ uniformly over $s$ and $t$.

\paragraph{Doubly time-uniform concentration.}
For any $s\!<\!t$,  $c>0$ and reference parameter $\theta_0 \in \Theta$, we define the regularized estimate
$\theta_{s+1:t,c}(\theta_0) = (\nabla \cL)^{-1}\left(\frac{1}{t-s+c}\left(\sum_{t'=s+1}^t F(X_{t'}) + c \nabla \cL(\theta_0)\right)\right)$ built from $(t\!-\!s)$ observations $X_{s+1},\ldots,X_t$. Similarly, we introduce the corresponding Bregman information gain $\gamma_{s+1:t,c}(\theta_0)$ as in Definition~\ref{def:info-gain}. The following result extends the Laplace method for exponential families (Theorem~\ref{thm:genericmom}) to control the Bregman deviation of $\theta_{s+1:t,c}(\theta_0)$ around $\theta_0$.

\begin{restatable}[Doubly time-uniform concentration]{theorem}{thmthree}\label{thm:Doubly-uniform}
Let $\delta\!\in\!(0,1]$ and
$g\colon\bN\!\rightarrow\!\bR_+^*$ such that $\sum_{t=1}^{\infty} 1/g(t)\!\leq\!1$ 
(e.g., $g(t)\!=\!\kappa (1\!+\!t) \log^2(1\!+\!t)$ where $\kappa=2.10975$).
Then, under the assumptions of Definition~\ref{def:info-gain}, it holds that
\beqan
\Pr_{\theta}\!\left[\exists t \in \Nat, \exists s\!<\!t:\! (t\!-\!s\!+\!c)\cB_\cL(\theta,\theta_{s+1:t,c}(\theta)) \!\geq\! \log \frac{g(t)}{\delta} + \gamma_{s+1:t,c}(\theta)\right] \leq \delta\,.
\eeqan
\end{restatable}





\begin{remark}
\citet{Maillard2018GLR} proves a doubly uniform concentration inequality for means of (sub)-Gaussian random variables and conjectures that it could be extended to other types of changes, such as changes of variance in a Gaussian family. Theorem~\ref{thm:Doubly-uniform} can be seen as a generalization of this result to generic exponential families. This naturally comes with additional challenges due to the intrinsic local geometry of exponential families. The proof of this result is deferred to Appendix~\ref{app:GLR}.
\end{remark}

\paragraph{A regularized GLR test.}
For a given false alarm (i.e., false positive) probability $\delta\in (0,1]$ and parameter $c >0$, we define the regularized GLR test $\tau^{\delta}_{c}(\cE)$ as follows:
\begin{align*}
&\Theta_{s, c}(\delta) = \left\lbrace \theta \in\Theta: (s+c) \cB_\cL(\theta,\theta_{1:s,c}(\theta)) \leq \log(1/\delta)+ \gamma_{1:s,c}(\theta) \right\rbrace\,,\\
&\Theta_{s+1:t, c}(\delta) = \left\lbrace \theta \in\Theta: (t-s+c)\cB_\cL(\theta,\theta_{s+1:t,c}(\theta)) \leq \log(g(t)/\delta) + \gamma_{s+1:t,c}(\theta) \right\rbrace\,,\\
&\tau^{\delta}_{c}(\cE) = \min\left\lbrace t \in \Nat : \exists s < t, \Theta_{s, c}(\delta/2) \cap \Theta_{s+1:t, c}(\delta/2) = \emptyset \right\rbrace \,.
\end{align*}
Note that, under the measure $\mathbb{P}_\theta$, $\theta\in\Theta_{s,c}(\delta/2)$ holds with probability at least $1-\delta/2$ by time-uniform concentration over $s$ (Theorem~\ref{thm:genericmom}), and $\theta\in\Theta_{s+1:t,c}(\delta/2)$ holds with probability at least $1\!-\!\delta/2$ by doubly time-uniform concentration over both $s$ and $t$ (Theorem~\ref{thm:Doubly-uniform}). Then, by a union-bound, this test is guaranteed to have a false alarm probability controlled by $\delta$. 
The factor $g(t)$ is essentially a function growing slightly faster than linearly; other common choices include $g(t)\propto t^{\eta}$ for some $\eta>1$, though it leads to a looser bound due to the faster growth.
Furthermore, if one is interested in detecting changes up to a known horizon $T$, one can replace the function $g(t)$ by the slightly tighter factor 
$g(t)=\kappa (1\!+\!t)\log^2(1\!+\!t)$ with $\kappa\!=\!\sum_{t=1}^T 1/g(t)$ 
and still guarantee a false alarm rate under $\delta$. In Appendix~\ref{app:xp_changepoint}, we report simulations of this GLR test to detect changes of variance in the family of centered Gaussian (Figure~\ref{fig:changepoint}), which answers an open question from \citet{Maillard2018GLR}.

\section{Conclusion}
We apply the method of mixture to derive a time-uniform deviation inequality for generic parametric exponential families expressed in terms of their Bregman divergences, highlighting the role of a quantity, the \emph{Bregman information gain}, that is related to the geometry of the family.
We specialize this general result to build confidence sets for classical examples. 
Our method compares favorably to the state-of-the art for Gaussian, Bernoulli and exponential distributions, and extends to other families like Chi-square, Pareto and two-parameter Gaussian, where no known time-uniform bound exists. Our method is also general enough to design GLR tests for change detection. 
An interesting direction for future work would be to consider the case where the log-partition function (and thus the Bregman divergence) is misspecified or unknown and has to be learned alongside the parameter.




\acks{
The authors acknowledge the funding of the French National Research Agency, the French Ministry of Higher Education and Research, Inria, the MEL and the I-Site ULNE regarding project R-PILOTE-19-004-APPRENF and Bandits
For Health (B4H). We thank the anonymous reviewers for their careful reading of the paper and their suggestions
for improvements. Experiments presented in this paper were carried out using the Grid’5000 testbed, supported by
a scientific interest group hosted by Inria and including CNRS, RENATER and several universities as well as other
organizations (see https://www.grid5000.fr).
}

\newpage

\bibliography{biblio.bib}

\begin{thebibliography}{46}
\providecommand{\natexlab}[1]{#1}
\providecommand{\url}[1]{\texttt{#1}}
\expandafter\ifx\csname urlstyle\endcsname\relax
  \providecommand{\doi}[1]{doi: #1}\else
  \providecommand{\doi}{doi: \begingroup \urlstyle{rm}\Url}\fi

\bibitem[Abbasi-Yadkori et~al.(2011)Abbasi-Yadkori, P{\'a}l, and
  Szepesv{\'a}ri]{abbasi2011improved}
Yasin Abbasi-Yadkori, D{\'a}vid P{\'a}l, and Csaba Szepesv{\'a}ri.
\newblock Improved algorithms for linear stochastic bandits.
\newblock In \emph{Advances in Neural Information Processing Systems}, pages
  2312--2320, 2011.

\bibitem[Amari(2016)]{amari2016information}
Shun-ichi Amari.
\newblock \emph{Information geometry and its applications}, volume 194.
\newblock Springer, 2016.

\bibitem[Auer et~al.(2002)Auer, Cesa-Bianchi, and Fischer]{Auer02}
P.~Auer, N.~Cesa-Bianchi, and P.~Fischer.
\newblock {Finite-time Analysis of the Multi-armed Bandit Problem}.
\newblock \emph{Machine Learning}, 47\penalty0 (2):\penalty0 235--256, 2002.

\bibitem[Basu et~al.(2022)Basu, Maillard, and Mathieu]{basu2022bandits}
Debabrota Basu, Odalric-Ambrym Maillard, and Timoth{\'e}e Mathieu.
\newblock Bandits corrupted by nature: Lower bounds on regret and robust
  optimistic algorithm.
\newblock \emph{arXiv preprint arXiv:2203.03186}, 2022.

\bibitem[Baudry et~al.(2020)Baudry, Kaufmann, and Maillard]{baudry2020sub}
Dorian Baudry, Emilie Kaufmann, and Odalric-Ambrym Maillard.
\newblock Sub-sampling for efficient non-parametric bandit exploration.
\newblock \emph{Advances in Neural Information Processing Systems},
  33:\penalty0 5468--5478, 2020.

\bibitem[Bentkus(2004)]{bentkus2004hoeffding}
Vidmantas Bentkus.
\newblock On hoeffding’s inequalities.
\newblock \emph{The Annals of Probability}, 32\penalty0 (2):\penalty0
  1650--1673, 2004.

\bibitem[Berend and Kontorovich(2013)]{berend2013concentration}
Daniel Berend and Aryeh Kontorovich.
\newblock On the concentration of the missing mass.
\newblock \emph{Electronic Communications in Probability}, 18:\penalty0 1--7,
  2013.

\bibitem[Boucheron et~al.(2013)Boucheron, Lugosi, and
  Massart]{boucheron2013concentration}
St{\'e}phane Boucheron, G{\'a}bor Lugosi, and Pascal Massart.
\newblock \emph{Concentration inequalities: A nonasymptotic theory of
  independence}.
\newblock Oxford university press, 2013.

\bibitem[Bregman(1967)]{bregman1967relaxation}
L.M. Bregman.
\newblock The relaxation method of finding the common point of convex sets and
  its application to the solution of problems in convex programming.
\newblock \emph{USSR Computational Mathematics and Mathematical Physics},
  7\penalty0 (3):\penalty0 200--217, 1967.
\newblock ISSN 0041-5553.

\bibitem[Bubeck(2010)]{bubeck2010bandits}
S{\'e}bastien Bubeck.
\newblock \emph{Bandits games and clustering foundations}.
\newblock PhD thesis, 2010.

\bibitem[Capp{\'e} et~al.(2013)Capp{\'e}, Garivier, Maillard, Munos, and
  Stoltz]{cappe2013kullback}
Olivier Capp{\'e}, Aur{\'e}lien Garivier, Odalric-Ambrym Maillard, R{\'e}mi
  Munos, and Gilles Stoltz.
\newblock Kullback-leibler upper confidence bounds for optimal sequential
  allocation.
\newblock \emph{The Annals of Statistics}, pages 1516--1541, 2013.

\bibitem[Carpentier et~al.(2011)Carpentier, Lazaric, Ghavamzadeh, Munos, and
  Auer]{carpentier2011upper}
Alexandra Carpentier, Alessandro Lazaric, Mohammad Ghavamzadeh, R{\'e}mi Munos,
  and Peter Auer.
\newblock Upper-confidence-bound algorithms for active learning in multi-armed
  bandits.
\newblock In \emph{International Conference on Algorithmic Learning Theory},
  pages 189--203. Springer, 2011.

\bibitem[Cesa-Bianchi and Lugosi(2006)]{cesa2006prediction}
Nicolo Cesa-Bianchi and G{\'a}bor Lugosi.
\newblock \emph{Prediction, learning, and games}.
\newblock Cambridge university press, 2006.

\bibitem[Chowdhury and Gopalan(2017)]{chowdhury2017kernelized}
Sayak~Ray Chowdhury and Aditya Gopalan.
\newblock On kernelized multi-armed bandits.
\newblock In \emph{International Conference on Machine Learning}, pages
  844--853. PMLR, 2017.

\bibitem[Chowdhury et~al.(2021)Chowdhury, Gopalan, and
  Maillard]{chowdhury2021reinforcement}
Sayak~Ray Chowdhury, Aditya Gopalan, and Odalric-Ambrym Maillard.
\newblock Reinforcement learning in parametric mdps with exponential families.
\newblock In \emph{International Conference on Artificial Intelligence and
  Statistics}, pages 1855--1863. PMLR, 2021.

\bibitem[Darling and Robbins(1967)]{darling1967iterated}
DA~Darling and Herbert Robbins.
\newblock Iterated logarithm inequalities.
\newblock \emph{Proceedings of the National Academy of Sciences}, 57\penalty0
  (5):\penalty0 1188--1192, 1967.

\bibitem[Durand et~al.(2018)Durand, Maillard, and Pineau]{durand2017streaming}
Audrey Durand, Odalric-Ambrym Maillard, and Joelle Pineau.
\newblock Streaming kernel regression with provably adaptive mean, variance,
  and regularization.
\newblock \emph{Journal of Machine Learning Research}, 19\penalty0
  (1):\penalty0 650--683, 2018.

\bibitem[Durrett(2019)]{durrett2019probability}
Rick Durrett.
\newblock \emph{Probability: theory and examples}, volume~49.
\newblock Cambridge university press, 2019.

\bibitem[Dwork et~al.(2014)Dwork, Roth, et~al.]{dwork2014algorithmic}
Cynthia Dwork, Aaron Roth, et~al.
\newblock The algorithmic foundations of differential privacy.
\newblock \emph{Found. Trends Theor. Comput. Sci.}, 9\penalty0 (3-4):\penalty0
  211--407, 2014.

\bibitem[Faury et~al.(2020)Faury, Abeille, Calauz{\`e}nes, and
  Fercoq]{faury2020improved}
Louis Faury, Marc Abeille, Cl{\'e}ment Calauz{\`e}nes, and Olivier Fercoq.
\newblock Improved optimistic algorithms for logistic bandits.
\newblock In \emph{International Conference on Machine Learning}, pages
  3052--3060. PMLR, 2020.

\bibitem[Garivier and Kaufmann(2016)]{garivier2016optimal}
Aur{\'e}lien Garivier and Emilie Kaufmann.
\newblock Optimal best arm identification with fixed confidence.
\newblock In \emph{Conference on Learning Theory}, pages 998--1027. PMLR, 2016.

\bibitem[Garivier(2013)]{garivierICB13}
Aurélien Garivier.
\newblock Informational confidence bounds for self-normalized averages and
  applications.
\newblock In \emph{2013 IEEE Information Theory Workshop (ITW)}, pages 1--5,
  2013.
\newblock \doi{10.1109/ITW.2013.6691311}.

\bibitem[Hao et~al.(2019)Hao, Abbasi~Yadkori, Wen, and
  Cheng]{hao2019bootstrapping}
Botao Hao, Yasin Abbasi~Yadkori, Zheng Wen, and Guang Cheng.
\newblock Bootstrapping upper confidence bound.
\newblock \emph{Advances in Neural Information Processing Systems},
  32:\penalty0 12123--12133, 2019.

\bibitem[Hoeffding(1963)]{hoeffding1963}
Wassily Hoeffding.
\newblock Probability inequalities for sums of bounded random variables.
\newblock \emph{Journal of the American Statistical Association}, 58\penalty0
  (301):\penalty0 13--30, 1963.
\newblock ISSN 01621459.

\bibitem[Holland and Haress(2021)]{holland2021learning}
Matthew Holland and El~Mehdi Haress.
\newblock Learning with risk-averse feedback under potentially heavy tails.
\newblock In \emph{International Conference on Artificial Intelligence and
  Statistics}, pages 892--900. PMLR, 2021.

\bibitem[Howard et~al.(2020)Howard, Ramdas, McAuliffe, and
  Sekhon]{howard2020time}
Steven~R Howard, Aaditya Ramdas, Jon McAuliffe, and Jasjeet Sekhon.
\newblock Time-uniform chernoff bounds via nonnegative supermartingales.
\newblock \emph{Probability Surveys}, 17:\penalty0 257--317, 2020.

\bibitem[Howard et~al.(2021)Howard, Ramdas, McAuliffe, and
  Sekhon]{howard2021time}
Steven~R Howard, Aaditya Ramdas, Jon McAuliffe, and Jasjeet Sekhon.
\newblock Time-uniform, nonparametric, nonasymptotic confidence sequences.
\newblock \emph{The Annals of Statistics}, 2021.

\bibitem[Jaksch et~al.(2010)Jaksch, Ortner, and Auer]{jaksch2010near}
Thomas Jaksch, Ronald Ortner, and Peter Auer.
\newblock Near-optimal regret bounds for reinforcement learning.
\newblock \emph{Journal of Machine Learning Research}, 11\penalty0 (4), 2010.

\bibitem[Kaufmann and Koolen(2021)]{kaufmann2021mixture}
Emilie Kaufmann and Wouter~M Koolen.
\newblock Mixture martingales revisited with applications to sequential tests
  and confidence intervals.
\newblock \emph{Journal of Machine Learning Research}, 22:\penalty0 246--1,
  2021.

\bibitem[Kearns and Saul(1998)]{kearns1998large}
Michael Kearns and Lawrence Saul.
\newblock Large deviation methods for approximate probabilistic inference.
\newblock In \emph{Proceedings of the Fourteenth conference on Uncertainty in
  artificial intelligence}, pages 311--319, 1998.

\bibitem[Kirschner and Krause(2018)]{kirschner2018information}
Johannes Kirschner and Andreas Krause.
\newblock Information directed sampling and bandits with heteroscedastic noise.
\newblock In \emph{Conference On Learning Theory}, pages 358--384. PMLR, 2018.

\bibitem[Korda et~al.(2013)Korda, Kaufmann, and Munos]{korda2013thompson}
Nathaniel Korda, Emilie Kaufmann, and Remi Munos.
\newblock Thompson sampling for 1-dimensional exponential family bandits.
\newblock \emph{Advances in neural information processing systems}, 26, 2013.

\bibitem[Kuchibhotla and Zheng(2021)]{pmlr-v139-kuchibhotla21a}
Arun~K Kuchibhotla and Qinqing Zheng.
\newblock Near-optimal confidence sequences for bounded random variables.
\newblock In Marina Meila and Tong Zhang, editors, \emph{Proceedings of the
  38th International Conference on Machine Learning}, volume 139 of
  \emph{Proceedings of Machine Learning Research}, pages 5827--5837. PMLR,
  18--24 Jul 2021.

\bibitem[Lattimore and Szepesv{\'a}ri(2019)]{LattimoreBanditAlgorithmsBook}
T.~Lattimore and C.~Szepesv{\'a}ri.
\newblock \emph{{Bandit Algorithms}}.
\newblock {C}ambridge {U}niversity {P}ress, 2019.

\bibitem[Laurent and Massart(2000)]{laurent2000adaptive}
Beatrice Laurent and Pascal Massart.
\newblock Adaptive estimation of a quadratic functional by model selection.
\newblock \emph{Annals of Statistics}, pages 1302--1338, 2000.

\bibitem[Maillard(2019{\natexlab{a}})]{Maillard2018GLR}
O.-A. Maillard.
\newblock {Sequential change-point detection: Laplace concentration of scan
  statistics and non-asymptotic delay bounds}.
\newblock In \emph{Algorithmic Learning Theory (ALT)}, 2019{\natexlab{a}}.

\bibitem[Maillard(2019{\natexlab{b}})]{maillard19HDR}
Odalric-Ambrym Maillard.
\newblock \emph{{Mathematics of Statistical Sequential Decision Making}}.
\newblock Habilitation {\`a} diriger des recherches, {Universit{\'e} de Lille
  Nord de France}, February 2019{\natexlab{b}}.

\bibitem[Pe{\~n}a et~al.(2008)Pe{\~n}a, Lai, and Shao]{pena2008self}
Victor~H Pe{\~n}a, Tze~Leung Lai, and Qi-Man Shao.
\newblock \emph{Self-normalized processes: Limit theory and Statistical
  Applications}.
\newblock Springer Science \& Business Media, 2008.

\bibitem[Raginsky and Sason(2018)]{raginsky2012concentration}
Maxim Raginsky and Igal Sason.
\newblock \emph{Concentration of Measure Inequalities in Information Theory,
  Communications, and Coding}.
\newblock Now Publishers, 2018.

\bibitem[Robbins(1970)]{robbins1970statistical}
Herbert Robbins.
\newblock Statistical methods related to the law of the iterated logarithm.
\newblock \emph{The Annals of Mathematical Statistics}, 41\penalty0
  (5):\penalty0 1397--1409, 1970.

\bibitem[Robbins and Pitman(1949)]{robbins1949application}
Herbert Robbins and EJG Pitman.
\newblock Application of the method of mixtures to quadratic forms in normal
  variates.
\newblock \emph{The annals of mathematical statistics}, pages 552--560, 1949.

\bibitem[Shafer and Vovk(2019)]{shafer2019game}
Glenn Shafer and Vladimir Vovk.
\newblock \emph{Game-Theoretic Foundations for Probability and Finance}, volume
  455.
\newblock John Wiley \& Sons, 2019.

\bibitem[Shun and McCullagh(1995)]{shun1995laplace}
Zhenming Shun and Peter McCullagh.
\newblock Laplace approximation of high dimensional integrals.
\newblock \emph{Journal of the Royal Statistical Society: Series B
  (Methodological)}, 57\penalty0 (4):\penalty0 749--760, 1995.

\bibitem[Ville(1939)]{ville1939etude}
Jean Ville.
\newblock Etude critique de la notion de collectif.
\newblock \emph{Bull. Amer. Math. Soc}, 45\penalty0 (11):\penalty0 824, 1939.

\bibitem[Waudby-Smith and Ramdas(2023)]{waudby2020estimating}
Ian Waudby-Smith and Aaditya Ramdas.
\newblock Estimating means of bounded random variables by betting.
\newblock \emph{Journal of the Royal Statistical Society: Series B (Statistical
  Methodology)}, 2023.

\bibitem[Zeitouni and Dembo(1998)]{zeitouni1998large}
A~Dembo~O Zeitouni and A~Dembo.
\newblock Large deviations techniques and applications.
\newblock \emph{Applications of Mathematics}, 38, 1998.

\end{thebibliography}

\newpage

\begin{appendix}
\section{Proof of the Main Result About Bregman Deviations}\label{app:proof}

In this section, we detail the proof of the main results regarding Bregman deviation inequalities, stated with its two variants  Theorem~\ref{thm:genericmom} and \ref{thm:momEFsequence}, 
We start in Section~\ref{sub:c} with the proof of Theorem~\ref{thm:genericmom} when considering  a numerical constant $c$ to build the regularizer. In Section~\ref{sub:legendre}, we then provide the proof for Theorem~\ref{thm:momEFsequence} when considering instead a  Legendre function $\cL_0$ as a regularizer, and also extend to the non i.i.d.~case.

\subsection{Proof of Bregman concentration using mixture parameter $c$}\label{sub:c}
The proof of this first result follows an adaptation of the method of mixture, combined with properties of Bregman divergences recalled in Lemma~\ref{lem:bregdual}.

\thmone*

\begin{myproof}{of Theorem~\ref{thm:genericmom}}

	{\bf Step 1. Martingale and mixture martingale construction.}
	
	Let us note that $\Esp_\theta[F(X)] = \nabla \cL(\theta)$ and $\log \Esp_\theta\left[ \exp \langle \lambda , F(X)\rangle\right]  = \cL(\theta+\lambda)- \cL(\theta)$.
	Hence, we deduce that 
	\beqan
	\log \Esp_\theta\left[ \exp \langle \lambda , F(X)- \Esp_\theta[F(X)]\rangle\right]  = \cB_{\cL,\theta}(\lambda) \eqdef \cL(\theta+\lambda) -\cL(\theta) - \langle \lambda, \nabla \cL (\theta)\rangle\,.
	\eeqan
	Now, let $\mu_n = \frac{1}{n}\sum_{t=1}^{n} F(X_t)$ and $\mu=\Esp_\theta[F(X)]$. For any $\lambda\in\Real^d$, the following quantity
	\beqan
	M_n^\lambda &=& \exp\bigg(\langle \lambda, n(\mu_n-\mu)\rangle -n \cB_{\cL,\theta}(\lambda)\bigg)\,,
	\eeqan
	is thus a martingale such that $\Esp[M_n^\lambda]=1$. 
	
	We now introduce the distribution $q(\theta|\alpha,\beta)  = H(\alpha,\beta)\cdot\exp(\langle \theta, \alpha\rangle - \beta \cL(\theta)) $ where $H$ is the normalization term. We further introduce the following quantity
	\beqan
	M_n = \int_{\Lambda_\theta} M_n^\lambda q(\theta+\lambda|\alpha,\beta)d\lambda\,,
	\eeqan
	where $\Lambda_\theta = \{ \lambda : \theta+\lambda \in\Theta\}$. 
	This also satisfies $\Esp[M_n]=1$. Further, we have the rewriting
	\beqan
	M_n =H(\alpha, \beta) \int_{\Lambda_\theta}\exp\bigg( \langle \lambda, n(\mu_n-\mu)\rangle - n\cB_{\cL,\theta}(\lambda) + \langle \lambda +\theta, \alpha\rangle - \beta \cL(\theta+\lambda)\bigg)d\lambda \,.
	\eeqan
	
	{\bf Step 2. Choice of parameters and duality properties.}
	Considering in particular the choice $\alpha=c \nabla \cL(\theta)$ and $\beta = c$, we get 
	\beqan
	M_n &=& H(c \nabla \cL(\theta),c) \int_{\Lambda_\theta}\exp\bigg( \langle \lambda, n(\mu_n-\mu)\rangle - n\cB_{\cL,\theta}(\lambda) + c\langle \lambda +\theta, \nabla \cL(\theta)\rangle - c\cL(\theta+\lambda)\bigg)d\lambda\\
	&=&G(\theta,c)\int_{\Lambda_\theta}\exp\bigg( \langle \lambda, n(\mu_n-\mu)\rangle - (n+c)\cB_{\cL,\theta}(\lambda)\bigg)d\lambda ~,
	\eeqan
	where we introduce the following quantity
	\beqan
	G(\theta,c) = H(c \nabla \cL(\theta),c)\exp\bigg(c\langle \theta,\nabla \cL(\theta)\rangle-c \cL(\theta)\bigg) &=& \frac{\exp\bigg(c\langle \theta,\nabla \cL(\theta)\rangle-c \cL(\theta)\bigg)}{\int_\Theta \exp\bigg(c\langle \theta',\nabla \cL(\theta)\rangle-c \cL(\theta')\bigg)d\theta'}\\
	&=& \bigg[
		\int_\Theta \exp\big(-c\cB_{\cL}(\theta',\theta)\big)d\theta'\bigg]^{-1}~.
	\eeqan
	At this point, let us introduce $x = \frac{n}{n+c}(\mu_n-\mu)$. We also consider the Legendre-Fenchel dual function $\cB_{\cL,\theta}^\star(x) =  \max_\lambda \langle \lambda, x\rangle - \cB_{\cL,\theta}(\lambda)$
	and denote $\lambda^\star_x$ its maximal point.
	Now, we note that $x+ \nabla \cL (\theta) = \nabla \cL(\theta+\lambda^\star_x)$ and thus 
	\beqan
	\langle \lambda , x\rangle -\cB_{\cL,\theta}(\lambda) -
	\langle \lambda^\star_x , x\rangle +\cB_{\cL,\theta}(\lambda^\star_x) &=&
	\langle \lambda -\lambda^\star_x, x+ \nabla \cL (\theta)\rangle + \cL(\theta+\lambda^\star_x)-\cL(\theta+\lambda)\\
	&=&
	\langle \lambda -\lambda^\star_x, \nabla \cL(\theta+\lambda^\star_x)\rangle + \cL(\theta+\lambda^\star_x)-\cL(\theta+\lambda)\,.
	\eeqan 
	
	Also, since it holds that $x = \frac{n}{n+c}(\mu_n-\nabla \cL(\theta))$, then the quantity
	$x+ \nabla \cL (\theta) = \nabla \cL(\theta+\lambda^\star_x)$
	rewrites
	$\frac{n}{n+c}\mu_n+ \frac{c}{n+c}\nabla \cL (\theta) = \nabla \cL(\theta+\lambda^\star_x)$, 
	which justifies to introduce the following regularized parameter estimate
	\beqan
	\theta_{n,c}(\theta) = (\nabla \cL)^{-1}\bigg(\frac{n}{n+c}\mu_n+ \frac{c}{n+c}\nabla\cL (\theta)\bigg)\,.
	\eeqan 
	

	{\bf Step 3. Martingale rewriting and conclusion.}
	Using this property, we note that 
	\beqan
	M_n &=& \exp\big( (n\!+\!c)\cB_{\cL,\theta}^\star(x)\big)\cdot G(\theta,c)\!\!\int_{\Lambda_\theta} \!\!\exp\!\bigg(\!(n\!+\!c)\Big[\langle \lambda \!-\!\lambda^\star_x, \nabla \cL(\theta\!+\!\lambda^\star_x)\rangle\!-\!\cL(\theta\!+\!\lambda)\!+\! \cL(\theta\!+\!\lambda^\star_x)\Big]\!\bigg)d\lambda \\
	&=&\exp\big( (n\!+\!c)\cB_{\cL,\theta}^\star(x)\big)\cdot G(\theta,c) \!\int_{\Lambda_\theta}\! \exp\!\bigg( \langle \lambda\!+\!\theta, (n\!+\!c)\nabla \cL(\theta\!+\!\lambda^\star_x)\rangle\!-\!(n\!+\!c)\cL(\theta\!+\!\lambda)\bigg)d\lambda\times\\
	&&
	\exp\bigg(\!-(n\!+\!c)\langle\lambda^\star_x\!+\!\theta, \nabla \cL(\theta\!+\!\lambda^\star_x)+(n\!+\!c)\cL(\theta\!+\!\lambda^\star_x)\rangle\bigg)\\
	&=&\exp\big( (n\!+\!c)\cB_{\cL,\theta}^\star(x)\big) \frac{G(\theta,c)}{G(\theta+\lambda^\star_x,n+c)}=\exp\big( (n\!+\!c)\cB_{\cL,\theta}^\star(x)\big) \frac{G(\theta,c)}{G( \theta_{n,c}(\theta),n+c)}\,.
	\eeqan
	We then apply simple Markov inequality, which yields for all constant $c$, 
	\beqan
	\forall \delta\in(0,1],\quad
	\Pr\bigg( \cB_{\cL,\theta}^\star\bigg(\frac{n}{n\!+\!c}(\mu_n\!-\!\mu)\bigg)
	\geq \frac{1}{n+c}\ln\bigg( \frac{G( \theta_{n,c}(\theta),n+c)}{G(\theta,c)}\frac{1}{\delta}\bigg) \bigg) \leq \delta.
	\eeqan
	To conclude, we use the duality property of the Bregman divergence (Lemma~\ref{lem:bregdual}), considering that $\nabla\cL$ is invertible.  Indeed, denoting $\alpha= n/(n+c)$, 
	and $\theta_n = (\nabla \cL)^{-1}(\mu_n)$, it comes
	\beqan
	\cB_{\cL,\theta}^\star\bigg( \alpha(\mu_n-\mu)\bigg) = \cB_{\cL}\big(\theta, \theta_\alpha\big)\,\qquad\text{where}\,\, \theta_\alpha=\nabla \cL^{-1}\big(\alpha \nabla \cL(\theta_n) + (1-\alpha) \nabla \cL (\theta)\big) = \theta_{n,c}(\theta)\,.
	\eeqan
	
	Last, we note that this extends from a single  $n$ to being time-uniform
	thanks to Doob's maximal inequality for nonnegative supermartingale applied to $M_n$ (also known as Vile's inequality \citet{ville1939etude}).
\end{myproof}

\subsection{Bregman concentration using Legendre function $\cL_0$}\label{sub:legendre}
%
%
In this section, we state a more general result that can handle the case of sequence $(X_t)_{t\geq 1}$ of random variables that are not independent, each having possibly different distribution from others. 

\begin{restatable}[Laplace method for non i.i.d. samples]{theorem}{thmtwo}
\label{thm:momEFsequence}
Suppose $\lbrace \cH_t \rbrace_{t=0}^{\infty}$ is a filtration such that for each $t \geq 1$, (i) $X_t$ is $\cH_t$-measurable, (ii) $F_t$ and $h_t$ are $\cH_{t-1}$-measurable and (iii) given $\cH_{t-1}$, $X_t\sim p_{\theta_0,t}$ where $p_{\theta_0,t}$ belongs to an exponential family with parameter $\theta_0$, feature function $F_t$ and base function $h_t$. Let $\cL_t$ be the log-partition function corresponding to $F_t$.
For any Legendre (i.e., strictly convex and  continuously differentiable) function $\cL_0:\Theta \to \Real$ such that $\exp(-\cL_0)$ is integrable and any $n \in \Nat$, we introduce the parameter estimate and Bregman information gain
	\begin{align*}
	\theta_{n,\cL_0} &= \Big(\sum_{t=0}^n\nabla \cL_t\Big)^{-1}\Big(\sum_{t=1}^{n} F_t(X_t)\Big)\,,\\ 
	\gamma_{n,\cL_0}&=	\log\bigg(\frac{\int_{\Theta}\exp\Big(-\cL_0(\theta')\Big)d\theta'}{\int_{\Theta}\exp\Big(-\sum_{t=0}^{n}\cB_{\cL_{t}}(\theta',\theta_{n,\cL_0})\Big)d\theta'}\bigg)\,,
	\end{align*}
	and then, for any $\delta \in (0,1]$, the set 
	\beqan
	\Theta_{n,\cL_0}(\delta) = \left\lbrace \theta: \sum_{t=0}^{n}\cB_{\cL_t} \left(\theta,\theta_{n,\cL_0}\right)-\cL_0(\theta) \leq \log\frac{1}{\delta} + \gamma_{n,\cL_0}\right\rbrace\,.
	\eeqan
	Then, the following time-uniform concentration holds:
	\begin{align*}
	    \Pr\big[\exists n\in\Nat : \theta_0 \notin \Theta_{n,\cL_0}(\delta)\big] \leq  \delta\,.
	\end{align*}
\end{restatable}
In contrast to Theorem~\ref{thm:genericmom} that involves a local regularization using the true parameter $\theta_0 \in \Theta$ of the family and a constant $c\!>\!0$, here
we make use of a global regularization, in the form of the Legendre function $\cL_0$. Concretely, the regularized parameter estimate $\theta_{n, \cL_0}$ and the Bregman information gain $\gamma_{n, \cL_0}$ do not depend on the true parameter $\theta_0$, thus making for a more explicit confidence set $\Theta_{n, \cL_0}$. The trade-off here is that the choice of regularizer is limited by the integrability assumption on $\exp(-\cL_0)$, which is critical to build an appropriate prior for the method of mixtures.

The proof of this result follows a similar line of proof as before. However the use of $\cL_0$ instead of $c$ induces a few changes that we detail below. In particular, we use a different prior to build the mixture of martingales.




\begin{myproof}{of Theorem~\ref{thm:momEFsequence}}	
	
{\bf Step 1. Martingale and mixture martingale construction.}
	For any $\lambda \in \Real^d$, we define
	\begin{align*}
	M_n^\lambda = \exp \left( \sum_{t=1}^{n}\left(\lambda^\top\left(F_{t}(X_t)-\Esp_\theta\left[F_t(X_t)\right]\right)-B_{\cL_{t},\theta}(\lambda)\right)\right)~,
	\end{align*}
	where we introduced  $B_{\cL_{t},\theta}(\lambda)= \cL_{t}(\theta+\lambda)-\cL_{t}(\theta) - \langle \lambda, \nabla  \cL_{t} (\theta)\rangle$
	for convenience.
	
	Note that $M_n^\lambda > 0$ and
	\begin{align*}
	\Esp_\theta\left[\exp\left(\lambda^\top F_{t}(X_t)\right)\right] &= \exp\left(\cL_{t}(\theta+\lambda)-\cL_{t}(\theta)\right)~,\\
	\Esp_\theta\left[\exp\left(\lambda^\top\Esp_\theta[F_{t}(X_t)\right)\right] &= \exp\left(\lambda^\top \cL_{t}(\theta)\right)~.
	\end{align*}
	Note that $M_n^\lambda$ is $\cH_n$-measurable and in fact $\Esp[M_n^\lambda | \cH_{n-1}] = M_{n-1}^{\lambda}$. Therefore $\lbrace M_n^\lambda \rbrace_{n=0}^{\infty}$ is a nonnegative martingale adapted to the filtration $\lbrace \cH_n \rbrace_{n=0}^{\infty}$ and actually satisfies $\Esp\left[M_n^\lambda\right]=1~$.
	For any prior density $q(\theta)$ for $\theta$, we now define a mixture of martingales
	\begin{align}
	M_{n} = \int_{\Lambda_\theta}M_n^\lambda q(\theta+\lambda) d\lambda~.
	\end{align}
	where $\Lambda_\theta = \{ \lambda : \theta+\lambda \in\Theta\}$. 
	Then $M_{n}$ is also a martingale and actually satisfies $\Esp\left[M_{n}\right]=1~$.
	Now consider the prior density 
	\begin{align}
	q(\theta+\lambda) = c_0\cdot\exp\left(-\cL_0(\theta+\lambda)\right)~,
	\end{align}
	where $c_0=\frac{1}{\int_{\Theta}\exp\left(-\cL_0(\theta')\right)d\theta'}$ (which is well-defined since $\exp(-\cL_0)$ is integrable over $\Theta$). We then have
	\begin{align*}
	M_{n} = c_0\int_{\Lambda_\theta}\exp\left(\lambda^\top S_n-\sum_{t=1}^{n}\cB_{\cL_{t},\theta}(\lambda)-\cL_0(\theta+\lambda)\right)d\lambda~,
	\end{align*}
	where we denote $S_n=\sum_{t=1}^{n}\left(F_{t}(X_t)-\Esp_\theta\left[F_{t}(X_t)\right]\right)$. Now, from the formula of parameter estimate, we have
	\begin{align}
	\quad\sum_{t=1}^{n}\nabla \cL_{t}(\theta_{n,\cL_0}) +\nabla\cL_0(\theta_{n,\cL_0}) = \sum_{t=1}^{n} F_{t}(X_t)~.
	\end{align}
	This yields
	$S_n = \sum_{t=1}^{n}\left(\nabla \cL_{t}(\theta_{n,\cL_0})-\nabla \cL_{t}(\theta)\right)+\nabla \cL_0(\theta_{n,\cL_0})$. 
	
	{\bf Step 2. Legendre function and Bregman properties.}
	We now introduce the function $\cL_{1:n}(\theta)=\sum_{t=1}^{n}\cL_{t}(\theta)$. Note that $\cL_{1:n}$ is a also Legendre function and its associated Bregman divergence satisfies
	\begin{align*}
	\cB_{\cL_{1:n}}(\theta',\theta)=\sum_{t=1}^{n}\left(\cL_{t}(\theta')-\cL_{t}(\theta) - (\theta'-\theta)^\top \nabla \cL_{t}(\theta)\right)=\sum_{t=1}^{n} \cB_{\cL_{t}}(\theta',\theta). 
	\end{align*} 
	In this notation, we can rewrite $\sum_{t=1}^{n}\cB_{\cL_t,\theta}(\lambda)=\cB_{\cL_{1:n},\theta}(\lambda)$ and $S_n=\nabla \cL_n(\theta_{n,\cL_0})-\nabla \cL_n(\theta)+\nabla \cL_0(\theta_{n,\cL_0})$. We then obtain
	\begin{align}
	M_{n} = c_0\cdot\exp\left(-\cL_0(\theta)\right)\int_{\Lambda_\theta}\exp\left(\lambda^\top x_n -\cB_{\cL_{1:n},\theta}(\lambda)+\lambda^\top x_0 -\cB_{\cL_0,\theta}(\lambda)\right)d\lambda~,
	\label{eqn:martingale-first}
	\end{align} 
	where we have introduced $x_0=\nabla \cL_0(\theta_{n,\cL_0})-\nabla \cL_0(\theta)$ and $x_n=\nabla \cL_{1:n}(\theta_{n,\cL_0})-\nabla \cL_{1:n}(\theta)$.

	We now have from the Bregman-duality property recalled in Lemma~\ref{lem:bregdual} that
	\begin{align*}
	\sup_{\lambda\in \Real^d}\left(\lambda^\top x_n-\cB_{\cL_{1:n},\theta}(\lambda)\right)&=\cB_{\cL_{1:n},\theta}^\star(x_n)
	\\&=\cB_{\cL_{1:n},\theta}^\star(\nabla \cL_{1:n}(\theta_{n,\cL_0})-\nabla \cL_{1:n}(\theta))= \cB_{\cL_{1:n}}(\theta,\theta_{n,\cL_0})~.
	\end{align*}
	Further, any optimal $\lambda$ must satisfy
	\begin{align*}
	\nabla \cL_{1:n}(\theta+\lambda) - \nabla \cL_{1:n}(\theta) = x_n \implies \nabla \cL_{1:n}(\theta+\lambda) = \nabla \cL_{1:n}(\theta_{n,\cL_0})~.
	\end{align*}
	One possible solution is $\lambda^\star=\theta_{n,\cL_0}-\theta$. We then have
	\begin{align}
	& \lambda^\top x_n-\cB_{\cL_{1:n},\theta}(\lambda)\nonumber\\
	&=\lambda^\top x_n-\cB_{\cL_{1:n},\theta}(\lambda)+\cB_{\cL_{1:n}}(\theta,\theta_{n,\cL_0})-\left({\lambda^\star}^\top x_n-\cB_{\cL_{1:n},\theta}(\lambda^\star)\right)\nonumber\\
	&= \cB_{\cL_{1:n}}(\theta,\theta_{n,\cL_0})+(\lambda-\lambda^\star)^\top \nabla \cL_n(\theta+\lambda^\star)+\cB_{\cL_{1:n},\theta}(\lambda^\star)-\cB_{\cL_{1:n},\theta}(\lambda)-(\lambda-\lambda^\star)^\top \nabla \cL_{1:n}(\theta)\nonumber\\
	&= \cB_{\cL_{1:n}}(\theta,\theta_{n,\cL_0})+(\lambda-\lambda^\star)^\top \nabla \cL_{1:n}(\theta+\lambda^\star) + \cL_{1:n}(\theta+\lambda^\star)-\cL_{1:n}(\theta+\lambda)~.
	\label{eqn:first-term}
	\end{align}
	Similarly, we have
	\begin{align}
	&\lambda^\top x_0-\cB_{\cL_0,\theta}(\lambda) = \cB_{\cL_0}(\theta,\theta_{n,\cL_0})+(\lambda-\lambda^\star)^\top \nabla \cL_0(\theta+\lambda^\star)+\cL_0(\theta+\lambda^\star)-\cL_0(\theta+\lambda)\,.
	\label{eqn:second-term}
	\end{align}

	{\bf Step 3. Martingale rewriting and conclusion.}
	Plugin-in (\ref{eqn:second-term}) and (\ref{eqn:first-term}) in (\ref{eqn:martingale-first}), we now obtain
	\begin{align*}
	M_{n} 
	& = c_0\exp\Big(\sum_{j \in \lbrace 0,1:n\rbrace}\cB_{\cL_j}(\theta,\theta_{n,\cL_0})-\cL_0(\theta)\Big)\\
	&\quad\quad\times\int_{\Lambda_\theta}\exp\Big(\sum_{j \in \lbrace 0,1:n\rbrace}\left((\lambda-\lambda^\star)^\top \nabla \cL_j(\theta+\lambda^\star) + \cL_j(\theta+\lambda^\star)-\cL_j(\theta+\lambda)\right)\Big)d\lambda\\
	& = c_0\exp\Big(\sum_{j \in \lbrace 0,1:n\rbrace}\cB_{\cL_j}(\theta,\theta_{n,\cL_0})-\cL_0(\theta)\Big)\cdot \exp\Big(-\sum_{j \in \lbrace 0,1:n\rbrace}\left((\theta+\lambda^\star)^\top \nabla \cL_j(\theta+\lambda^\star) -\cL_j(\theta+\lambda^\star)\right)\Big)\\
	&\quad\quad\times\int_{\Lambda_\theta}\exp\Big(\sum_{j \in \lbrace 0,1:n\rbrace}\left((\theta+\lambda)^\top \nabla \cL_j(\theta+\lambda^\star) -\cL_j(\theta+\lambda)\right)\Big)d\lambda\\
	& = \frac{c_0}{c_n}\exp\Big(\sum_{j \in \lbrace 0,1:n\rbrace}\cB_{\cL_j}(\theta,\theta_{n,\cL_0})-\cL_0(\theta)\Big)\\
	&\quad\quad\times \frac{\int_{\Lambda_\theta}\exp\left(\sum_{j \in \lbrace 0,1:n\rbrace}\left((\theta+\lambda)^\top \nabla \cL_j(\theta+\lambda^\star) -\cL_j(\theta+\lambda)\right)\right)d\lambda}{\int_{\Theta}\exp\left(\sum_{j \in \lbrace 0,1:n\rbrace}\left((\theta')^\top \nabla \cL_j(\theta+\lambda^\star) -\cL_j(\theta')\right)\right)d\theta'}\\
	& = \frac{c_0}{c_n}\exp\Big(\sum_{t=1}^{n}\cB_{\cL_{t}}(\theta,\theta_{n,\cL_0})+\cB_{\cL_{0}}(\theta,\theta_{n,\cL_0})-\cL_0(\theta)\Big)~,
	\end{align*}
	where we have introduced
	$c_n=\frac{\exp\left(\sum_{j \in \lbrace 0,1:n\rbrace}\left((\theta+\lambda^\star)^\top \nabla \cL_j(\theta+\lambda^\star) -\cL_j(\theta+\lambda^\star)\right)\right)}{\int_{\Theta}\exp\left(\sum_{j \in \lbrace 0,1:n\rbrace}\left((\theta')^\top \nabla \cL_j(\theta+\lambda^\star) -\cL_j(\theta')\right)\right)d\theta'}~.$
	
	We then obtain the following result by a simple application of Markov's inequality, which yields
	\begin{align}
	&\Pr\left[\sum_{t=1}^{n}\cB_{\cL_{t}}(\theta,\theta_{n,\cL_0})+\cB_{\cL_0}(\theta,\theta_{n,\cL_0})-\cL_0(\theta)\geq \log \left(\frac{c_n}{c_0}\frac{1}{\delta}\right)\right]\nonumber\\ &= \Pr\left[M_{n}\geq \frac{1}{\delta}\right] \leq\delta\cdot\Esp\left[M_{n}\right] = \delta~.
	\end{align}
	Finally, since $\lambda^\star=\theta_{n,\cL_0}-\theta$, we have the more explicit form
	\begin{align*}
	\frac{c_n}{c_0}&=\frac{\int_{\Theta}\exp\Big(-\cL_0(\theta')\Big)d\theta'}{\int_{\Theta}\exp\Big(-\sum\limits_{j \in \lbrace 0,1:n\rbrace}\cB_{\cL_j}(\theta',\theta+\lambda^\star)\Big)d\theta'}\\
	&=\frac{\int_{\Theta}\exp\Big(-\cL_0(\theta')\Big)d\theta'}{\int_{\Theta}\exp\Big(-\sum_{t=1}^{n}\cB_{\cL_{t}}(\theta',\theta_{n,\cL_0})-\cB_{\cL_0}(\theta',\theta_{n,\cL_0})\Big)d\theta'}~.
	\end{align*}
	We obtain the time-uniform bound by applying Doob's maximal inequality for nonnegative supermartingale.
\end{myproof}

The general result of Theorem~\ref{thm:momEFsequence} specifies straightforwardly to the case when all observations have same law, yielding the following corollary that we state now for completeness. 

\begin{corollary}\label{cor:momEF}
	Let	$\mu_n = \frac{1}{n}\sum_{t=1}^{n} F(X_t)$.
	For all Legendre function $\cL_0$ such that $\exp(-\cL_0)$ is integrable over $\Theta$, let  
	\beqan
	\theta_{n,\cL_0}\! = \big(\nabla \cL\!+ \!\frac{1}{n}\nabla \cL_0\big)^{-1}(\mu_n)\,\text{ and }\, \gamma_{n,\cL_0}\!= \log \left(\frac{\int_{\Theta}\exp\Big(- \cL_{0}(\theta')\Big)d\theta'}{\int_{\Theta}\exp\Big(-\!n\cB_\cL \left(\theta',\theta_{n,\cL_0}\right)\!- \!
		\cB_{\cL_0}\left(\theta',\theta_{n,\cL_0}\right) \Big)d\theta'}\right)\,.
	\eeqan
	Then for all $\delta\in(0,1],$
	\beqan
	\Pr\bigg(\exists n\in\Nat,\,\, n\cB_\cL \left(\theta,\theta_{n,\cL_0}\right) + \cB_{\cL_0}\left(\theta,\theta_{n,\cL_0}\right)\geq \log\frac{1}{\delta} + \cL_0(\theta)+\gamma_{n,\cL_0}\bigg) \leq \delta\,.
	\eeqan
\end{corollary}

\paragraph{Choice of Legendre function $\cL_0$} A natural choice for the Legendre regularizer $\cL_0$ is to use the log-partition function of the exponential family at hand. If $\cL_0=c\cL$, where $c>0$ is a scaling coefficient, the formula above simplify to
\beqan
	\theta_{n,\cL_0}\! = \frac{\theta_{n}}{1+\frac{c}{n}}\,\text{ and }\, \gamma_{n,\cL_0}\!= \log \left(\frac{\int_{\Theta}\exp\Big(- c\cL(\theta')\Big)d\theta'}{\int_{\Theta}\exp\Big(-\!(n+c)\cB_\cL \left(\theta',\theta_{n,\cL_0}\right)\Big)d\theta'}\right)\,,
\eeqan
where $\theta_{n}=\left( \nabla \cL \right)^{-1}(\mu_n)$ is the standard maximum likelihood estimate of the parameter $\theta_0$.

In the special case of the one-dimensional Gaussian family $\cN(\theta_0, \sigma^2)$ where the variance $\sigma^2$ is known, the log-partition function regularizer defined by $\cL_0(\theta)=\frac{\theta^2}{2\sigma^2}$ satisfies the integrability assumption $\int_{\bR} e^{-c\cL_0(\theta)}d\theta < \infty$ with $c>0$. Straightforward calculations show that the resulting confidence set is the same as the one derived from Theorem~\ref{thm:genericmom} with local regularization $c$ (see Appendix~\ref{app:gaussian_known_variance}). For many other standard families however, this integrability assumption may fail, as in the case of the exponential distributions $\cE(-\theta)$, for which $\cL(\theta)=-\log(-\theta)$ (see Appendix~\ref{app:exponential}).

For other choices of Legendre function, computing the regularized parameter estimate $\theta_{n, \cL_0}$ and the information gain $\gamma_{n, \cL_0}$ requires inverting the function $\nabla \cL + \frac{1}{n}\nabla \cL_0$ and computing a tedious integral, both of which seldom result in closed-form expressions. For these reasons, we recommend as a first step the use of local regularization (Theorem~\ref{thm:genericmom}) over Legendre regularizers. In Appendix~\ref{app:bandits}, we detail an application of Theorem~\ref{thm:momEFsequence} to Gaussian contextual bandits using a quadratic regularizer.

\section{Properties of Bregman Divergences}\label{app:technical}
In this section, we detail the derivation of some useful properties of the Bregman divergence that are used in the proof of the Bregman deviation inequalities.

\lemone*

\begin{myproof}{of Lemma~\ref{lem:bregdual}}

	The first equality is immediate, since $\Esp_{\theta}[F(X)] = \nabla {\cL}(\theta)$ and
	\beqan
	\log \Esp_{\theta}\left[ \exp(\langle \lambda, F(X)\rangle )\right]&=&
	\log \int\exp(\langle \lambda, F(x)\rangle + \langle{\theta}, F(x)\rangle - {\cL}(\theta))dx\\
	&=& {\cL}(\lambda+\theta) - {\cL}(\theta)\,.
	\eeqan
	We now turn to the duality formula. Using the definition of each terms, we have
	\beqan
	\cB_{{\cL},\theta'}^\star(\nabla{\cL}(\theta)-\nabla{\cL}(\theta')) &=& \sup_\lambda \langle \lambda, \nabla{\cL}(\theta)-\nabla{\cL}(\theta') \rangle - \Big[{\cL}(\theta'+\lambda) - {\cL}(\theta') - \langle \lambda, \nabla {\cL}(\theta')\rangle \Big]\\
	&=&\sup_\lambda \langle \lambda, \nabla{\cL}(\theta)\rangle - {\cL}(\theta'+\lambda) + {\cL}(\theta')\,.
	\eeqan
	An optimal $\lambda$ must satisfy $\nabla {\cL}(\theta)=\nabla {\cL}(\theta'+\lambda)$. Hence, provided that $\nabla {\cL}$ is invertible, this means $\lambda = \theta-\theta'$. Plugin-in this value, we obtain
	\beqan
	\cB_{{\cL},\theta'}^\star(\nabla{\cL}(\theta)-\nabla{\cL}(\theta')) =\langle \theta-\theta', \nabla{\cL}(\theta)\rangle- {\cL}(\theta)+{\cL}(\theta')
	= \cB_{\cL}(\theta',\theta)\,.
	\eeqan
	The remaining equality $\cB_{\cL}(\theta',\theta)=\cB_{{\cL}^\star}(\nabla{\cL}(\theta),\nabla{\cL}(\theta'))$ is a standard result.
	Finally, regarding the generalization, we note that 	
	\beqan
	\cB_{{\cL},\theta'}^\star\Big(\alpha(\nabla{\cL}(\theta)-\nabla{\cL}(\theta'))\Big)
	&=&\sup_\lambda \langle \lambda, \alpha\nabla{\cL}(\theta)+(1-\alpha)\nabla \cL(\theta')\rangle - {\cL}(\theta'+\lambda) + {\cL}(\theta')\,.
	\eeqan
	Hence, an optimal $\lambda$ must now satisfy $\alpha\nabla{\cL}(\theta)+(1-\alpha)\nabla \cL(\theta')=\nabla {\cL}(\theta'+\lambda)$, that is $\lambda =  \theta_\alpha-\theta'$. This further yields
	\beqan
	\cB_{{\cL},\theta'}^\star\Big(\alpha(\nabla{\cL}(\theta)-\nabla{\cL}(\theta'))\Big)
	&=&\langle \theta_\alpha-\theta', \nabla {\cL}(\theta_\alpha)\rangle - {\cL}(\theta_\alpha) + {\cL}(\theta')= \cB_{\cL}(\theta',\theta_\alpha)\,.
	\eeqan
\end{myproof}

\section{Specification to Illustrative Exponential Families}\label{app:specific}

In this section, we provide the technical derivations to specify our generic concentration result to a few classical distributions. The results are summarized in Table~\ref{table:bregman_cs_full}.

\subsection{Gaussian with unknown mean, known variance}\label{app:gaussian_known_variance}
Consider $X \sim \cN(\mu,\sigma^2)$, where $\mu$ is unknown and $\sigma$ is known. This corresponds to an exponential family of distributions, with parameter $\theta=\mu$, feature function $F(x)=\frac{1}{\sigma^2}x$ and log-partition function $\cL(\theta)=\frac{\theta^2}{2\sigma^2}$. The Bregman divergence between two parameters $\theta'$ and $\theta$ associated with $\cL$ is given by $\cB_{\cL}(\theta',\theta)=\texttt{KL}(P_\mu,P_{\mu'})=\frac{(\mu-\mu')^2}{2\sigma^2}$.
We further have $\cL'(\theta)=\frac{\theta}{\sigma^2}=\frac{\mu}{\sigma^2}$ and $\cL''(\theta)=\frac{1}{\sigma^2}$, implying that $\cL'$ is invertible. Denoting $S_n= \sum_{t=1}^{n}X_t$, we obtain that
\begin{align*}
\mu_{n,c}(\mu) =\theta_{n,c}(\theta)=\frac{S_n+c\mu}{n+c}~.
\end{align*}
The Bregman deviations simplify as follows
\begin{align*}
(n+c)\cdot\cB_{\cL}(\theta,\theta_{n,c}(\theta))&=(n+c)\frac{(\theta_{n,c}(\theta)-\theta)^2}{2\sigma^2}= \frac{1}{n+c}\frac{\left( S_n -n\mu\right)^2}{2\sigma^2}~.
\end{align*}
Now, on the other hand, let us see that the information gain is explicitly given by
\begin{align*}
\gamma_{n,c}(\mu) &=\frac{1}{2}\log \frac{n+c}{c}~.
\end{align*}
We obtain from Theorem~\ref{thm:genericmom}
w.p. $\geq 1-\delta$, $\forall n \in \Nat$,
\begin{align*}
\frac{1}{n+c}\frac{(S_n -n\mu)^2}{2\sigma^2} \leq  \log(1/\delta)+ \frac{1}{2}\log \frac{n+c}{c}~.
\end{align*}

\subsection{Gaussian with known mean, unknown variance}
We consider $X\sim \cN\left(\mu, \sigma^2\right)$, where $\mu$ is known and $\sigma$ is unknown. This corresponds to a one-dimensional exponential family model with parameter $\theta=-\frac{1}{2\sigma^2}\in\bR^*_{-}$, feature function $F(x)=\left(x-\mu\right)^2$ and log-partition function $\cL(\theta) = -\frac{1}{2}\log\left( -2\theta \right)$. The first two derivatives of $\cL$ are given by $\cL'(\theta) = -\frac{1}{2\theta}$ and $\cL''(\theta) = \frac{1}{2\theta^2}$, which shows that $\cL'$ is invertible on the domain $\Theta = \bR^*_-$. The Bregman divergence between two parameters $\theta'$ and $\theta$ is therefore $\cB_{\cL}(\theta',\theta) = \frac{1}{2}\log\left( \theta / \theta' \right) + \frac{1}{2}\left( \theta' / \theta - 1 \right)$. 

Let $Q_n = \sum_{t=1}^n \left( X_t - \mu\right)^2$. A short calculation shows that the following expression holds:
\begin{align*} 
\theta_{n, c}(\theta) &= -\frac{n+c}{c / \theta - 2 Q_n}\,.
\end{align*}
To compute the Bregman information gain, note that the above expression and a change of variable implies 
\begin{align*}
\int_{-\infty}^0 \exp\left( -c\cB_{\cL}\left(\theta', \theta \right) \right) d\theta' &= -\theta \left( 2/c\right)^{c/2 + 1} e^{c/2} \Gamma\left( c/2 + 2\right) \,.
\end{align*}
Combining the last two lines yields
\begin{align*}
    \gamma_{n, c}(\theta) &= \log\frac{\theta}{\theta_{n, c}(\theta)} - (1\!+\!\log 2)\frac{n}{2} - (\frac{c}{2}\!+\!1)\log c + (\frac{n\!+\!c}{2}\!+\!1)\log(n\!+\!c) - \frac{\log \Gamma\left( \frac{n\!+\!c}{2}\!+\!2\right)}{\log\Gamma\left(\frac{c}{2}\!+\!2\right)}\\
    &= \log\frac{c\!-\!2\theta Q_n}{n\!+\!c} - (1\!+\!\log 2)\frac{n}{2} - (\frac{c}{2}\!+\!1)\log c + (\frac{n\!+\!c}{2}\!+\!1)\log(n\!+\!c) - \frac{\log \Gamma\left( \frac{n\!+\!c}{2}\!+\!2\right)}{\log\Gamma\left(\frac{c}{2}\!+\!2\right)}\\
    & = \log\frac{Q_n / \sigma^2 + c}{n\!+\!c} - (1\!+\!\log 2)\frac{n}{2} - (\frac{c}{2}\!+\!1)\log c + (\frac{n\!+\!c}{2}\!+\!1)\log(n\!+\!c) - \frac{\log \Gamma\left( \frac{n\!+\!c}{2}\!+\!2\right)}{\log\Gamma\left(\frac{c}{2}\!+\!2\right)}\,.
\end{align*}
Moreover, we deduce from the expression of $\theta_{n,c}(\theta)$ and the Bregman divergence that:
\begin{align*} 
\cB_{\cL}(\theta', \theta_{n,c}(\theta)) &= -\frac{1}{2}\log\left( \frac{c\theta'/\theta - 2\theta' Q_n}{n\!+\!c} \right) + \frac{1}{2}\left( \frac{c\theta'/\theta - 2\theta' Q_n}{n\!+\!c} - 1 \right) \,.
\end{align*}
So, we have
\begin{align*} 
(n+c)\cB_{\cL}(\theta, \theta_{n,c}(\theta)) &= -\frac{n+c}{2}\log\left(\frac{Q_n/\sigma^2 + c}{n\!+\!c} \right) + \frac{Q_n/\sigma^2 + c}{2} - \frac{n}{2}\,.
\end{align*}
After some simple algebra and using the natural parametrization $\theta=-\frac{1}{2\sigma^2}$, we derive from Theorem~\ref{thm:genericmom} that w.p. $\geq 1-\delta$, $\forall n \in \Nat$,
\begin{align}\label{eq:gaussianvar_confseq}
& -(\frac{n\!+\!c}{2}\!+\!1)\log \frac{Q_n/\sigma^2+c}{n\!+\!c} + \frac{Q_n}{2\sigma^2} \nonumber \\
&\leq \log \frac{1}{\delta} -\frac{n}{2}\log 2 - (\frac{c}{2}\!+\!1)\log c + (\frac{n\!+\!c}{2}\!+\!1)\log(n\!+\!c) + \log \Gamma\left( \frac{c}{2}\!+\!2\right) - \log \Gamma\left( \frac{n\!+\!c}{2}\!+\!2\right)\,.
\end{align}



\subsection{Gaussian with unknown mean and variance}

We consider $X\sim \cN\left(\mu, \sigma^2\right)$, where both $\mu\in\bR$ and $\sigma>0$ are unknown. These distributions form a two-dimensional exponential family with parameter $\theta=\begin{pmatrix}
\theta_1 \\
\theta_2
\end{pmatrix}=\begin{pmatrix}
\frac{\mu}{\sigma^2} \\
-\frac{1}{2\sigma^2}
\end{pmatrix}$ belonging to the domain $\Theta = \bR \times \bR^*_{-}$. The corresponding feature and log-partition functions writes $F(x)=\begin{pmatrix}
x \\
x^2
\end{pmatrix}$ and $\cL(\theta) = -\frac{1}{2}\log(-\theta_2) - \frac{\theta_1^2}{4\theta_2}$ respectively. The gradient and Hessian of $\cL$ follows from straightforward calculations and write:
\begin{align*}
    \nabla \cL(\theta) = \begin{pmatrix}
        -\frac{\theta_1}{2\theta_2} \\
        -\frac{1}{2\theta_2} + \frac{\theta_1^2}{4\theta_2^2}
        \end{pmatrix}\,\,,
    \nabla^2 \cL(\theta) = \begin{pmatrix}
        -\frac{1}{2\theta_2} & \frac{\theta_1}{2\theta_2^2} \\
        \frac{\theta_1}{2\theta_2^2} & \frac{1}{2\theta_2^2} - \frac{\theta_1^2}{2\theta_2^3}
        \end{pmatrix}\,.
\end{align*}
In particular, $\text{Tr }\nabla^2\cL(\theta)=-\frac{1}{2\theta_2} + \frac{1}{2\theta_2^2} - \frac{\theta_1^2}{2\theta_2^3} > 0$ and $\det \nabla^2 \cL(\theta) = -\frac{1}{4\theta_2^3} > 0$, therefore the symmetric matrix $\nabla^2 \cL(\theta)$ is positive definite (both its eigenvalues are positive). The inverse of the gradient mapping $\nabla \cL\colon \Theta \rightarrow \cP $, where the image set is $\cP = \left\lbrace (x, y)\in\bR^2, y > x^2\right\rbrace$, can be explicitly computed as:
\begin{align*}
    \left(\nabla\right)^{-1} \cL \begin{pmatrix} x \\ y\end{pmatrix} = \frac{1}{y-x^2}\begin{pmatrix}
        x\\
        -\frac{1}{2}
        \end{pmatrix}\,.
\end{align*}
The expression of the Bregman divergence between two parameters $\theta$ and $\theta'$ can be calculated from the above results and reads:
\begin{align*}
    \cB_{\cL}(\theta', \theta) &= \frac{1}{2}\log \frac{\theta_2}{\theta'_2} + \frac{\theta'_2}{2\theta_2} - \theta'_2 \left(\frac{\theta'_1}{2\theta'_2} - \frac{\theta_1}{2\theta_2}\right)^2 - \frac{1}{2}.
\end{align*}
Let $S_n=\sum_{t=1}^n X_t$ and $Q_n = \sum_{t=1}^n X_t^2$ and $V_{n,c}=\frac{1}{n+c}Q_n - \left(\frac{1}{n+c}S_n\right)^2$. We have that
\begin{align*}
    \theta_{n,c}(\theta) &= \frac{1}{V_{n, c} - \frac{c}{n\!+\!c}\frac{1}{2\theta_2} + \frac{nc}{(n\!+\!c)^2}\frac{\theta_1^2}{4\theta_2^2} + \frac{c}{(n\!+\!c)^2}\frac{\theta_1}{\theta_2}S_n}\begin{pmatrix}
        \frac{1}{n\!+\!c}S_n - \frac{c}{2(n\!+\!c)}\frac{\theta_1}{\theta_2} \\
        -\frac{1}{2}
        \end{pmatrix}\,.
\end{align*}
To compute the Bregman information gain, we will need to evaluate the integral $\int_{\Theta} \exp\left( -c \cB_{\cL}(\theta', \theta)\right)d\theta'$. As the integrand is nonnegative, we can integrate first along $\theta_1'$ (Fubini's theorem), which writes:
\begin{align*}
    \int_{\Theta} \exp\left( -c \cB_{\cL}(\theta', \theta)\right)d\theta' &= \int_{-\infty}^0 \int_{-\infty}^{\infty} \exp\left( -\frac{c}{2}\log\left(\frac{\theta_2}{\theta'_2}\right) - \frac{c}{2}\frac{\theta'_2}{\theta_2} + c\theta'_2\left( \frac{\theta'_1}{2\theta'_2} - \frac{\theta_1}{2\theta_2} \right)^2 + \frac{c}{2}\right) d\theta_1' d\theta_2'\\
    &= e^{\frac{c}{2}} \int_{-\infty}^0 \left( \frac{\theta'_2}{\theta_2} \right)^{\frac{c}{2}} e^{-\frac{c}{2}\frac{\theta'_2}{\theta_2}} \int_{-\infty}^{\infty} \exp\left(c\theta'_2\left( \frac{\theta'_1}{2\theta'_2} - \frac{\theta_1}{2\theta_2} \right)^2\right) d\theta_1' d\theta_2'\,.
\end{align*}
Let $m=\frac{\theta_1}{2\theta_2}$ and $s=\frac{1}{\sqrt{-2c\theta'_2}}$. The integral w.r.t $\theta'_1$ can be rewritten using the change of variable $y=\frac{\theta'_1}{2\theta'_2}$ as:
\begin{align*}
    \int_{-\infty}^{\infty} \exp\left(c\theta'_2\left( \frac{\theta'_1}{2\theta'_2} - \frac{\theta_1}{2\theta_2} \right)^2\right) d\theta_1' &= -2\theta_2'\sqrt{2\pi}s \underbrace{\int_{-\infty}^{\infty} \frac{1}{\sqrt{2\pi}s} e^{-\frac{\left( y-m\right)^2}{2s^2}} dy}_{=1}\,.
\end{align*}
Therefore, the above calculation simplifies to:
\begin{align*}
    \int_{\Theta} \exp\left( -c \cB_{\cL}(\theta', \theta)\right)d\theta'
    &= 2\sqrt{\frac{\pi}{c}} e^{\frac{c}{2}} \int_{-\infty}^0 \left( \frac{\theta'_2}{\theta_2} \right)^{\frac{c}{2}} e^{-\frac{c}{2}\frac{\theta'_2}{\theta_2}} \sqrt{-\theta_2'}  d\theta_2'\,,
\end{align*}
which after a linear change of variable on $\theta'_2$ can be related to the Gamma function as follows:
\begin{align*}
    \int_{\Theta} \exp\left( -c \cB_{\cL}(\theta', \theta)\right)d\theta'
    &= 2\sqrt{\frac{\pi}{c}} \left(\frac{2}{c}\right)^{\frac{c+1}{2}+1} \Gamma\left(\frac{c+1}{2}+1\right)\left(-\theta_2\right)^{\frac{3}{2}} \,.
\end{align*}
The Bregman information gain is thus:
\begin{align*}
    \gamma_{n,c}(\theta) &= -\left(\frac{1\!+\!\log 2}{2}\right)n - \left(\frac{c}{2}+2\right)\log c + \left(\frac{n\!+\!c}{2} + 2\right)\log(n\!+\!c) \\
    &\quad + \log \Gamma\left(\frac{c\!+\!3}{2}\right) - \log \Gamma\left(\frac{n\!+\!c\!+\!3}{2}\right)+ \frac{3}{2}\log\frac{\theta_2}{\theta_{n,c}(\theta)_2}\,,
\end{align*}
with $\frac{\theta_2}{\theta_{n,c}(\theta)_2} = -2\theta_2 V_{n,c} + \frac{c}{n\!+\!c} - \frac{nc}{(n\!+\!c)^2}\frac{\theta_1^2}{2\theta_2} - \frac{2c\theta_1}{(n\!+\!c)^2}S_n$. Now, applying the result of Theorem~\ref{thm:genericmom} shows that w.p. $\geq 1-\delta$, $\forall n \in \Nat$,
\begin{align*}
    & -\frac{n\!+\!c\!+\!3}{2}\log\left( \frac{\theta_2}{\theta_{n,c}(\theta)_2} \right) -\theta_2 (n\!+\!c)V_{n,c} - \frac{n\theta_1^2}{4\theta_2} - \frac{\theta_2}{n\!+\!c}S_n^2 - \theta_1 S_n \\
    &\quad \leq \log \frac{1}{\delta} - \frac{n}{2}\log 2 - \left(\frac{c}{2}\!+\!2\right)\log c + \left(\frac{n\!+\!c}{2}\!+\!2\right)\log \left(n\!+\!c\right) + \log \Gamma\left(\frac{c\!+\!3}{2}\right) - \log \Gamma\left(\frac{n\!+\!c\!+\!3}{2}\right) \,.
\end{align*}
After expanding the ratio $\frac{\theta_2}{\theta_{n,c}(\theta)_2}$ and substituting the natural parametrization in terms of $\mu$ and $\sigma$, we finally obtain that w.p. $\geq 1-\delta$, $\forall n \in \Nat$,
\begin{align*}
    & -\frac{n\!+\!c\!+\!3}{2}\log\left( \frac{1}{\sigma^2}V_{n,c} + \frac{c}{n\!+\!c} + \frac{nc}{(n\!+\!c)^2}\frac{\mu^2}{\sigma^2} - \frac{2c}{(n\!+\!c)^2}\frac{\mu}{\sigma^2}S_n\right) + \frac{n\!+\!c}{2\sigma^2}V_{n,c} + \frac{n\mu^2}{2\sigma^2} + \frac{1}{2(n\!+\!c)\sigma^2}S_n^2 - \frac{\mu}{\sigma^2} S_n \nonumber \\
    &\qquad \leq \log \frac{1}{\delta} - \frac{n}{2}\log 2 - \left(\frac{c}{2}\!+\!2\right)\log c + \left(\frac{n\!+\!c}{2}\!+\!2\right)\log \left(n\!+\!c\right)  + \log \Gamma\left(\frac{c\!+\!3}{2}\right) - \log \Gamma\left(\frac{n\!+\!c\!+\!3}{2}\right) \,.
\end{align*}
To simplify this formula, we introduce the standardized sum of squares $Z_n(m, s) = \sum\limits_{t=1}^n \left(\frac{X_t - m}{s}\right)^2$ for $(m, s)\in\bR\times\bR_+^*$. After rearranging terms and denoting $\hat \mu_n = S_n/n$, the above formula reads:
\begin{align*}
    &\frac{1}{2}Z_n(\mu, \sigma) -\frac{n+c+3}{2}\log\left(\frac{n}{n\!+\!c} Z_n(\hat \mu_n, \sigma) + \frac{c}{n\!+\!c}Z_n(\mu, \sigma) + c \right)\\
    &\leq \log\frac{1}{\delta} - \frac{n}{2}\log 2 - \left(\frac{c}{2}\!+\!2\right)\log c + \frac{1}{2}\log \left(n\!+\!c\right) + \log\Gamma\left(\frac{c\!+\!3}{2}\right) - \log\Gamma\left(\frac{n\!+\!c\!+\!3}{2}\right)\,.
\end{align*}


\subsection{Bernoulli}
We consider $X\sim\text{Bernoulli}(\mu)$. This corresponds to an exponential family model, with parameter $\theta=\log\frac{\mu}{1-\mu}$, feature function $F(x)=x$ and log-partition function $\cL(\theta)=\log(1+\exp(\theta))$. The Bregman divergence between two parameters $\theta'$ and $\theta$ associated with $\cL$ is given by $\cB_{\cL}(\theta',\theta)=\texttt{kl}(\mu,\mu')=\mu \log \frac{\mu}{\mu'}+(1-\mu)\log\frac{1-\mu}{1-\mu'}$. 
We further have $\cL'(\theta)=\frac{\exp(\theta)}{1+\exp(\theta)}=\mu$ and $\cL''(\theta)=\frac{\exp(\theta)}{(1+\exp(\theta))^2}=\mu(1-\mu)$. Therefore $\cL'$ is invertible and we have the expression $\theta_{n,c}(\theta)=(\cL')^{-1} \left( \frac{1}{n+c}\sum_{t=1}^{n}X_t+\frac{c}{n+c}\frac{\exp(\theta)}{1+\exp(\theta)}\right)$. Then, denoting $S_n=\sum_{t=1}^{n}X_t$, we get
\begin{align*}
\mu_{n,c}(\mu):=\cL'(\theta_{n,c}(\theta))=\frac{\sum_{t=1}^{n}X_t+c\mu}{n+c}= \frac{S_n+c\mu}{n+c}~.
\end{align*}
Therefore, the Bregman deviation specifies to the following closed-form formula
\begin{align*}
(n+c)\cdot\cB_{\cL}(\theta,\theta_{n,c}(\theta))&=(n+c)\cdot\texttt{kl}(\mu_{n,c}(\mu),\mu)\\
& = (S_n+c\mu)\log \frac{\mu_{n,c}(\mu)}{\mu} + \left(n-S_n+c(1-\mu)\right)\log \frac{1-\mu_{n,c}(\mu)}{1-\mu} ~.
\end{align*}
Now, we observe that
\begin{align*}
\int_{\Real^d} \exp\big(-\!c\cB_\cL(\theta',\theta)\big)d\theta' &= \frac{\text{B}(c\mu,c(1-\mu))}{\mu^{c\mu}(1-\mu)^{c(1-\mu)}}~,\\
\int_{\Real^d} \exp\big(-\!(n+c)\cB_\cL(\theta',\theta_{n,c}(\theta))\big)d\theta' &= \frac{\text{B}((n+c)\mu_{n,c}(\mu),(n+c)(1-\mu_{n,c}(\mu)))}{\mu_{n,c}(\mu)^{(n+c)\mu_{n,c}(\mu)}(1-\mu_{n,c}(\mu))^{(n+c)(1-\mu_{n,c}(\mu))}}~.
\end{align*}
Therefore, we deduce that the Bregman information gain rewrites
\begin{align*}
\gamma_{n,c}(\mu)=& (S_n + c\mu)\log \mu_{n,c}(\mu) + \left(n-S_n+c(1-\mu)\right)\log (1-\mu_{n,c}(\mu))-c\mu \log \mu - c(1-\mu)\log (1-\mu)\\&+
\log \frac{\Gamma(c\mu)\Gamma(c(1-\mu))}{\Gamma(S_n+c\mu)\Gamma(n-S_n+c(1-\mu))} + \log \frac{\Gamma(n+c)}{\Gamma(c)}~. 
\end{align*}
Combining the above and using Theorem~\ref{thm:genericmom}, we obtain that w.p. $\geq 1-\delta$, $\forall n \in \Nat$,
\begin{align*}
S_n \log \frac{1}{\mu} + (n-S_n)\log \frac{1}{1-\mu} + \log \frac{\Gamma(S_n+c\mu)\Gamma(n-S_n+c(1-\mu))}{\Gamma(c\mu)\Gamma(c(1-\mu))} \leq \log \frac{1}{\delta} + \log \frac{\Gamma(n+c)}{\Gamma(c)}~. 
\end{align*}

\subsection{Exponential}\label{app:exponential}
We consider $X\sim\text{Exp}(1/\mu)$ with unknown mean $\mu$. The distribution of $X$ is supported on $[0, +\infty)$ with density $p_{\mu}(x)=\frac{1}{\mu}e^{-\frac{x}{\mu}}$. This corresponds to an exponential family model with parameter $\theta=-\frac{1}{\mu}$, feature function $F(x)=x$ and log-partition function $\cL(\theta)=\log(-\frac{1}{\theta})$. The Bregman divergence between two parameters $\theta'$ and $\theta$ associated with $\cL$ is given by $\cB_{\cL}(\theta',\theta)=\texttt{KL}(P_\mu,P_{\mu'})=\frac{\mu}{\mu'}-1-\log\frac{\mu}{\mu'}$. 

We have $\cL'(\theta)=-\frac{1}{\theta}=\mu$ and $\cL''(\theta)=\frac{1}{\theta^2}=\mu^2$. Therefore $\cL'$ is invertible and we have 
\begin{align*}
\mu_{n,c}(\mu):=\cL'(\theta_{n,c}(\theta))=\frac{S_n+c\mu}{n+c}~.
\end{align*}
Therefore, we deduce that the Bregman divergence takes the following form
\begin{align*}
(n+c)\cdot\cB_{\cL}(\theta,\theta_{n,c}(\theta))&=(n+c)\cdot\texttt{KL}(P_{\mu_{n,c}(\mu)},P_{\mu})=\frac{S_n}{\mu}-(n+c)\log\frac{\mu_{n,c}(\mu)}{\mu}-n~.
\end{align*}
Now, we observe that
\begin{align*}
\int_{\Real^d} \exp\big(-\!c\cB_\cL(\theta',\theta)\big)d\theta' &= \frac{\Gamma (c+1)}{c\mu}\left(\frac{e}{c}\right)^c= \frac{\Gamma (c)}{\mu}\left(\frac{e}{c}\right)^c,\\
\int_{\Real^d} \exp\big(-\!(n+c)\cB_\cL(\theta',\theta_{n,c}(\theta))\big)d\theta' &= \frac{\Gamma (n+c+1)}{(n+c)\mu_{n,c}(\mu)}\left(\frac{e}{n+c}\right)^{n+c}=\frac{\Gamma (n+c)}{\mu_{n,c}(\mu)}\left(\frac{e}{n+c}\right)^{n+c}~.
\end{align*}
Therefore, the Bregman information gain writes explicitly as follows
\begin{align*}
\gamma_{n,c}(\theta) = \log \frac{\mu_{n,c}(\mu)}{\mu} + \log\left( \Gamma(c)\left(\frac{e}{c}\right)^c\right) - \log\left( \Gamma(n+c)\left(\frac{e}{n+c}\right)^{n+c}\right).
\end{align*}
We can now specify the inequality $(n+c)\cdot\cB_{\cL}(\theta,\theta_{n,c}(\theta)) \leq \log(1/\delta) + \gamma_{n,c}(\theta)$.
Combining the above, we obtain w.p. $\geq 1-\delta$, $\forall n \in \Nat$,
\begin{align*}
\frac{S_n}{\mu}-\left(n+c+1\right)\log \left(\frac{S_n+c\mu}{(n+c)\mu}\right) \leq  (n+c)\log (n+c) + \log \frac{\Gamma(c)}{\Gamma(n+c)}+\log \frac{1}{\delta}- c \log c ~.
\end{align*}


\subsection{Gamma with fixed shape}
We consider $X\sim\text{Gamma}(\lambda,k)$ with fixed shape $k >0$ and unknown scale $\lambda > 0$.\footnote{Note that $\text{Gamma}(\lambda,k)$ with $k=1$ is $\text{Exp}(1/\lambda)$.} The distribution of $X$ is supported on $[0,\infty)$ with density $p_\lambda(x)=\frac{1}{\Gamma(k)\lambda^k}x^{k-1}e^{- \frac{x}{\lambda}}$. This corresponds to an exponential family model with parameter $\theta=-\frac{1}{\lambda}$, feature function $F(x)=x$ and log-partition function $\cL(\theta)=\log(\lambda^k)=k\log(-\frac{1}{\theta})$. The Bregman divergence between two parameters $\theta'$ and $\theta$ associated with $\cL$ is given by $\cB_{\cL}(\theta',\theta)=\texttt{KL}(p_\lambda, p_{\lambda'})=k\left(\frac{\lambda}{\lambda'}-1-\log\left(\frac{\lambda}{\lambda'}\right)\right)$.

Note that $\cL'(\theta)=-\frac{k}{\theta}=k\lambda$ and $\cL''(\theta)=\frac{k}{\theta^2}=k\lambda$. Therefore $\cL'$ is invertible, and we get
\begin{align*}
k\lambda_{n,c}(\lambda) :=\cL'(\theta_{n,c}(\theta))=\frac{\sum_{t=1}^{n}X_t+ck\lambda }{n+c}
\end{align*}
Therefore, the Bregman divergence takes the following form
\begin{align*}
(n+c)\cdot\cB_{\cL}(\theta,\theta_{n,c}(\theta))&=(n+c)\cdot\texttt{KL}(p_{\lambda_{n,c}(\lambda)}, p_{\lambda})\\&=k(n+c)\left(\frac{\sum_{t=1}^{n}X_t+ck\lambda}{(n+c)k\lambda}\right)-k(n+c)\log\left(\frac{\sum_{t=1}^{n}X_t+ck\lambda}{(n+c)k\lambda}\right)-k(n+c)~.
\end{align*}
Now, we observe that
\begin{align*}
\int_{\Real^d} \exp\big(-\!c\cB_\cL(\theta',\theta)\big)d\theta' &= \frac{\Gamma (ck)}{\lambda}\left(\frac{e}{ck}\right)^{ck},\\
\int_{\Real^d} \exp\big(-\!(n+c)\cB_\cL(\theta',\theta_{n,c}(\theta))\big)d\theta' &=\frac{\Gamma ((n+c)k)}{\lambda_{n,c}(\lambda)}\left(\frac{e}{(n+c)k}\right)^{(n+c)k}~.
\end{align*}
Therefore, the Bregman information gain writes explicitly as follows
\begin{align*}
\gamma_{n,c}(\theta) = \log \left(\frac{\sum_{t=1}^{n}X_t+ck\lambda}{(n+c)k\lambda}\right) + \log\left( \Gamma(ck)\left(\frac{e}{ck}\right)^{ck}\right) - \log\left( \Gamma((n+c)k)\left(\frac{e}{(n+c)k}\right)^{(n+c)k}\right).
\end{align*}
Using the above with Theorem~\ref{thm:genericmom}, we obtain w.p. $\geq 1-\delta$, $\forall n \in \Nat$,
\begin{align}\label{eqn:confset_gamma}
    &k(n+c)\left(\frac{\sum_{t=1}^{n}X_t+ck\lambda}{(n+c)k\lambda}\right)-(k(n+c)+1)\log\left(\frac{\sum_{t=1}^{n}X_t+ck\lambda}{(n+c)k\lambda}\right)\nonumber\\&\leq \log \frac{1}{\delta}+ \log \frac{\Gamma(ck)}{\Gamma((n+c)k)}+(n+c)k\log ((n+c)k) +ck- ck \log ck.
\end{align}

\subsection{Weibull with fixed shape}

We consider $X\sim\text{Weibull}(\lambda,k)$ with fixed shape $k >0$ and unknown scale $\lambda > 0$.\footnote{Note that $\text{Weibull}(\lambda,k)$ with $k=1$ is $\text{Exp}(1/\lambda)$.} The distribution of $X$ is supported on $[0,\infty)$ with density $p_\lambda(x)=\frac{k}{\lambda}\left( \frac{x}{\lambda}\right)^{k-1}e^{-\left( \frac{x}{\lambda}\right)^{k}}$. This corresponds to an exponential family model with parameter $\theta=-\frac{1}{\lambda^k}$, feature function $F(x)=x^k$ and log-partition function $\cL(\theta)=\log(\lambda^k)=\log(-\frac{1}{\theta})$. The Bregman divergence between two parameters $\theta'$ and $\theta$ associated with $\cL$ is given by $\cB_{\cL}(\theta',\theta)=\texttt{KL}(P_\lambda,P_{\lambda'})=\left(\frac{\lambda}{\lambda'}\right)^k-1-\log\left(\frac{\lambda}{\lambda'}\right)^k$.

Note that $\cL'(\theta)=-\frac{1}{\theta}=\lambda^k$ and $\cL''(\theta)=\frac{1}{\theta^2}=\lambda^{2k}$. Therefore $\cL'$ is invertible, and we get
\begin{align*}
(\lambda_{n,c}(\lambda))^k :=\cL'(\theta_{n,c}(\theta))=\frac{\sum_{t=1}^{n}X_t^k+c\lambda^k}{n+c}.
\end{align*}
Therefore, the Bregman divergence takes the following form
\begin{align*}
(n+c)\cdot\cB_{\cL}(\theta,\theta_{n,c}(\theta))&=(n+c)\cdot\texttt{KL}(p_{\lambda_{n,c}(\lambda)}, p_{\lambda})\nonumber\\&=(n+c)\left(\frac{\sum_{t=1}^{n}X_t^k+c\lambda^k}{(n+c)\lambda^k}\right)-(n+c)\log\left(\frac{\sum_{t=1}^{n}X_t^k+c\lambda^k}{(n+c)\lambda^k}\right)-(n+c)~.
\end{align*}
Now, we observe that
\begin{align*}
\int_{\Real^d} \exp\big(-\!c\cB_\cL(\theta',\theta)\big)d\theta' &= \frac{\Gamma (c)}{\lambda^k}\left(\frac{e}{c}\right)^c,\\
\int_{\Real^d} \exp\big(-\!(n+c)\cB_\cL(\theta',\theta_{n,c}(\theta))\big)d\theta' &=\frac{\Gamma (n+c)}{(\lambda_{n,c}(\lambda))^k}\left(\frac{e}{n+c}\right)^{n+c}~.
\end{align*}
Therefore, the Bregman information gain writes explicitly as follows
\begin{align*}
\gamma_{n,c}(\theta) = \log \left(\frac{\sum_{t=1}^{n}X_t^k+c\lambda^k}{(n+c)\lambda^k}\right) + \log\left( \Gamma(c)\left(\frac{e}{c}\right)^c\right) - \log\left( \Gamma(n+c)\left(\frac{e}{n+c}\right)^{n+c}\right).
\end{align*}
Using the above with Theorem~\ref{thm:genericmom}, we obtain w.p. $\geq 1-\delta$, $\forall n \in \Nat$,
\begin{align}\label{eqn:confset_weibull}
    &(n+c)\left(\frac{\sum_{t=1}^{n}X_t^k+c\lambda^k}{(n+c)\lambda^k}\right)-(n+c+1)\log\left(\frac{\sum_{t=1}^{n}X_t^k+c\lambda^k}{(n+c)\lambda^k}\right)\nonumber\\&\leq \log \frac{1}{\delta}+ \log \frac{\Gamma(c)}{\Gamma(n+c)}+(n+c)\log (n+c) +c- c \log c.
\end{align}

\subsection{Pareto with fixed scale}
We consider $X\sim \text{Pareto}\left(\alpha\right)$, where $\alpha>0$ is unknown.\footnote{We assume scale is fixed to the value $1$.} The distribution of $X$ is supported in $[1, +\infty)$, with density $p_{\alpha}(x)=\frac{\alpha}{x^{\alpha\!+\!1}}$, which corresponds to a one-dimensional exponential family model with parameter $\theta=-\alpha-1\in(-\infty, -1)$, feature function $F(x)=\log x$ and log-partition function $\cL(\theta) = -\log\left(-1\!-\!\theta\right)$. The first two derivatives of $\cL$ are given by $\cL'(\theta)=-\frac{1}{1+\theta}$ and $\cL''(\theta) = \frac{1}{(1+\theta)^2}$, therefore $\cL'$ is invertible on the domain $(-\infty, -1)$. Using these expressions, the Bregman divergence between two parameters $\theta'$ and $\theta$ writes $\cB_{\cL}(\theta',\theta) = -\log\left(-1-\theta'\right) + \log\left(-1-\theta\right) + \frac{\theta'\!-\!\theta}{1\!+\!\theta}$.
Using the shorthand $L_n = \sum_{t=1}^n \log X_t$, it follows from the definition that:
\begin{align*} 
\theta_{n, c}(\theta) &= -1 + \frac{n\!+\!c}{\frac{c}{1\!+\!\theta} - L_n}\,.
\end{align*}
To compute the Bregman information gain, we rewrite the following integral thanks to an affine change of variable in order to relate it to the Gamma function:
\begin{align*}
\int_{-\infty}^{-1}\exp \left(-c\cB_{\cL}(\theta', \theta)\right) d\theta' &= \left(-1\!-\!\theta\right)^{-c} e^{\frac{c\theta}{1\!+\!\theta}}\int_{-\infty}^{-1} \left(-1\!-\!\theta'\right)^c e^{-\frac{c\theta'}{1\!+\!\theta}}d\theta'\nonumber \\
&= \left(-1\!-\!\theta\right) \Gamma(c) \left(\frac{e}{c}\right)^c \,.
\end{align*}
The expression of the Bregman information then follows immediately:
\begin{align*}
\gamma_{n,c}(\theta) &= - \log\left(\frac{n\!+\!c}{(-1-\theta)L_n + c}\right) - n - c\log c + (n\!+\!c)\log (n\!+\!c) + \log \Gamma(c) - \log \Gamma(n\!+\!c) \,.
\end{align*}
Moreover, we deduce from the expression of $\theta_{n,c}(\theta)$ and the Bregman divergence that:
\begin{align*}
\cB_{\cL}(\theta', \theta_{n,c}(\theta)) &= -\log(-1-\theta') +\log\left( \frac{n\!+\!c}{L_n - \frac{c}{1\!+\!\theta}}\right) - \frac{\left(L_n\!-\!\frac{c}{1\!+\!\theta}\right)\left(\theta'\!+\!1\right)}{n\!+\!c} - 1 \,.
\end{align*}
Therefore, Theorem~\ref{thm:genericmom} combined with the natural parameter $\theta=-1-\alpha$ yields that w.p. $\geq 1-\delta$, $\forall n \in \Nat$,
\begin{align}\label{eqn:confset_pareto}
    \alpha L_n-(n\!+\!c\!+\!1)\log\left( \alpha L_n+c\right)  &\leq \log \frac{1}{\delta} - c\log c - \log(n\!+\!c) + \log \Gamma(c) - \log \Gamma(n\!+\!c) \,.
\end{align}







\subsection{Chi-square}
We finally consider $X\sim\chi^2(k)$ or, equivalently, $X \sim \texttt{Gamma}\left(\frac{k}{2},\frac{1}{2}\right)$, i.e., $p_k(x)=\frac{(\frac{1}{2})^{\frac{k}{2}}}{\Gamma \left(\frac{k}{2}\right)}x^{\frac{k}{2}-1}e^{-\frac{x}{2}}$, $x \geq 0$, $k \in \Nat$ (or $k\in\Real_+\setminus\{0\}$ if one considers \texttt{Gamma} distributions). 
This corresponds to an exponential family model with parameter $\theta=\frac{k}{2}-1$, feature function $F(x)=\log x$ and log-partition function $\cL(\theta)=(\theta+1)\log 2+\log \Gamma (\theta+1)$. The Bregman divergence between two parameters $\theta'$ and $\theta$ associated with $\cL$ is given by
\begin{align*}
\cB_{\cL}(\theta',\theta)=\texttt{KL}(P_k,P_{k'})=\frac{1}{2}(k-k')\psi_0(k/2)-\log \frac{\Gamma (k/2)}{\Gamma (k'/2)}~,
\end{align*}
where $\psi_0(x)=\frac{d}{dx}\log \Gamma (x)$ denotes the digamma function.

We further have $\cL'(\theta)=\log 2 + \psi_0(\theta+1)$ and $\cL''(\theta)=\psi_1(\theta+1)$, where $\psi_1(x)=\frac{d^2}{dx^2}\log \Gamma (x)$ denotes the trigamma function. Therefore $\cL'$ is invertible, and the parameter estimate is given by
\begin{align*}
\log2 + \psi_0(\theta_{n,c}(\theta)+1)=\frac{1}{n+c}\sum_{t=1}^{n}\log X_t+\frac{c}{n+c}(\log 2 + \psi_0(\theta+1))~,
\end{align*}
yielding
\begin{align*}
k_{n,c}(k):=2(1+\theta_{n,c}(\theta)) &= 2\psi_0^{-1}\left( \frac{1}{n+c}\sum_{t=1}^{n}\log X_t+\frac{c}{n+c} \psi_0(k/2)-\frac{n}{n+c}\log 2\right)\\
&=2\psi_0^{-1}\left( \frac{1}{n+c}K_n+\frac{c}{n+c} \psi_0(k/2)\right)\,,
\end{align*}
where $K_n = \sum\limits_{t=1}^n \log \frac{X_t}{2}$.
Therefore, the Bregman divergence rewrites as follows
\begin{align*}
(n+c)\cdot\cB_\cL(\theta,\theta_{n,c}(\theta)) &= (n+c)\cdot\texttt{KL}(P_{k_{n,c}(k)},P_k)\\
&=\frac{1}{2}(k_{n,c}(k)-k)\left(K_n + c \;\psi_0\left(\frac{k}{2}\right)\right)-(n+c)\log \frac{\Gamma \left(\frac{k_{n,c}(k)}{2}\right)}{\Gamma \left(\frac{k}{2}\right)}~.
\end{align*}
Now, we see that
\begin{align*}
\int_{\Real^d} \exp\big(\!-\!c\cB_\cL(\theta',\theta)\big)d\theta' \!&=\! \int \frac{1}{2}\exp \left(\!-\frac{c}{2}(k-k')\psi_0(k/2)\right) \left(\frac{\Gamma (k/2)}{\Gamma (k'/2)}\right)^c dk'~,\\
\int_{\Real^d} \exp\big(\!-\!(n+c)\cB_\cL(\theta',\theta_{n,c}(\theta))\big)d\theta' \!&=\! \int \frac{1}{2}\exp \left(\!-\frac{n\!+\!c}{2}(k_{n,c}(k)-k')\psi_0(k_{n,c}(k)/2)\right) \left(\frac{\Gamma (k_{n,c}(k)/2)}{\Gamma (k'/2)}\right)^{n\!+\!c}\!dk'~.
\end{align*}
Therefore, the Bregman information gain writes 
\begin{align*}
\gamma_{n,c}(\theta)&=c\log \Gamma \left(\frac{k}{2}\right) - (n+c)\log \Gamma \left(\frac{k_{n,c}(k)}{2}\right)-c\;\frac{k}{2}\psi_0\left(\frac{k}{2}\right) +(n+c)\frac{k_{n,c}(k)}{2}\psi_0 \left( \frac{k_{n,c}(k)}{2}\right)\nonumber\\
& \quad+ \log \frac{\int \left( \Gamma \left(k'/2\right)\right)^{-c} \exp \left(c\; \frac{k'}{2}\psi_0\left(\frac{k}{2}\right)\right)dk'}{\int \left( \Gamma \left(k'/2\right)\right)^{-(n+c)} \exp \left((n+c) \frac{k'}{2}\psi_0\left(\frac{k_{n,c}(k)}{2}\right)\right)dk'}\nonumber\\
& = c\log \Gamma \left(\frac{k}{2}\right) - (n+c)\log \Gamma \left(\frac{k_{n,c}(k)}{2}\right)-c\;\frac{k}{2}\psi_0\left(\frac{k}{2}\right) +\frac{k_{n,c}(k)}{2}\left(K_n + c \psi_0\left(\frac{k}{2}\right) \right)\nonumber\\
& \quad+ \log \frac{\int \left( \Gamma \left(k'/2\right)\right)^{-c} \exp \left(c\; \frac{k'}{2}\psi_0\left(\frac{k}{2}\right)\right)dk'}{\int \left( \Gamma \left(k'/2\right)\right)^{-(n+c)} \exp \left(\frac{k'}{2}\left(K_n + c\psi_0\left(\frac{k}{2}\right)\right)\right)dk'} \nonumber\\
& = c\log \Gamma \left(\frac{k}{2}\right) - (n+c)\log \Gamma \left(\frac{k_{n,c}(k)}{2}\right)-c\;\frac{k}{2}\psi_0\left(\frac{k}{2}\right) +\frac{k_{n,c}(k)}{2}\left(K_n + c \psi_0\left(\frac{k}{2}\right) \right)\nonumber\\
& \quad + \log J\left(c, c\psi_0\left( \frac{k}{2}\right)\right) - \log J\left(n\!+\!c, K_n\!+\!c\psi_0\left( \frac{k}{2}\right)\right)\,,
\end{align*}
where we define the auxiliary function $J(a, b) = \int \exp\left( -a \log \Gamma\left( \frac{k'}{2}\right) + b\frac{k'}{2}\right) dk'$. Combining the above with Theorem~
\ref{thm:genericmom}, and after some simple algebra, we obtain that w.p. $\geq 1-\delta$, $\forall n \in \Nat$,
\begin{align}\label{eq:chi2_confseq}
&n \log \Gamma \left( \frac{k}{2}\right)-\frac{k}{2}K_n - \log J\left(c, c\psi_0\left( \frac{k}{2}\right)\right) + \log J\left(n\!+\!c, K_n\!+\!c\psi_0\left( \frac{k}{2}\right)\right) \leq \log \frac{1}{\delta}\,.
\end{align}


\begin{remark}\label{rmk:chi2_gamma}
The integral terms $\int dk'$ in the above derive from the martingale construction in \ref{sub:c} and the mixture distribution $q(\theta|\alpha,\beta)$ over the parameter $\theta\in\Theta$ of the exponential family. In the case of $\texttt{Gamma}\left(\frac{k}{2}, \frac{1}{2}\right)$ with unknown shape $\frac{k}{2}>0$, we have $\theta=\frac{k}{2}$ and $\Theta=\left(0, +\infty\right)$, therefore $dk'$ corresponds to the Lebesgue measure over $\left(0, +\infty\right)$. When restricted to the Chi-square family, $\Theta=\mathbb{N}$, and $dk'$ is instead the counting measure, effectively turning integrals into discrete sums. 
In Appendix~\ref{app:xps}, we report figures using both versions, see Figure~\ref{fig:confidencebands_chi2} and Figure~\ref{fig:confidencebandschi2_comparison}.
\end{remark}
\begin{remark}\label{rmk:chi2_compute}
The ratio of integrals (or infinite sums) in (\ref{eq:chi2_confseq}) can be efficiently implemented using a simple integration scheme (or by truncation). Indeed, for a given $k_{\max}\in\mathbb{N}$, $b\ll 1$ and $B\gg 1$, we define $x_{k'}= b + \frac{k'}{k_{\max}} \left(B-b\right)$ for $k'=0, \dots, k_{\max}$, so that
\begin{align*}
  & \log \frac{\int_{0}^{+\infty} \left( \Gamma \left(\frac{k'}{2}\right)\right)^{-c} \exp \left(\frac{k'}{2}c\psi_0\left(\frac{k}{2}\right)\right)dk'}{\int_{0}^{+\infty} \left( \Gamma \left(\frac{k'}{2}\right)\right)^{-(n+c)} \exp \left(\frac{k'}{2}\left(c\psi_0\left(\frac{k}{2}\right)+\sum_{t=1}^{n}\log \frac{X_t}{2}\right)\right)dk'} \\
  &\approx \underset{k'=1, \dots, k_{\max}}{\emph{logsumexp}}\left( -c \log\Gamma \left(\frac{x_{k'}}{2}\right) + \frac{c x_{k'}}{2}\psi_0\left(\frac{k}{2} \right) + \log\left(x_{k'}-x_{k'-1}\right)\right) \\
  &\quad -\underset{k'=1, \dots, k_{\max}}{\emph{logsumexp}}\left(-\left(n+c\right) \Gamma\left(\frac{x_{k'}}{2}\right) + \frac{x_{k'}}{2}\left(c\psi_0\left(\frac{k}{2}+\sum_{t=1}^{n}\log \frac{X_t}{2}\right)\right) + \log\left(x_{k'} - x_{k'-1}\right) \right).
  \end{align*}
  Similarly, we have
   \begin{align*}
   & \log \frac{\sum\limits_{k'=1}^{+\infty} \left( \Gamma \left(\frac{k'}{2}\right)\right)^{-c} \exp \left(\frac{k'}{2}c\psi_0\left(\frac{k}{2}\right)\right)}{\sum\limits_{k'=1}^{+\infty} \left( \Gamma \left(\frac{k'}{2}\right)\right)^{-(n+c)} \exp \left(\frac{k'}{2}\left(c\psi_0\left(\frac{k}{2}\right)+\sum_{t=1}^{n}\log \frac{X_t}{2}\right)\right)} \\ &\approx \underset{k'=1, \dots, k_{\max}}{\emph{logsumexp}}\left( -c \log\Gamma \left(\frac{k'}{2}\right) + \frac{k'}{2}c\psi_0\left(\frac{k}{2} \right) \right) \\
   &\quad -\underset{k'=1, \dots, k_{\max}}{\emph{logsumexp}}\left( -\left(n+c\right) \Gamma \left(\frac{k'}{2}\right) + \frac{k'}{2}\left(c\psi_0\left(\frac{k}{2}+\sum_{t=1}^{n}\log \frac{X_t}{2}\right)\right) \right),
\end{align*}
The final steps correspond to the right-rectangular scheme over $(b, B)$ with $k_{\max}$ steps and the truncation to the first $k_{\max}$ terms respectively. Note the use of $\underset{k'=1, \dots, k_{\max}}{\emph{logsumexp}}(z) = \log \sum\limits_{k'=1}^{k_{\max}} \exp\left(z_{k'}\right)$ for $z\in\mathbb{R}^{k_{\max}}$, which is efficiently implemented in many libraries for scientific computing and better handles summation of large numbers. Empirically, we found that $\log b = -10, \log B = 10$ and $k_{\max} = 2000$ provided sufficient accuracy and that using finer approximation schemes did not significantly impact the numerical results.
\end{remark}

\subsection{Poisson}
We consider $X\sim \text{Poisson}\left(\lambda\right)$, where $\lambda>0$ is unknown. We recall that the distribution of $X$ is supported on $\bN$ with probability mass function $\bP\left(X=k\right) = \frac{\lambda^k e^{-\lambda}}{k!}$. This corresponds to a one-dimensional exponential family model with parameter $\theta=\log \lambda \in\bR$, feature function $F(x)=x$ and log-partition function $\cL(\theta) = e^\theta$ (which is invertible on $\bR$). The Bregman divergence between two parameters $\theta'$ and $\theta$ is therefore $\cB_{\cL}(\theta',\theta) = e^{\theta'} - e^{\theta} - (\theta' - \theta)e^{\theta}$. 

Using the shorthand $S_n = \sum_{t=1}^n X_t$, it follows from the definition that:
\begin{align*} 
\theta_{n, c}(\theta) &= \log\left( \frac{S_n\!+\!ce^{\theta}}{n\!+\!c}\right)\,.
\end{align*}
The Bregman information gain is expressed using the auxiliary function $I(a, b) = \int_{-\infty}^{+\infty} e^{-a e^{\theta} + b\theta} d\theta$ as: 
\begin{align*}
\gamma_{n,c}(\theta) &= c(1 - \theta)e^{\theta} - (n\!+\!c)\left(1 - \theta_{n,c}(\theta)\right) e^{\theta_{n,c}(\theta)} + \log I\left(c, c e^{\theta}\right) - \log I\left(n\!+\!c, (n\!+\!c)e^{\theta_{n,c}(\theta)}\right)\nonumber \\
&= c(1 - \theta)e^{\theta} - \left(1 - \log\left( \frac{S_n\!+\!ce^{\theta}}{n\!+\!c}\right)\right) \left(S_n\!+\!ce^{\theta}\right) + \log I\left(c, c e^{\theta}\right) - \log I\left(n\!+\!c, S_n\!+\!ce^{\theta}\right)\,.
\end{align*}
Moreover, we deduce from the expression of $\theta_{n,c}(\theta)$ and the Bregman divergence that:
\begin{align*} 
\cB_{\cL}(\theta', \theta_{n,c}(\theta)) &= e^{\theta'} - \left(\frac{S_n\!+\!ce^{\theta}}{n\!+\!c}\right) - \left(\theta' - \log\left( \frac{S_n\!+\!ce^{\theta}}{n\!+\!c}\right) \right) \left( \frac{S_n\!+\!ce^{\theta}}{n\!+\!c}\right) \,.
\end{align*}
Therefore, Theorem~\ref{thm:genericmom} combined with the natural parametrization $\theta=\log \lambda$ yields that w.p. $\geq 1-\delta$, $\forall n \in \Nat$,
\begin{align*}
    n\lambda - S_n \log \lambda &\leq \log \frac{1}{\delta} + \log I\left(c, c \lambda\right) - \log I\left(n\!+\!c, S_n\!+\!c\lambda\right) \,.
\end{align*}

\begin{remark}\label{rmk:poisson_compute}
    Although, to the best of our knowledge, the integral $I(a, b)$ does not have a closed-form expression, it can be numerically estimated up to arbitrary precision. We recommend the same implementation as discussed in Remark~\ref{rmk:chi2_compute}, using the \emph{logsumexp} operator for stability, and refer to the code for further details.
\end{remark}

\subsection{Summary table}

We summarize the high probability confidence sets of Theorem~\ref{thm:genericmom} derived in this Table~\ref{table:bregman_cs_full}.


\begin{table}
  \caption{Summary of Bregman confidence sets given by Theorem~\ref{thm:genericmom} for representative families. Throughout, the following notations are used:\\
  {\footnotesize
  $S_n=\sum\limits_{t=1}^{n}X_t$,\quad $\hat\mu_n = \frac{S_n}{n}$,\quad $Q_n(\mu)=\sum\limits_{t=1}^n \left(X_t-\mu\right)^2$,\quad $Z_n(\mu, \sigma)=\frac{Q_n(\mu)}{\sigma^2}$,\\
  $S_n^{(k)}=\sum\limits_{t=1}^{n}X_t^k$,\quad $L_n=\sum\limits_{t=1}^{n}\log X_t$,\quad $K_n=\sum\limits_{t=1}^{n}\log \frac{X_t}{2}$,\\
  $I(a, b) = \int_{-\infty}^{+\infty} e^{-a e^{\theta} + b\theta} d\theta$,\\
  $J(a, b) = \int_{0}^{\infty} \exp\left( -a \log \Gamma\left( \frac{k}{2}\right) \!+\! \frac{bk}{2}\right) dk$ ($dk$ is the Lebesgue measure if $k\in\bR_+$ and the counting measure if $k\in\Nat$).\\
  }
  }
  \label{table:bregman_cs_full}
  \centering
  \begin{tabular}{lll}
    \toprule
    Name     & Parameters     & Formula \\
    \midrule
    Gaussian & $\mu\in\bR$  & $\frac{1}{n+c}\frac{\left(S_n -n\mu\right)^2}{2\sigma^2} \leq   \log\frac{1}{\delta}+ \frac{1}{2}\log \frac{n+c}{c}$ \label{eqn:confsetGaussian_full} \\[4.0ex]
    Gaussian & $\sigma\in\bR_+$ & \makecell[l]{$\frac{Q_n(\mu)}{2\sigma^2} - \left(\frac{n\!+\!c}{2}\!+\!1\right)\log \left(\frac{Q_n(\mu)+c}{2\sigma^2}\right) $ \\ $\quad\leq \log \frac{1}{\delta}\!+\!\log \frac{\Gamma\left(\frac{c}{2}\!+\!2\right)}{\Gamma\left(\frac{n\!+\!c}{2}\!+\!2\right)}\!-\!\frac{n}{2}\log 2\!-\!(\frac{c}{2}\!+\!1)\log c\!+\!(\frac{n\!+\!c}{2}\!+\!1)\log(n\!+\!c)$} \label{eqn:gauss-var_full} \\[6.0ex]
    Gaussian & \makecell[l]{$\mu\in\bR$\\ $\sigma\in\bR_+$} & 
    \makecell[l]{$\frac{1}{2}Z_n(\mu, \sigma) -\frac{n+c+3}{2}\log\left(\frac{n}{n\!+\!c} Z_n(\hat \mu_n, \sigma) + \frac{c}{n\!+\!c}Z_n(\mu, \sigma) + c \right)$ \\ $\quad\leq \log\frac{1}{\delta} - \frac{n}{2}\log 2 - \left(\frac{c}{2}\!+\!2\right)\log c + \frac{1}{2}\log \left(n\!+\!c\right)$ \\ $\qquad+ \log\Gamma\left(\frac{c\!+\!3}{2}\right) - \log\Gamma\left(\frac{n\!+\!c\!+\!3}{2}\right)$} \label{eqn:gauss-mean--var_full} \\[6.0ex]
    Bernoulli & $\mu\in[0, 1]$  & \makecell[l]{$S_n \log \!\frac{1}{\mu} \!+\! (n\!-\!S_n)\log \!\frac{1}{1\!-\!\mu} \!+\! \log \!\frac{\Gamma(S_n\!+\!c\mu)\Gamma(n\!-\!S_n\!+\!c(1\!-\!\mu))}{\Gamma(c\mu)\Gamma(c(1\!-\!\mu))}$ \\ $\quad\leq \log \!\frac{1}{\delta} \!+\! \log\! \frac{\Gamma(n\!+\!c)}{\Gamma(c)}$} \label{eqn:confsetBernoulli_full} \\[5.0ex]
    Exponential & $\mu\in\bR_+$  & \makecell[l]{$\frac{S_n}{\mu}\!-\!(n\!+\!c\!+\!1) \log\left(\frac{S_n}{\mu}+c \right)$ \\ $\quad\leq \log \frac{1}{\delta} \!+\! \log \frac{\Gamma(c)}{\Gamma(n\!+\!c)}\!-\! \log (n\!+\!c)  \!-\! c \log c$} \label{eqn:confsetExpo_full} \\[5.0ex]
    Gamma & $\lambda\in\bR_+$  & \makecell[l]{$\frac{S_n}{\lambda}\!-\!((n\!+\!c)k\!+\!1) \log\!\left(\!\frac{S_n}{\lambda}\!+ck\! \right)$ \\ $\quad\leq \log \frac{1}{\delta} \!+\! \log \frac{\Gamma(ck)}{\Gamma((n\!+\!c)k)}\!-\! \log ((n\!+\!c)k) - ck \log ck$} \label{eqn:confsetGamma_full} \\[5.0ex]
    Weibull & $\lambda\in\bR_+$  & \makecell[l]{$\frac{S_n^{(k)}}{\lambda^k} -(n+c+1) \log\left(\frac{S_n^{(k)}}{\lambda^k}+c \right)$\\ $\quad\leq \log \frac{1}{\delta}+ \log \frac{\Gamma(c)}{\Gamma(n+c)}-\log(n+c)-c\log c$} \label{eqn:confsetWeibull_full} \\[5.0ex]
    Pareto & $\alpha\in\bR$  & \makecell[l]{$\alpha L_n-(n\!+\!c\!+\!1)\log\left( \alpha L_n+c\right)$ \\ $\quad\leq \log \frac{1}{\delta}+ \log \frac{\Gamma(c)}{\Gamma(n+c)}-\log(n+c)-c\log c$} \label{eqn:confsetPareto_full} \\[5.0ex]
    Poisson & $\lambda\in\bR_+$  & \makecell[l]{$n\lambda \!-\! S_n \log \lambda$\\ $\quad\leq \log \frac{1}{\delta} \!+\! \log I\left(c, c \lambda\right) - \log I\left(n\!+\!c, S_n\!+\!c\lambda\right)$} \label{eqn:confsetPoisson_full} \\[5.0ex]
    Chi-square & \makecell[l]{$k\in\Nat$\\ or $k\in\bR_+$}  & \makecell[l]{$n \log \Gamma \left( \frac{k}{2}\right)\!-\!\frac{k}{2}K_n\nonumber \!-\! \log J\left(c, c\psi_0\left( \frac{k}{2}\right)\right)$ \\ $\quad + \log J\left(n\!+\!c, K_n\!+\!c\psi_0\left( \frac{k}{2}\right)\right)\leq \log \frac{1}{\delta}$} \label{eqn:confsetXhi2_full} \\[2.5ex]
    \bottomrule
  \end{tabular}
\end{table}

\newpage
\section{Empirical Comparison with Existing Time-uniform Confidence Sequences}
\label{app:xps}

We illustrate the time-uniform confidence sequences derived from Bregman concentration on several instances of classical exponential families detailed in \ref{sub:classicalfamilies}. In each setting, when available, we also report confidence sequences based on existing methods in the literature\footnote{The code is provided in the supplementary material for reproducibility.}.

\begin{figure}[H]
	\centering
	\includegraphics[width=0.49\linewidth]{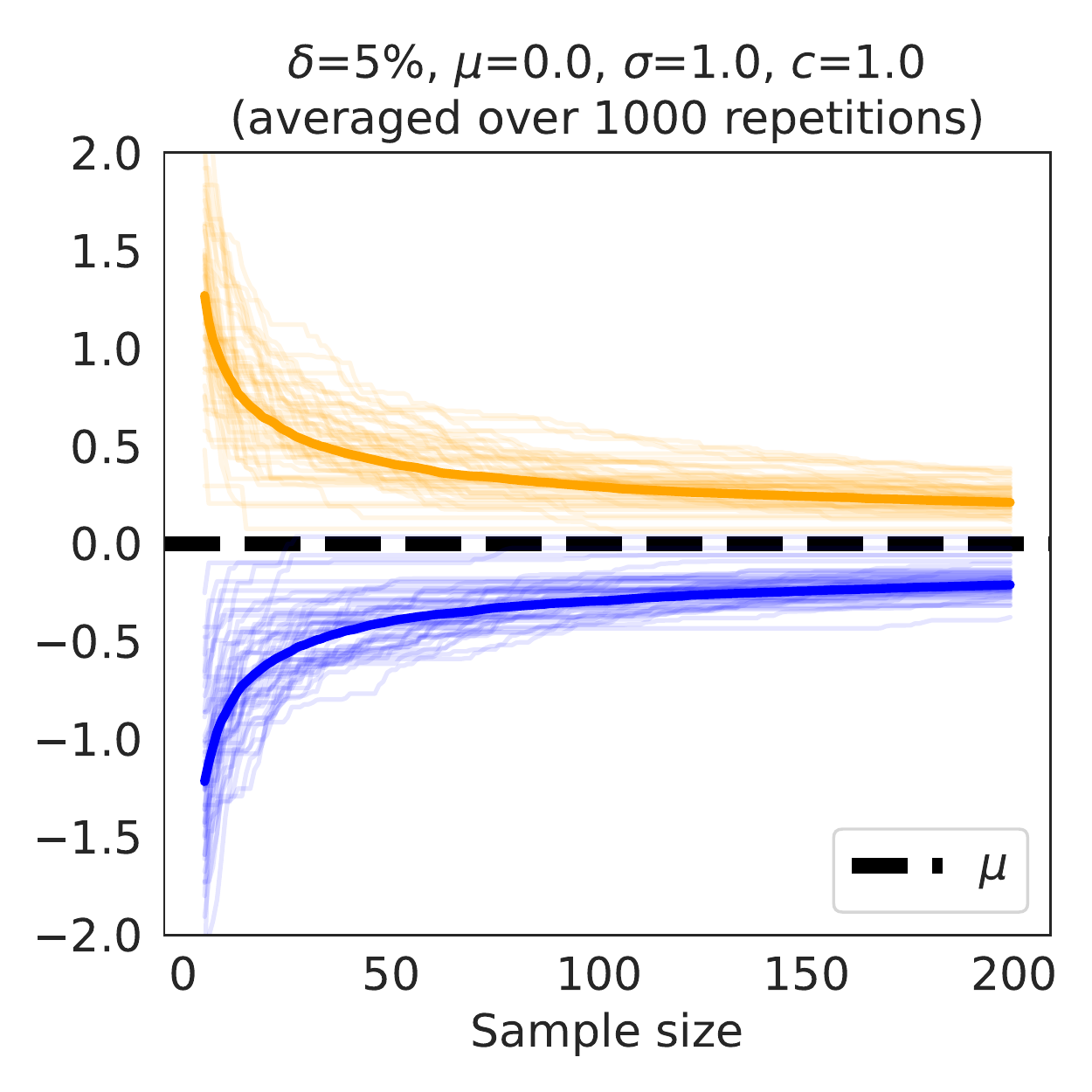}
	\includegraphics[width=0.49\linewidth]{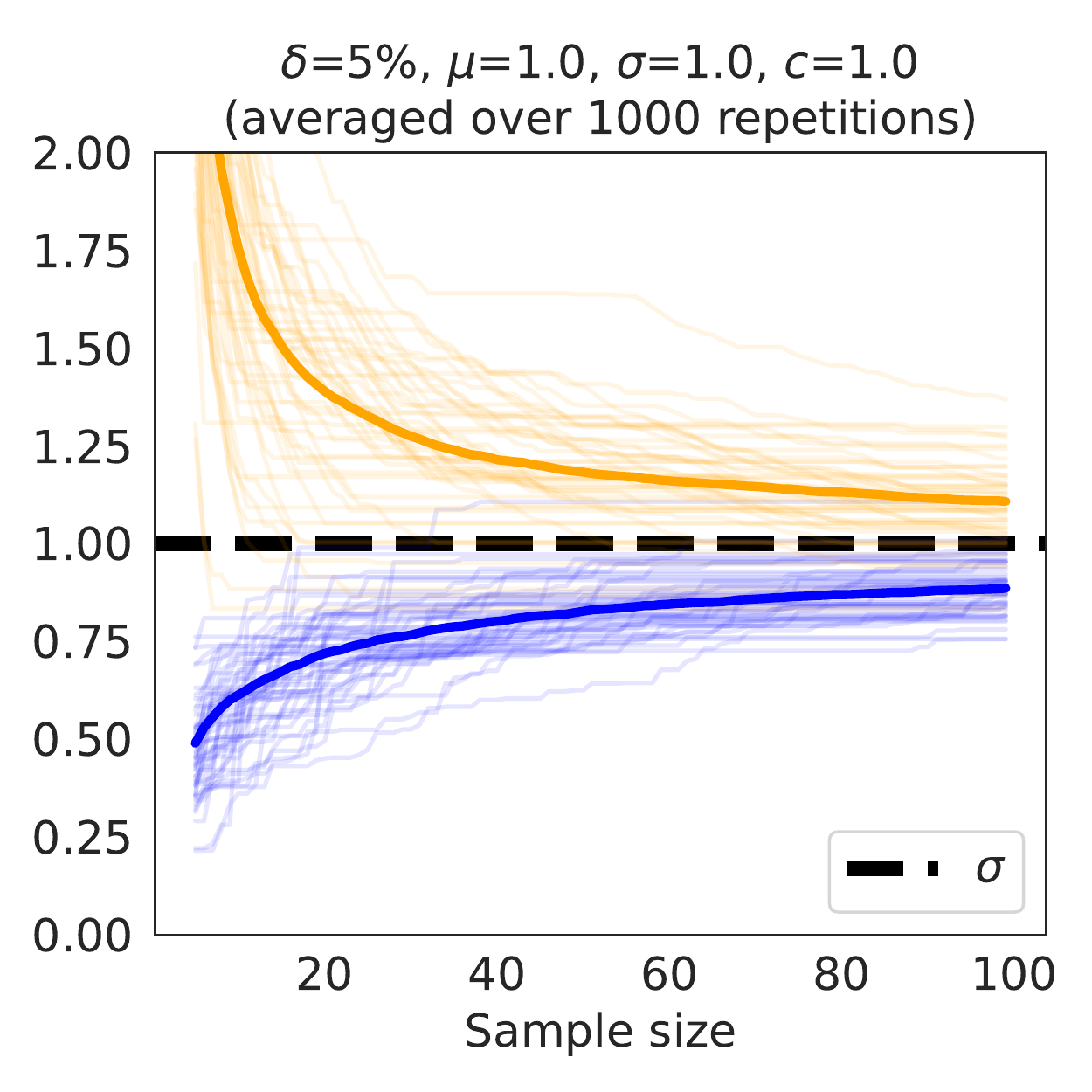}
	\caption{Examples of confidence upper and lower envelopes around unknown expectation $\mu=0$ for $\cN\left(\mu, 1\right)$ (left, cf. Table~\ref{eqn:confsetGaussian_full}) and unknown standard deviation $\sigma=1$ for $\cN\left(1, \sigma\right)$ (right, cf. Table~\ref{eqn:gauss-var_full}), as a function of the number of observations $n$. The thick lines indicate the median curve over 1000 replicates.}
	\label{fig:confidencebands_gaussian}
\end{figure}

\begin{figure}[H]
	\centering
    \includegraphics[width=0.49\linewidth]{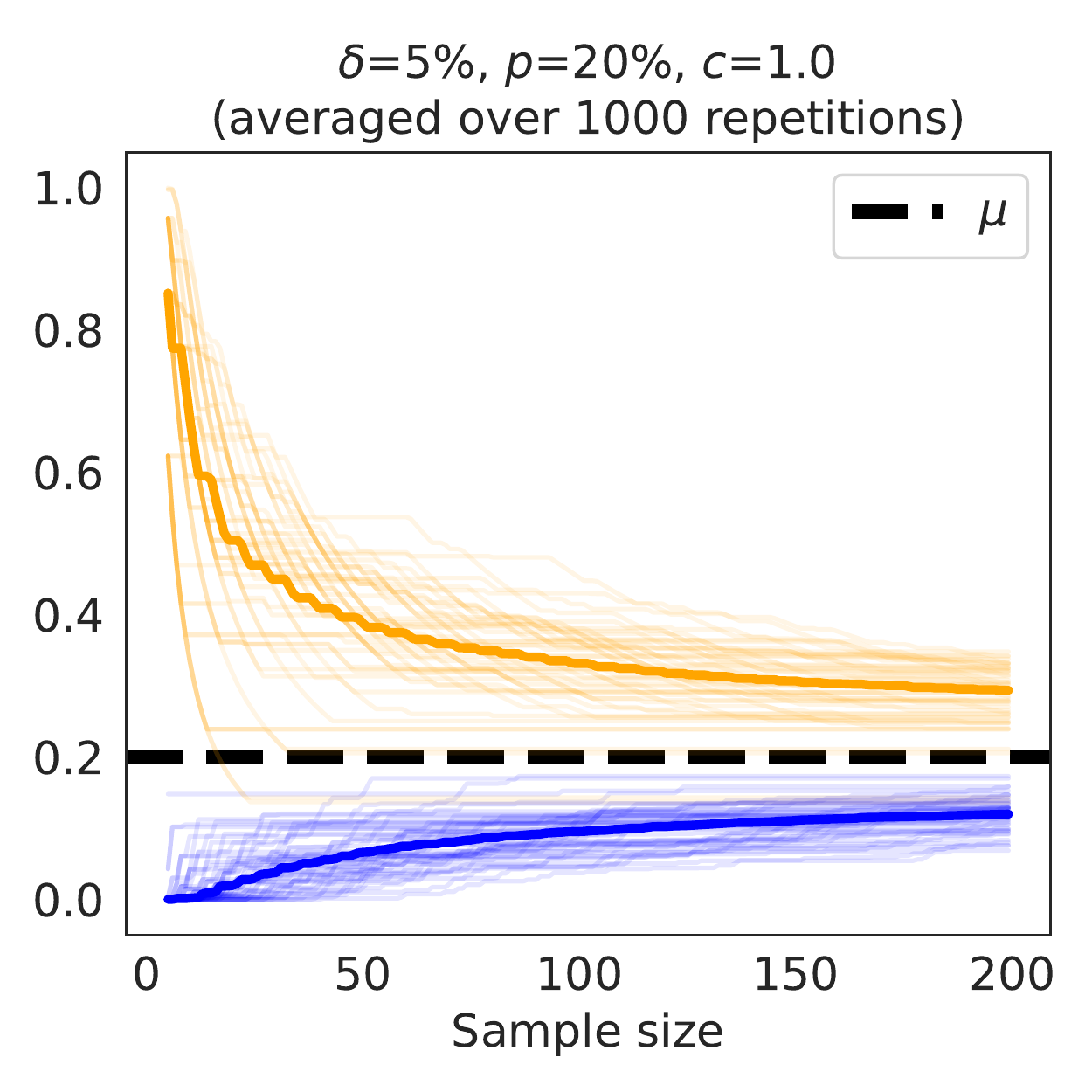}
	\includegraphics[width=0.49\linewidth]{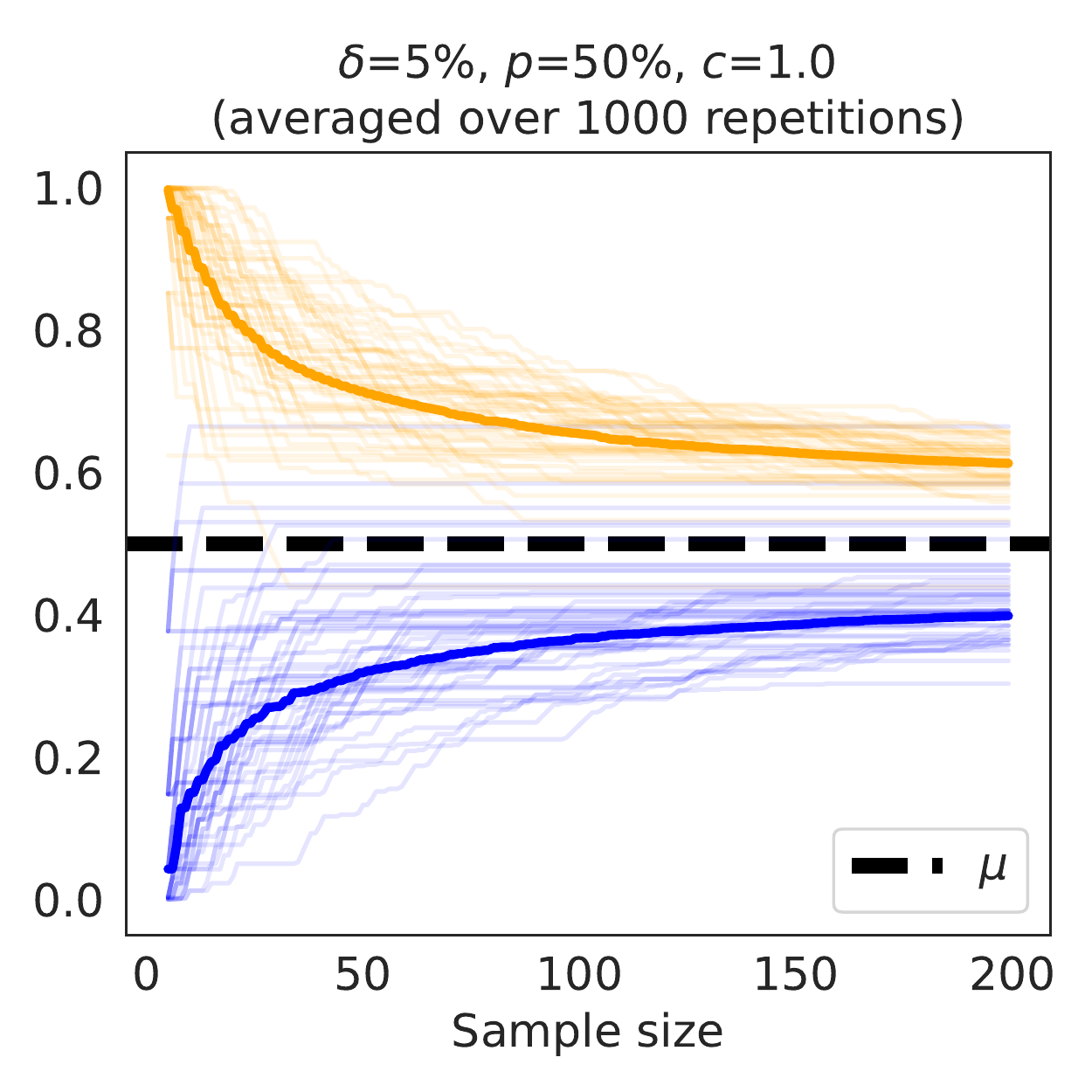}\\
	\includegraphics[width=0.49\linewidth]{figsLaplace/bregman_bernoulli_80.pdf}
	\includegraphics[width=0.49\linewidth]{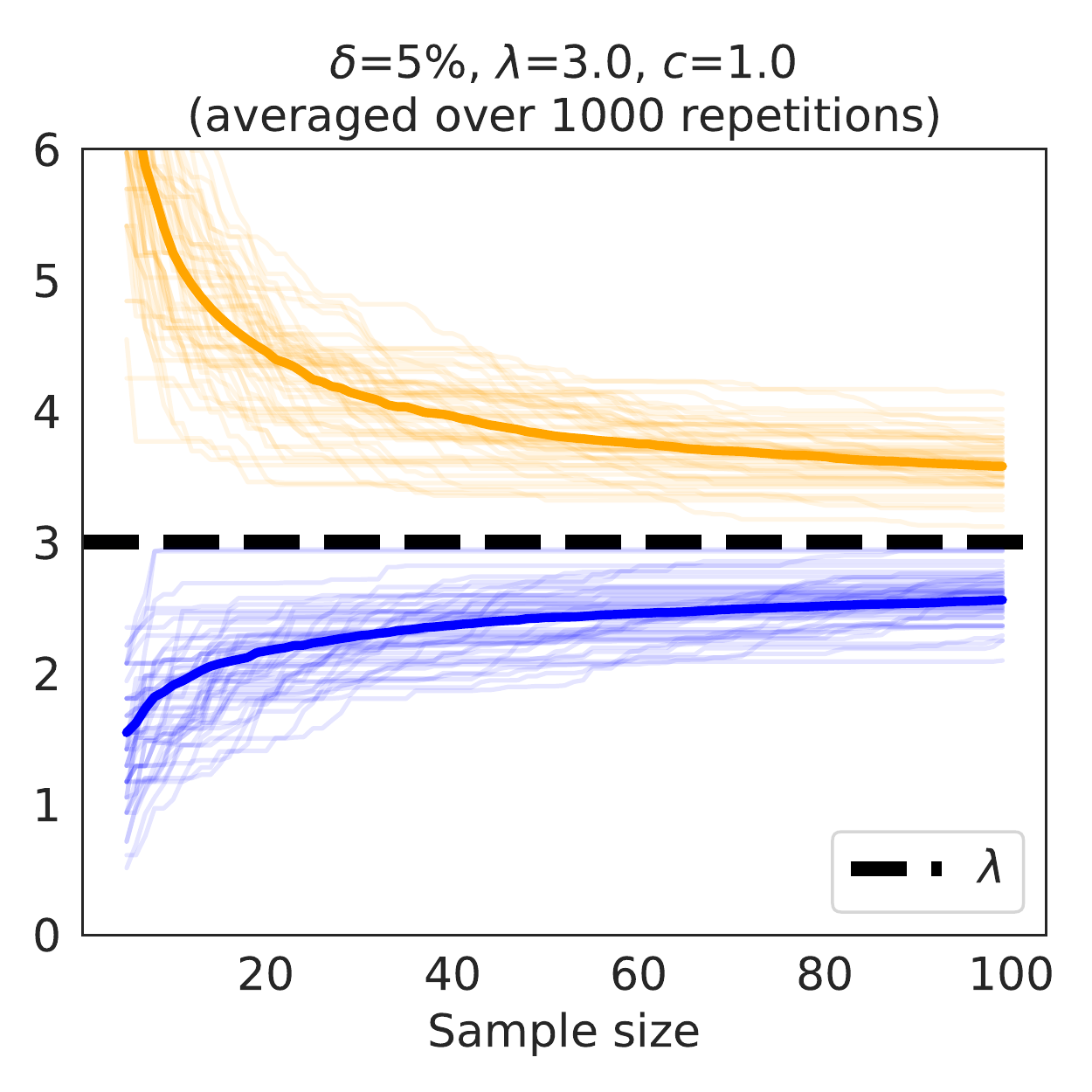}
	\caption{Examples of confidence upper and lower envelopes around expectation $\mu$ for discrete distributions. From top to bottom: $\text{Bernoulli}(p)$, $p\in\{0.2, 0.5, 0.8\}$ (cf. Table~\eqref{eqn:confsetBernoulli_full}) and $\text{Poisson}(3)$ (cf. Table~\ref{eqn:confsetPoisson_full}) on several realizations (each dashed lines) as a function of the number of observations $n$. The thick lines indicate the median curve over 1000 replicates.}
	\label{fig:confidencebands_discrete}
\end{figure}

\begin{figure}[H]
	\centering
	\includegraphics[width=0.49\linewidth]{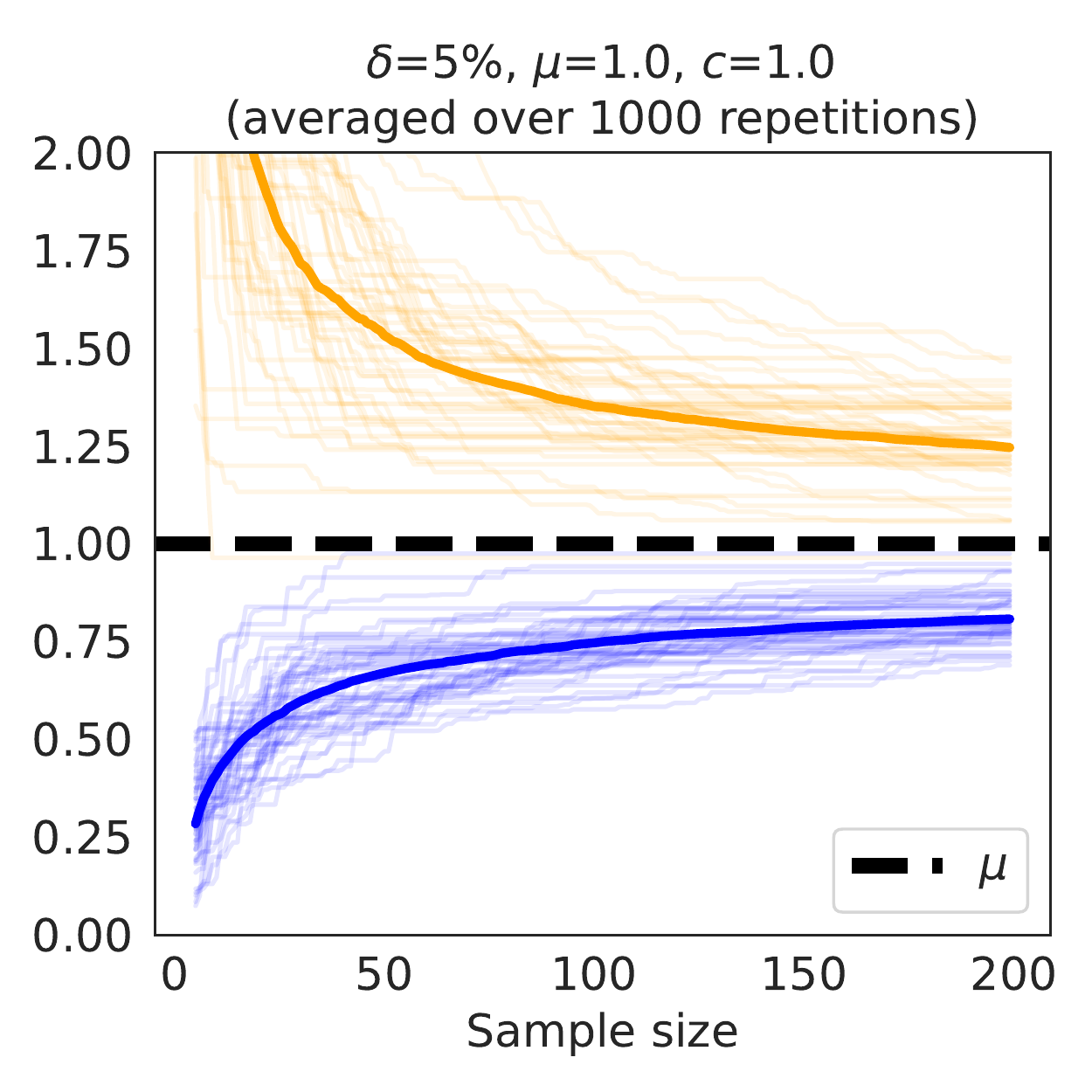}
    \includegraphics[width=0.49\linewidth]{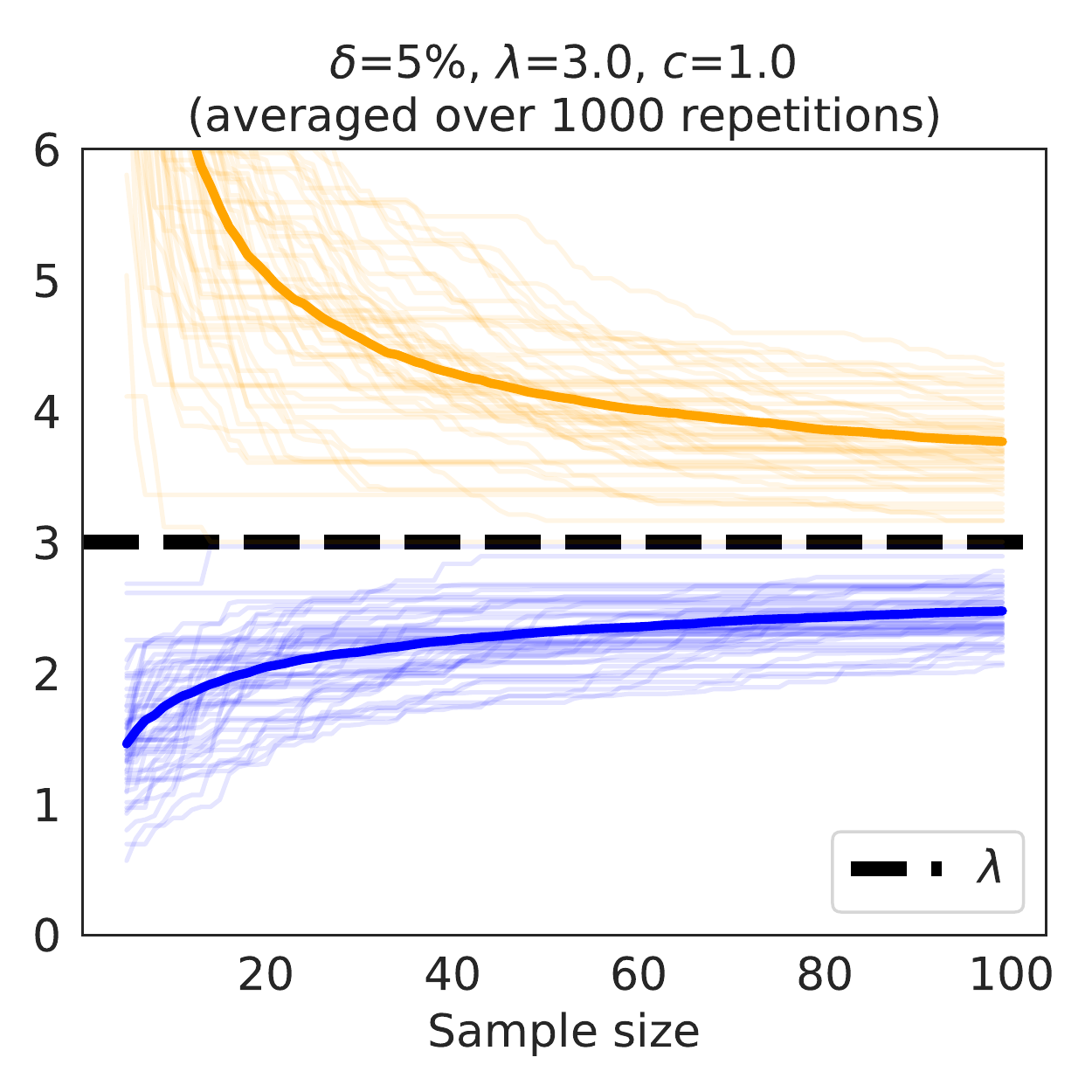}\\
	\includegraphics[width=0.49\linewidth]{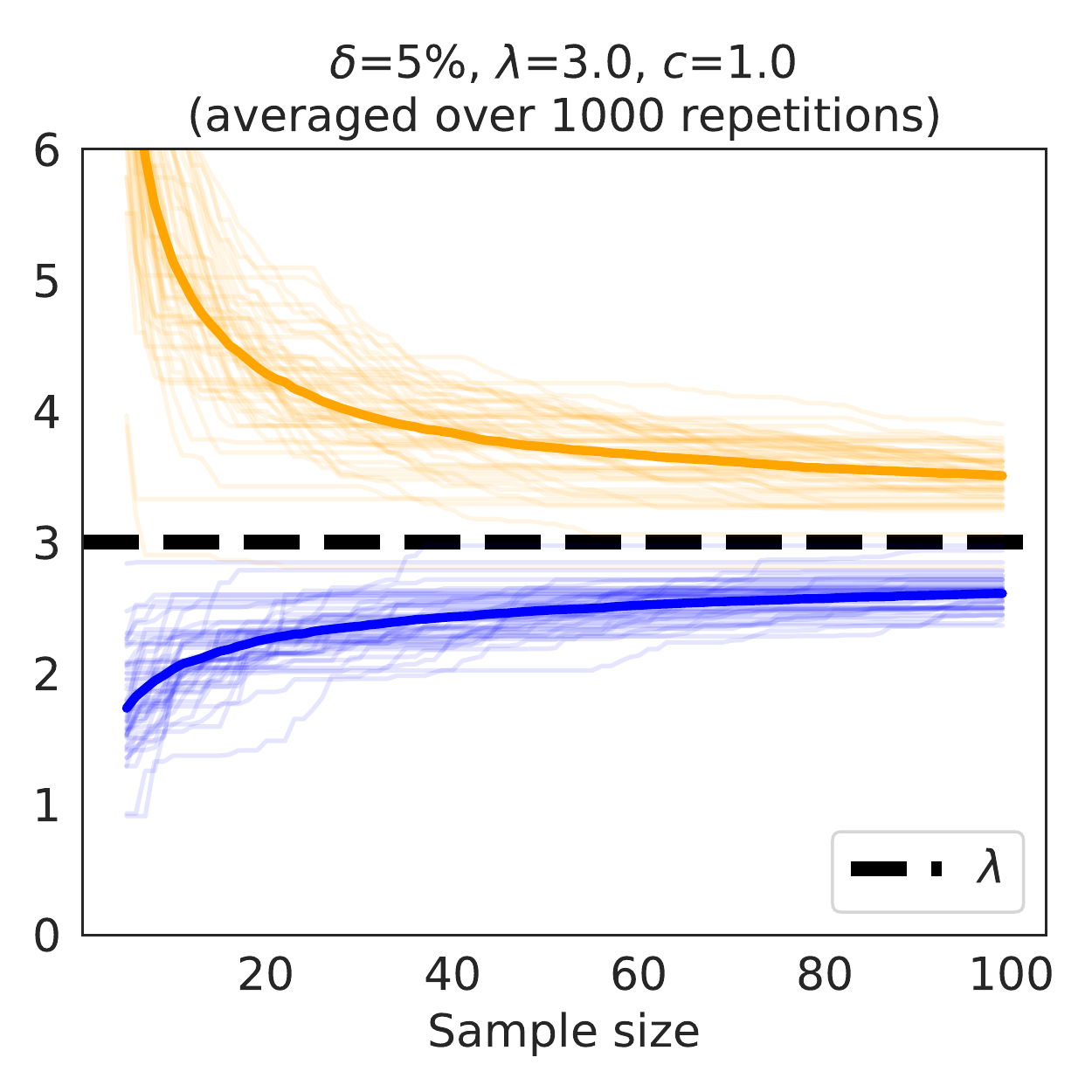}
    \includegraphics[width=0.49\linewidth]{figsLaplace/bregman_pareto_50.pdf}
	\caption{Examples of confidence upper and lower envelopes (from top to bottom) around scale parameter $\lambda$ for $\text{Exp}(1)$ (cf. Table~\eqref{eqn:confsetExpo_full}), fixed shape $\text{Gamma}(3, 2)$ (cf. Table~\ref{eqn:confsetGamma_full}), fixed shape $\text{Weibull}(3, 2)$ (cf. Table~\ref{eqn:confsetWeibull_full}), and exponent $\alpha$ for $\text{Pareto}(\frac{1}{2})$ (cf. Table~\ref{eqn:confsetPareto_full}), on several realizations (each dashed lines) as a function of the number of observations $n$. The thick lines indicate the median curve over 1000 replicates.}
	\label{fig:confidencebands_exp}
\end{figure}

\begin{figure}[H]
	\centering
	\includegraphics[width=0.49\linewidth]{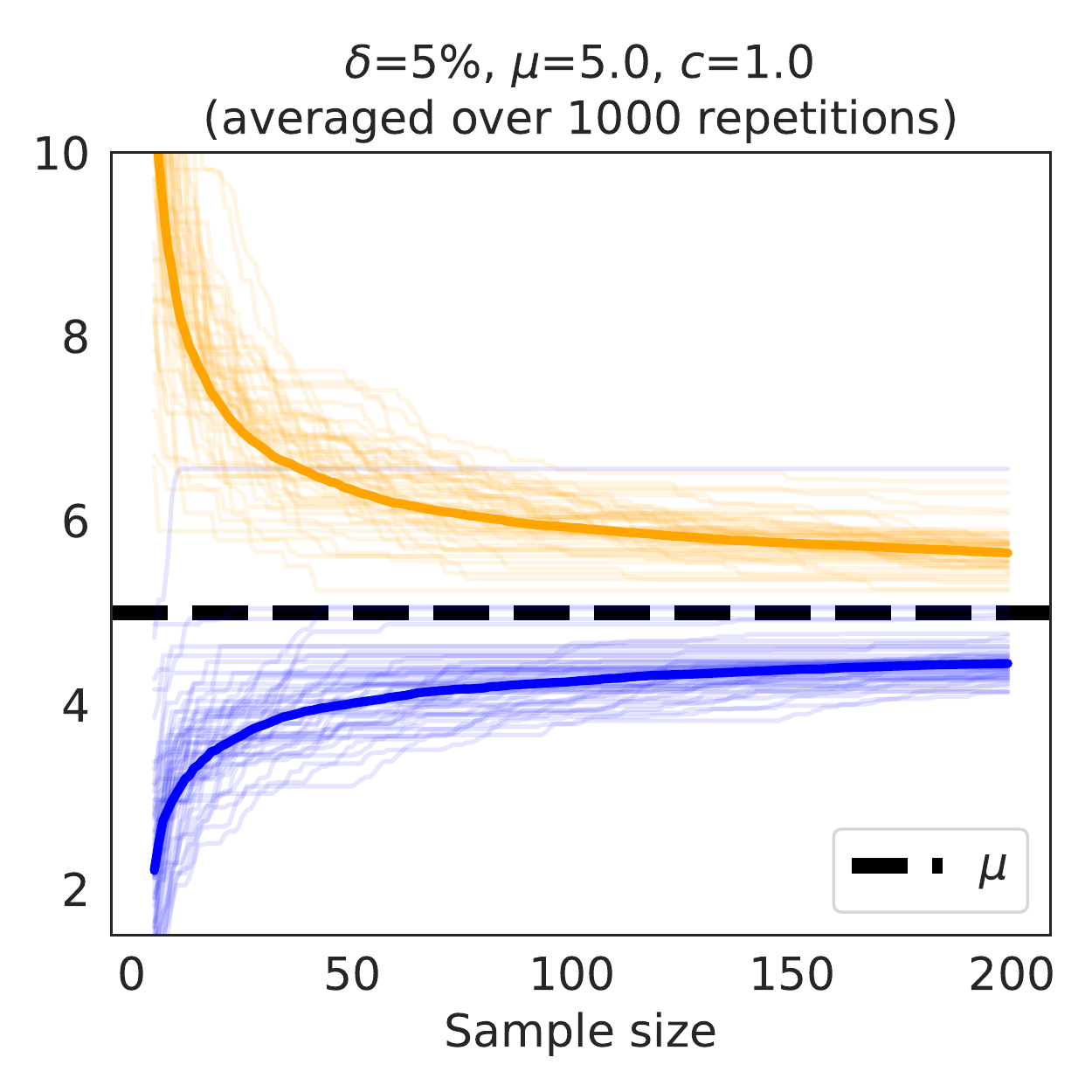}
	\includegraphics[width=0.49\linewidth]{figsLaplace/bregman_chi2_discrete_prior500.pdf}
	\caption{Examples of confidence upper and lower envelopes around expectation $\mu$ for (continuous prior) $\text{Gamma}\left(\frac{k}{2}\right)$ with unknown shape $\frac{k}{2}>0$ and (discrete prior) Chi-square $\chi^2(k)$, $k\in\mathbb{N}^*$ (cf. Table~\eqref{eqn:confsetXhi2_full} and Remark~\ref{rmk:chi2_gamma}) on several realizations (each dashed lines), as a function of the number of observations $n$. In particular for the Chi-square, confidence lower and upper bounds are ceiled and floored to integers. The thick lines indicate the median curve over 1000 replicates.}
	\label{fig:confidencebands_chi2}
\end{figure}

In what follows, we fix $\delta\in(0,1)$ the uniform confidence level. For each confidence sequence $\left(\Theta_n\right)_{n\in\Nat}(\delta)$, we report in the figures the intersection sequence $\left(\cap_{k\leq n} \Theta_k(\delta)\right)_{n\in\Nat}$, which also holds with confidence $1 - \delta$, for $n$ up to $200$. 
Typical realizations of Bregman confidence sequences are reported in Figure~\ref{fig:confidencebands_gaussian} (Gaussian), Figure~\ref{fig:confidencebands_discrete} (Bernoulli, Poisson), Figure~\ref{fig:confidencebands_exp} (Exponential, Gamma, Weibull, Pareto), and Figure~\ref{fig:confidencebands_chi2} (Chi-square).

\subsection{Gaussian with unknown mean and variance}\label{app:gauss}

\begin{figure}[H]
	\centering
	\includegraphics[width=0.49\linewidth]{figsLaplace/bregman_gaussian_2d_100_100.pdf}
	\includegraphics[width=0.49\linewidth]{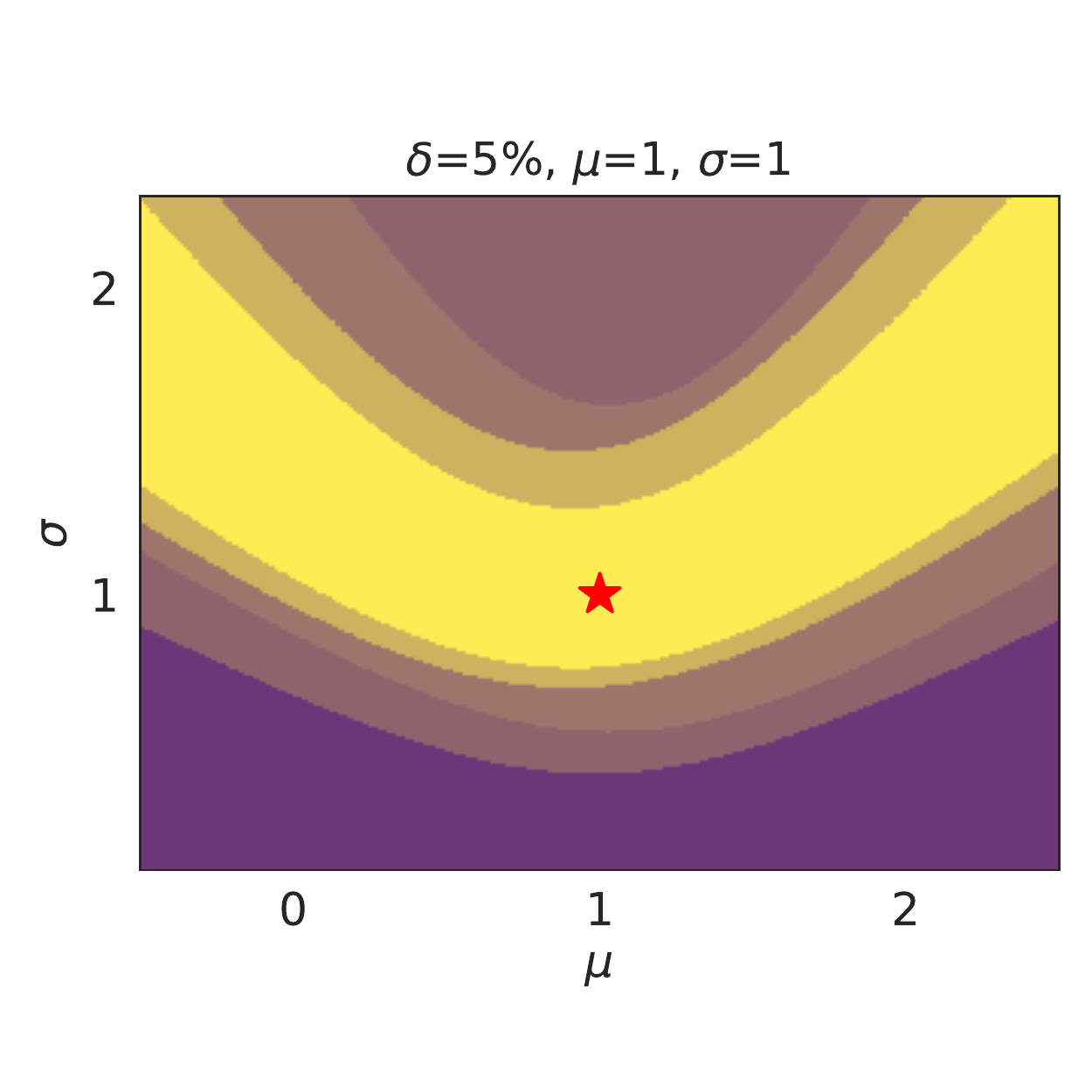}\\
	\caption{Example of time-uniform, joint confidence sets around $(\mu, \sigma)$ for $\cN\left(1, 1\right)$ (cf. Table~\ref{eqn:gauss-mean--var_full} for $n\in\left\lbrace 10, 25, 50, 100\right\rbrace$ observations (smaller confidence sets correspond to larger sample sizes). The red star indicates the true parameters $(\mu, \sigma) = (1, 1)$.}
	\label{fig:confidencesets_gaussian2d}
\end{figure}

We consider the two-dimensional family $\left\lbrace \cN\left(\mu, \sigma^2\right), \left(\mu, \sigma\right)\in \bR \times \bR^*_+ \right\rbrace$. To the best of our knowledge, there does not exist time-uniform joint confidence sets for $\left(\mu, \sigma\right)$ prior to this work. In order to compare ourselves against some baseline, we derive a crude one based on Chi-square quantiles and a union bound.

Let $n\in\Nat, \delta\in(0, 1)$, $X_1, \dots, X_n$ i.i.d samples from $\cN(\mu_0, \sigma_0^2)$. The standardized sum of squares 
\[Z_t(\mu_0, \sigma_0) = \sum\limits_{s=1}^t \left(\frac{X_t - \mu_0}{\sigma_0}\right)^2\]
follows a $\chi^2(t)$ distribution, therefore, denoting by $q_{\chi^2(t)}$ the corresponding quantile function, we have:
\[\bP\left( q_{\chi^2(t)}\left(\frac{\delta}{2}\right) \leq Z_t(\mu_0, \sigma_0) \leq q_{\chi^2(t)}\left(1 - \frac{\delta}{2}\right)\right) \geq 1 - \delta\,.\]
By a union bound argument, the intersection of $n$ such events with confidence $\frac{\delta}{n}$, i.e
\[ \Theta^{Z}_{t, n}(\delta) = \bigcap\limits_{s=1}^t \left\lbrace (\mu, \sigma) \in \bR \times \bR_+^*\colon\ q_{\chi^2(s)}\left(\frac{\delta}{2n}\right) \leq Z_s(\mu, \sigma) \leq q_{\chi^2(s)}\left(1 - \frac{\delta}{2n}\right) \right\rbrace\,\]
describes a sequence of confidence sets at level $\delta$ that hold uniformly over $t\in\{1, \dots, n\}$.

We report our confidence sets (cf. Equation~\ref{eqn:gauss-mean--var_full}) and the above on Figure~\ref{fig:confidencesets_gaussian2d}. The most striking drawback of $\Theta^{Z}_{t, n}(\delta)$ is that it is not convex nor even bounded; in particular, projecting onto the axes of $\mu$ and $\sigma$ only provides trivial confidence sets, $\bR$ and $\bR_+^*$ respectively, rendering this result vacuous. To better grasp this phenomenon, let us informally consider $\mu=\alpha + \mu_0$ and $\sigma=\alpha$ for some $\alpha>0$. We have:
\begin{align*}
    Z_n(\mu, \sigma) &= \sum\limits_{t=1}^n \left(\frac{X_t - \mu_0}{\alpha} - 1\right)^2\\
    &= \underbrace{\frac{1}{\alpha^2}\sum\limits_{t=1}^n \left(X_t - \mu_0\right)^2}_{\approx \frac{\sigma_0^2}{\alpha^2}n} - \underbrace{\frac{2}{\alpha} \sum\limits_{t=1}^n \left(X_t - \mu_0\right)}_{\approx \frac{2}{\alpha}\sqrt{n}} + n\,.
\end{align*}
Hence, for $\alpha\rightarrow +\infty$, $Z_{n}(\mu_0 + \alpha, \alpha) \approx n$. On the other hand, Lemma~1 in \citet{laurent2000adaptive} shows that $q_{\chi^2(n)}\left(1 - \delta/2n\right) = \mathcal{O}\left(n + \sqrt{n\log n}\right)$ and $q_{\chi^2(s)}\left(\delta/2n\right)=\mathcal{O}\left(n\right)$. Therefore, even with increasing the sample size $n$, there exists arbitrary large $\alpha$ such that $(\mu_0+\alpha, \alpha)$ may belong to $\Theta^{Z}_{t, n}(\delta)$.

By contrast, our Bregman confidence sets are convex and bounded, which we interpret as the result of exploiting the true geometry of the two-dimensional Gaussian family. In addition to the unboundedness, $\Theta^{Z}_{t, n}(\delta)$ is built using a crude union bound, which is not anytime (depends on the terminal time $n$) and rather loose.

\subsection{Bernoulli}
\label{app:bernoulli_comparison}

\begin{figure}[H]
	\centering
	\includegraphics[width=0.49\linewidth]{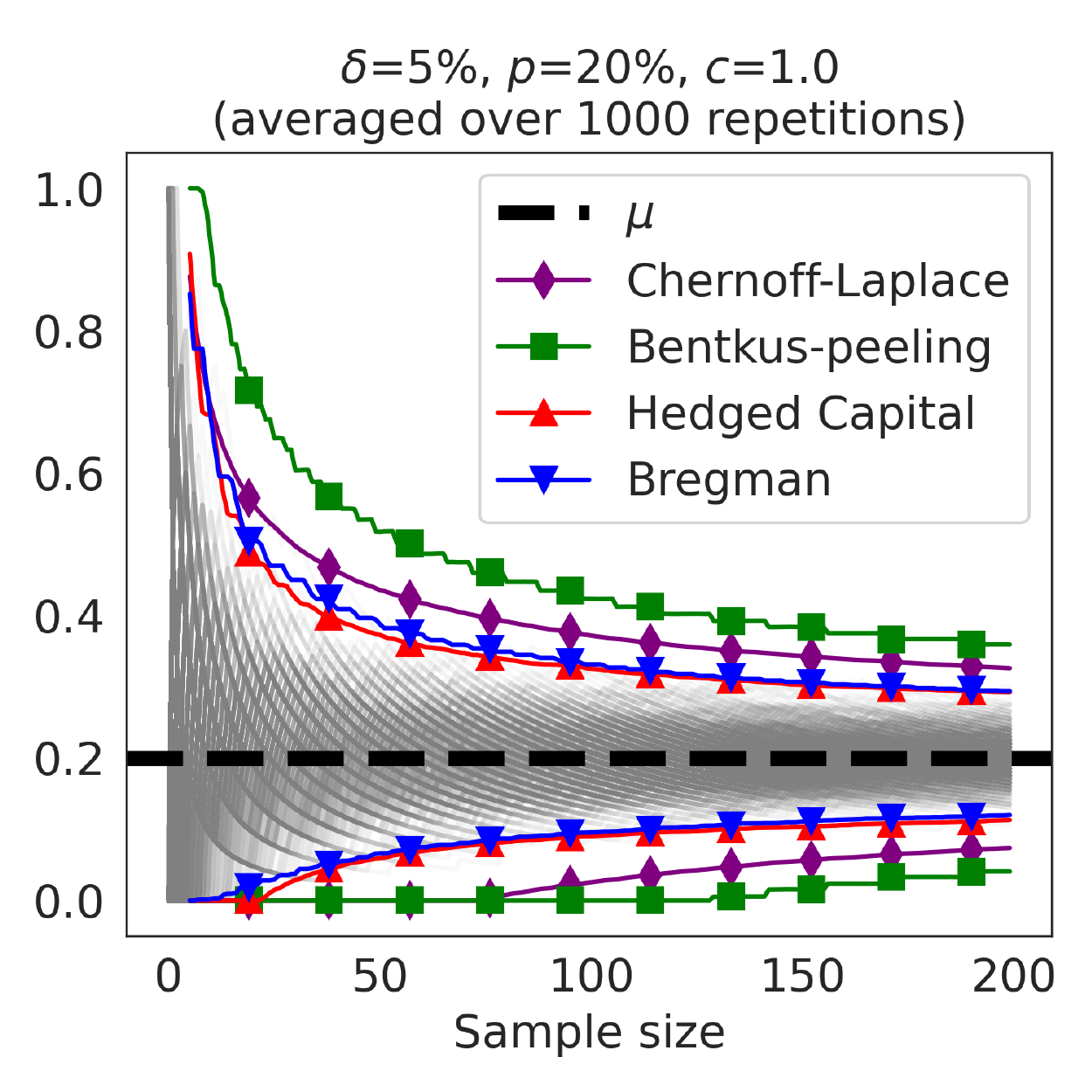}
	\includegraphics[width=0.49\linewidth]{figsLaplace/bernoulli_comparison_80.pdf}\\
	\includegraphics[width=0.49\linewidth]{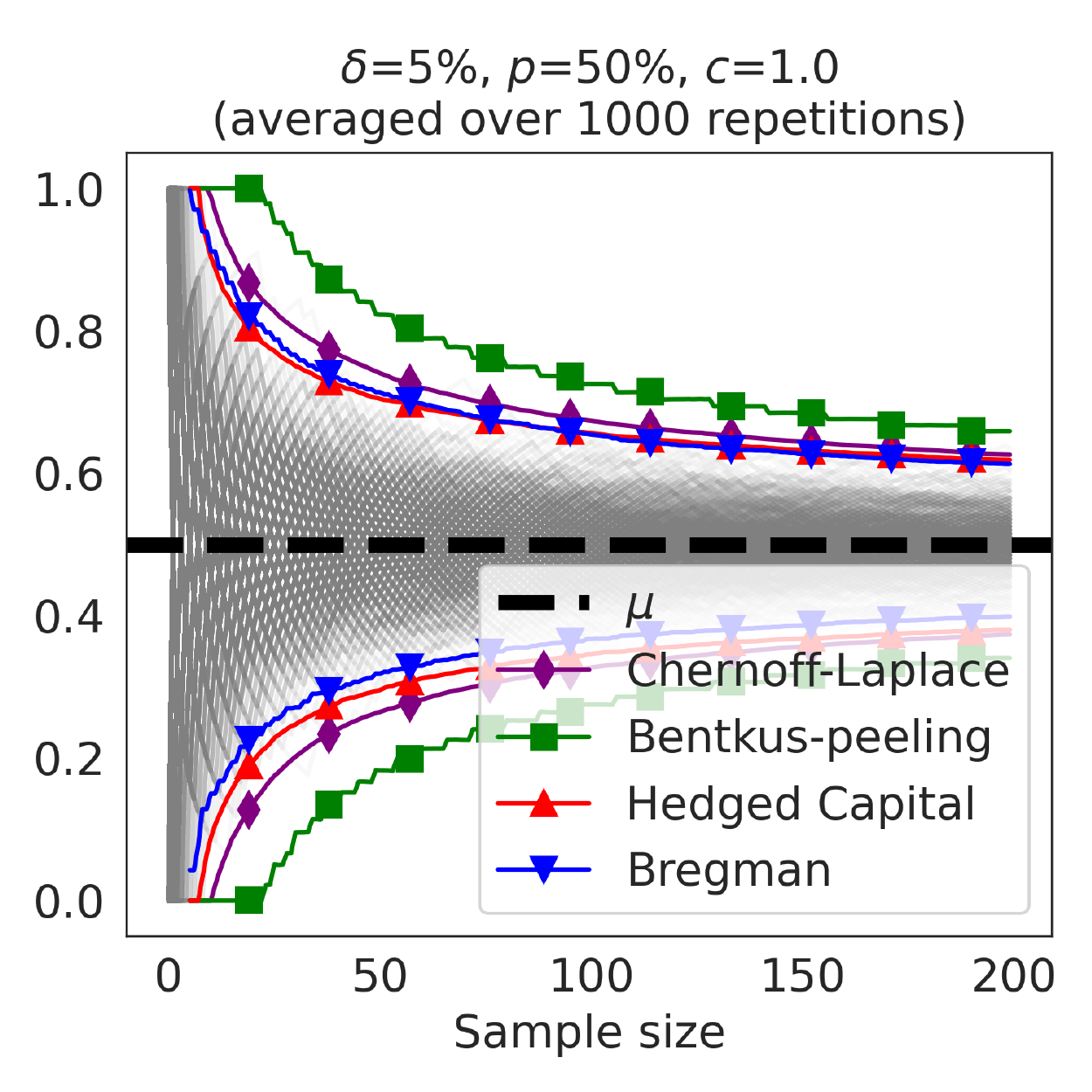}
	\caption{Confidence upper and lower envelopes built for $\text{Bernoulli}(\mu)$, $\mu\in\{0.2, 0.5, 0.8\}$, as a function of the number of observations $n$, averaged over $1000$ independent simulations. Grey lines are trajectories of empirical means $\widehat{\mu}_n$. Confidence bounds are clipped between $0$ and $1$.}
	\label{fig:confidencebandsbernoulli_comparison}
\end{figure}

We consider the Bernoulli distribution $\text{Bernoulli}(\mu)$ for some unknown $\mu\in[0, 1]$. Confidence bounds are displayed in Figure~\ref{fig:confidencebandsbernoulli_comparison} for $\mu\in\{0.2, 0.5, 0.8\}$. We detail below a few alternative state-of-the-art bounds.

\paragraph{Chernoff-Laplace method for sub-Gaussian distributions}

We recall that $B(\mu)$ being supported in $[0, 1]$, it is $1/2$-sub-Gaussian (Hoeffding lemma, \citet{hoeffding1963}). The method of mixtures for exponential martingales \citep{pena2008self} shows that $\bP\left( \forall n\in\bN, \mu \in \Theta^{\text{Laplace}}_n(\delta) \right) \geq 1-\delta$, with
\[\Theta^{\text{Laplace}}_n(\delta) = \left[\hat\mu_{n} - \sqrt{\left( 1 + \frac{1}{n}\right) \frac{\log\left( 2\sqrt{1+n}/\delta\right)}{2n}}, \hat\mu_{n} + \sqrt{\left( 1 + \frac{1}{n}\right) \frac{\log\left( 2\sqrt{1+n}/\delta\right)}{2n}} \right].\]

Note that a sharper sub-Gaussian control holds for $B(\mu)$, see \citet{kearns1998large}, \citet{berend2013concentration}. However, the optimal sub-Gaussian parameter they suggest ($\frac{1/2-\mu}{\log(1/\mu -1)}$ instead of $1/2$) is a function of the unknown $\mu$.

\paragraph{Bentkus concentration with geometric time-peeling}
\citet[Theorem 4]{pmlr-v139-kuchibhotla21a} show that $\bP\left( \forall n\in\bN, \mu \in \Theta^{\text{Bentkus}}_n(\delta) \right) \geq 1-\delta$ with
\[\Theta^{\text{Bentkus}}_n(\delta) = \left[\hat\mu_{n} - \frac{1}{n}q\left(\frac{\delta}{2h(k_n)}; c_n, 1/2, 1\right), \hat\mu_{n} + \frac{1}{n}q\left(\frac{\delta}{2h(k_n)}; c_n, 1/2, 1\right) \right].\]
Here $q(\delta'; N, A, B)$ is the Bentkus bound, where $\delta'$ is the confidence level, $N$ is the sample size, $B$ is the \emph{almost sure} upper bound of the random variables and $A^2$ is the variance upper bound. The function $h(k)$ is defined as $h(k)=\zeta(1.1)(k+1)^{1.1}$, where $\zeta(\cdot)$ is the Riemann-zeta function. The integer $k_n$ and the real $c_n$ are defined as $k_n=\min\{k\in\Nat: \lceil \eta^k \rceil \leq n \leq \lfloor \eta^{k+1} \rfloor\}$ and $c_n=\lfloor \eta^{k_{n}+1}\rfloor$, with $\eta=1.1$. The referenced theorem is stated with an empirical overestimate of the standard deviation instead of the fixed bound $1/2$, which would lead to replace the $\delta/2$ present in the confidence set with $\delta/3$ by union bound; we found this step to be of negligible, and even slightly detrimental impact, in the case of Bernoulli distributions.

\paragraph{Hedged Capital martingale method}

\citet[Theorem 3]{waudby2020estimating} present a nonnegative martingale construction from observations $X_1, \dots, X_n$ bounded in $[0, 1]$. Following the recommendations of the authors, we define the following quantities:
\begin{align*}
    &\cK_n(m) = \max(\cK_n^{+}, \cK_n^-(m)),\\
    &\cK_n^{\pm}(m) = \prod_{k=1}^n \left( 1 \pm \lambda_k^{\pm}(m)\left( X_k - m\right)\right),\\
    &\lambda_k^{+}(m) = \min\left(\lvert \lambda_k \rvert, \frac{1/2}{m}\right), \lambda_k^{-}(m) = \min\left(\lvert \lambda_k \rvert, \frac{1/2}{1-m}\right),\\
    &\lambda_k = \sqrt{\frac{2\log(2/\delta)}{\hat\sigma_{k-1}^2 k \log(k+1)}},\\
    &\hat\sigma_k^2 = \frac{1/4 + \sum_{i=1}^k \left(X_i - \hat \mu_i\right)^2}{k + 1},\\
    &\hat\mu_k = \frac{1/2 + \sum_{i=1}^k X_i}{k + 1}.
\end{align*}
Then, it holds that $\bP\left( \forall n\in\bN, \mu \in \Theta^{\text{HC}}_n(\delta) \right) \geq 1-\delta$, where
$\Theta^{\text{HC}}_n(\delta) = \left\lbrace m\in[0, 1]: \cK_n(m) < 1 / \delta\right\rbrace$.

\subsection{Gaussian}
\label{app:gaussian_comparison}

\begin{figure}[H]
	\centering
	\includegraphics[width=0.6\linewidth]{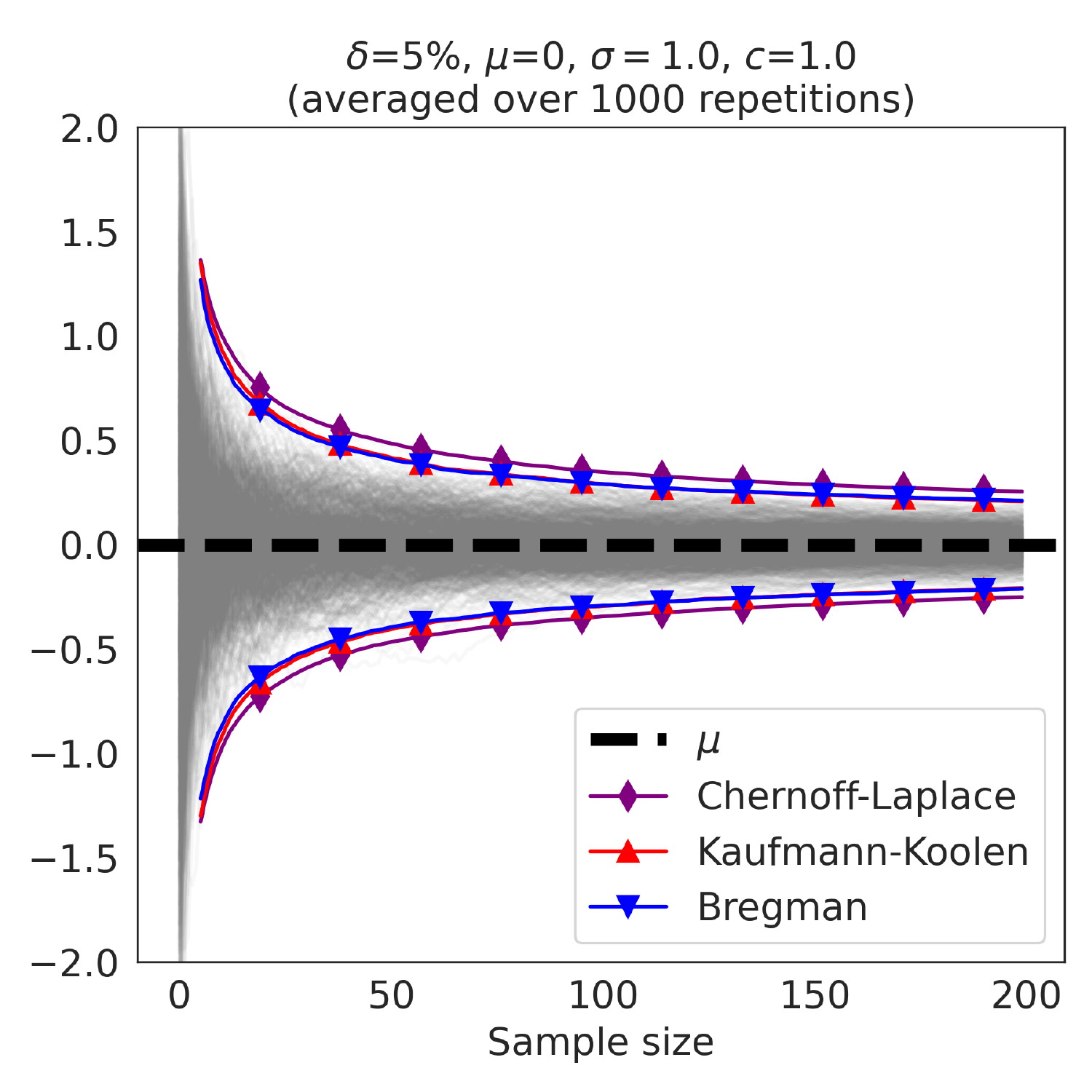}
	\caption{Confidence upper and lower envelopes built for Gaussian distributions $\cN(0, 1)$ as a function of the number of observations $n$, averaged over $1000$ independent simulations. Grey lines are trajectories of empirical means $\widehat{\mu}_n$.}
	\label{fig:confidencebandsgaussian_comparison}
\end{figure}

We consider Gaussian distributions $\cN(\mu, \sigma^2)$ for some unknown $\mu\in\bR$ and known variance $\sigma^2$. The confidence bounds are displayed in Figure~\ref{fig:confidencebandsgaussian_comparison} for $\mu=0$.

\paragraph{Chernoff-Laplace}
Similarly to the Bernoulli case, $\cN(\mu, \sigma^2)$ is $\sigma$-sub-Gaussian, therefore\\ 
$\bP\left( \forall n\in\bN, \mu \in \Theta^{\text{Laplace}}_n(\delta) \right) \geq 1-\delta$, with
\[\Theta^{\text{Laplace}}_n(\delta) = \left[\hat\mu_{n} - \sigma\sqrt{\left( 1 + \frac{1}{n}\right) \frac{\log\left( 2\sqrt{1+n}/\delta\right)}{n}}, \hat\mu_{n} + \sigma\sqrt{\left( 1 + \frac{1}{n}\right) \frac{\log\left( 2\sqrt{1+n}/\delta\right)}{n}} \right].\]

\paragraph{Kaufmann-Koolen}
\citet{kaufmann2021mixture} introduces a martingale construction for exponential families to derive time-uniform deviation inequalities under bandit sampling. However, application of their result is limited to Gaussian distribution with known variance and Gamma distribution with known shape, which is just a scaled version of exponential distribution.
Restricting Corollary 10 of \citet{kaufmann2021mixture} to the case of a single arm yields $\bP\left( \forall n\in\bN, \mu \in \Theta^{\text{KK}}_n(\delta) \right) \geq 1-\delta$, with
\begin{align*}
    &\Theta^{\text{KK}}_n(\delta) = \left\lbrace \cB_{\cL}\left( \hat\mu_n, \mu\right) \leq \frac{2}{n}\log\left(4 + \log(n)\right) + \frac{1}{n}C^g\left(\log 1/\delta\right)\right\rbrace,\\
    &g\colon \lambda\in (1/2, 1]\mapsto 2\lambda\left(1 - \log(4\lambda)\right) + \log \zeta(2\lambda) - \frac{1}{2}\log(1-\lambda),\\
    &C^g\colon x\in(0, +\infty) \mapsto \min_{\lambda\in(1/2, 1]}\frac{g(\lambda) + x}{\lambda},\\
    &\cL(\theta)=\frac{\theta^2}{2\sigma^2} \quad \text{(log-partition function of $\cN(\cdot, \sigma^2)$)}.
\end{align*}

\subsection{Exponential}
\label{app:exponential_comparison}

\begin{figure}[H]
	\centering
	\includegraphics[width=0.6\linewidth]{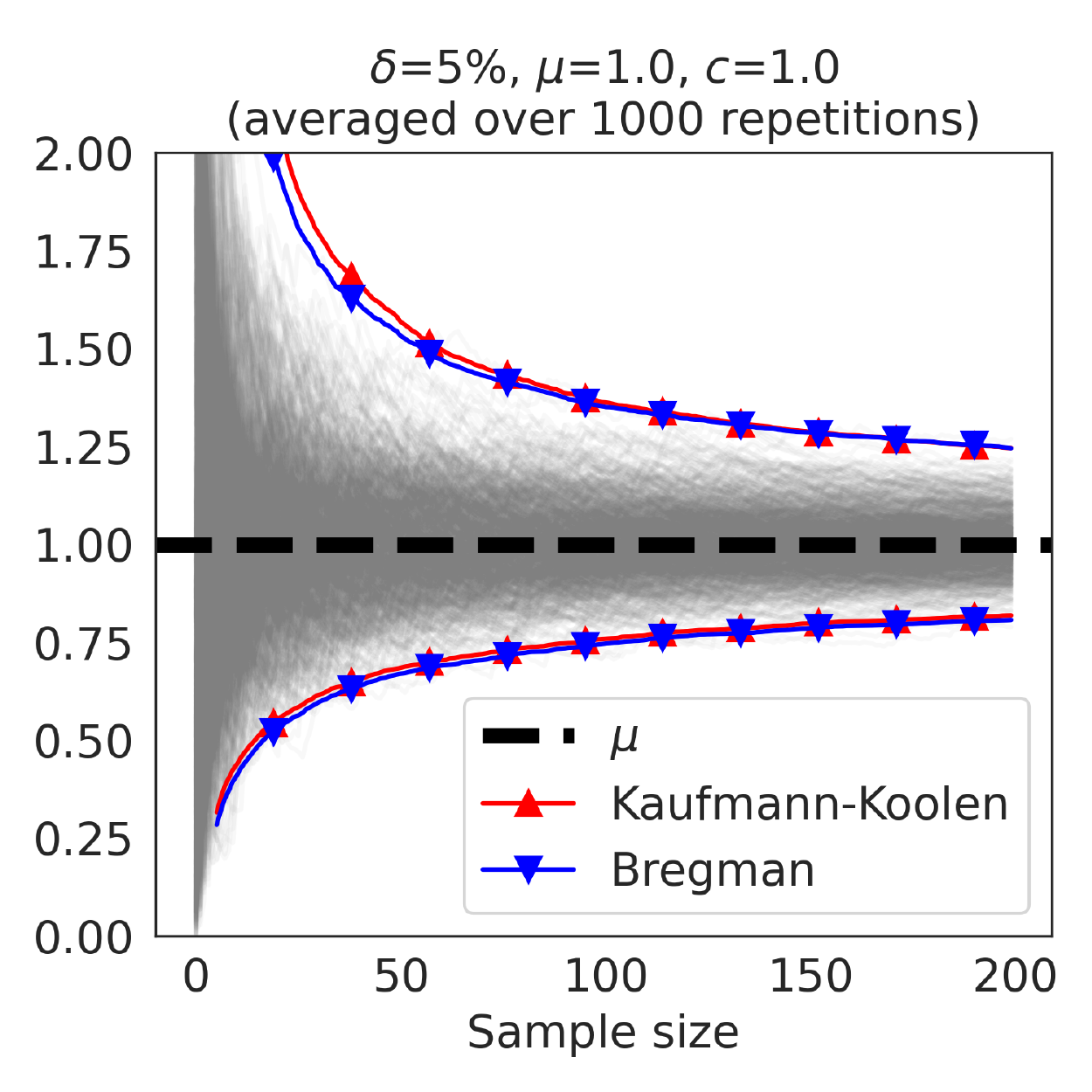}\\
	\caption{Confidence upper and lower envelopes built for Exponential distributions $\text{Exp}(1)$ as a function of the number of observations $n$, averaged over $1000$ independent simulations. Grey lines are trajectories of empirical means $\widehat{\mu}_n$.}
	\label{fig:confidencebandsexponential_comparison}
\end{figure}

We now consider exponential distributions $\text{Exp}(1/\mu)$ for some unknown mean $\mu\in\bR$. We report in Figure~\ref{fig:confidencebandsexponential_comparison} the confidence bounds for the case when $\mu=1$.

\paragraph{Kaufmann-Koolen}
\citet[Corollary 12]{kaufmann2021mixture} show that $\bP\left( \forall n\in\bN, \mu \in \Theta^{\text{KK}}_n(\delta) \right) \geq 1-\delta$, with the same definition as for the Gaussian case except
\begin{align}
    &g\colon \lambda\in (1/2, 1]\mapsto 2\lambda\left(1 - \log(4\lambda)\right) + \log \zeta(2\lambda) - \log(1-\lambda),\\
    &\cL(\theta)= \log(-1/\mu) \quad \text{(log-partition function of $\cE(1/\mu)$)}.
\end{align}

\subsection{Chi-square}
We consider $\text{Gamma}\left( \frac{k}{2}, \frac{1}{2}\right)$ and $\chi^2\left(k\right)$ distributions for some unknown $k>0$ and $k\in\mathbb{N}^*$ respectively. They are in fact the same distributions, however we distinguish both as the restriction on the domain for $k$ (real or integer) bears two consequences:
\begin{itemize}
    \item the mixture distribution in the martingale construction is either continuous or discrete, resulting in integrals or sums in the expression of the confidence sequence (Equation~\ref{eqn:confsetXhi2_full});
    \item confidence bounds are constrained to be integers in the discrete case, thus ceiling and flooring the lower and upper bounds respectively.
\end{itemize}

\begin{figure}[H]
	\centering
	\includegraphics[width=0.6\linewidth]{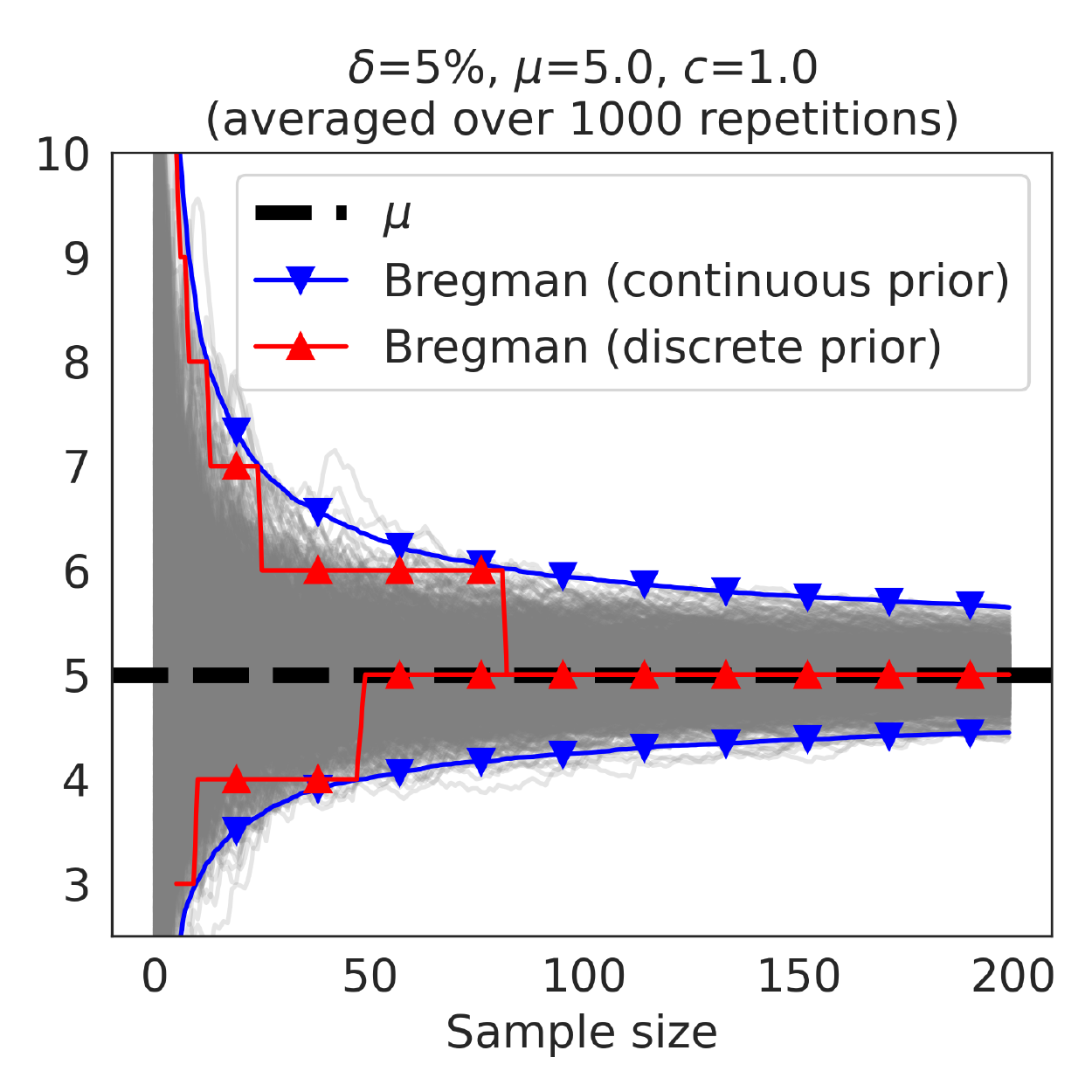}
	\caption{Confidence upper and lower envelopes built for $\text{Gamma}\left(\frac{5}{2}, \frac{1}{2}\right)$ (blue) and $\chi^2\left(5\right)$ (red) distributions as a function of the number of observations $n$, over $1000$ independent simulations (colored lines are median of simulated bounds). Grey lines are trajectories of empirical means $\widehat{\mu}_n$.}
	\label{fig:confidencebandschi2_comparison}
\end{figure}

We find that the former is negligible as both the continuous and discrete mixtures yield the same bound within numerical precision. The latter however allows to drastically shrink the size of the confidence sequences, resulting in perfect identification the mean within $95\%$ confidence with less than $100$ observations in half the simulations (see Figure~\ref{fig:confidencebandschi2_comparison}).

\subsection{Key Observations}
On the studied examples, confidence intervals based on time-uniform Bregman concentration are either comparable with state-of-the art methods for the corresponding setting or result in sharper bounds, especially for small sample sizes (lower bounds for Bernoulli compared to Hedged Capital, upper bound for Exponential compared to Kaufmann-Koolen). In particular, due to the formulation in terms of Bregman divergence, our intervals are naturally asymmetric when the underlying distribution is, and respect the support constraints (for instance Bernoulli Bregman bounds are in $[0, 1]$ without the need for clipping). Moreover, we also provide bounds in novel settings for which, to the best of our knowledge, time-uniform confidence sets are lacking (Chi-square, Poisson, Weibull, Pareto, mean-variance for Gaussian).

\subsection{Tuning of The Regularization Parameter $c$}\label{app:tuning}
\begin{figure}[H]
	\centering
	\begin{subfigure}[t]{.45\linewidth}
        \includegraphics[width=\linewidth]{figsLaplace/bernoulli_c_sensi_50.pdf}
        \caption{Bernoulli}
    \end{subfigure}
    \begin{subfigure}[t]{.45\linewidth}
	   \includegraphics[width=\linewidth]{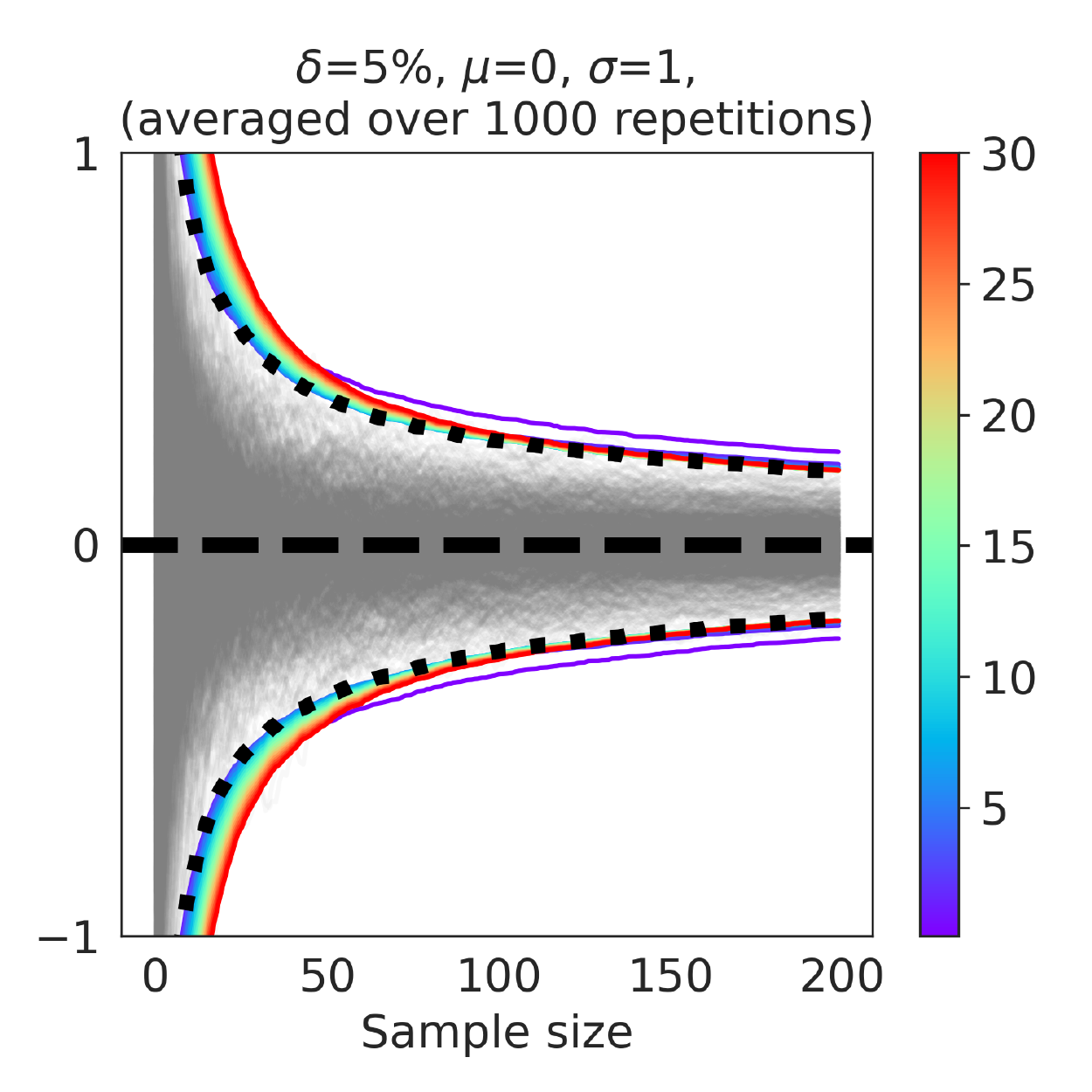}\\
	   \caption{Gaussian}
    \end{subfigure}\\
    \begin{subfigure}[t]{.45\linewidth}
        \includegraphics[width=\linewidth]{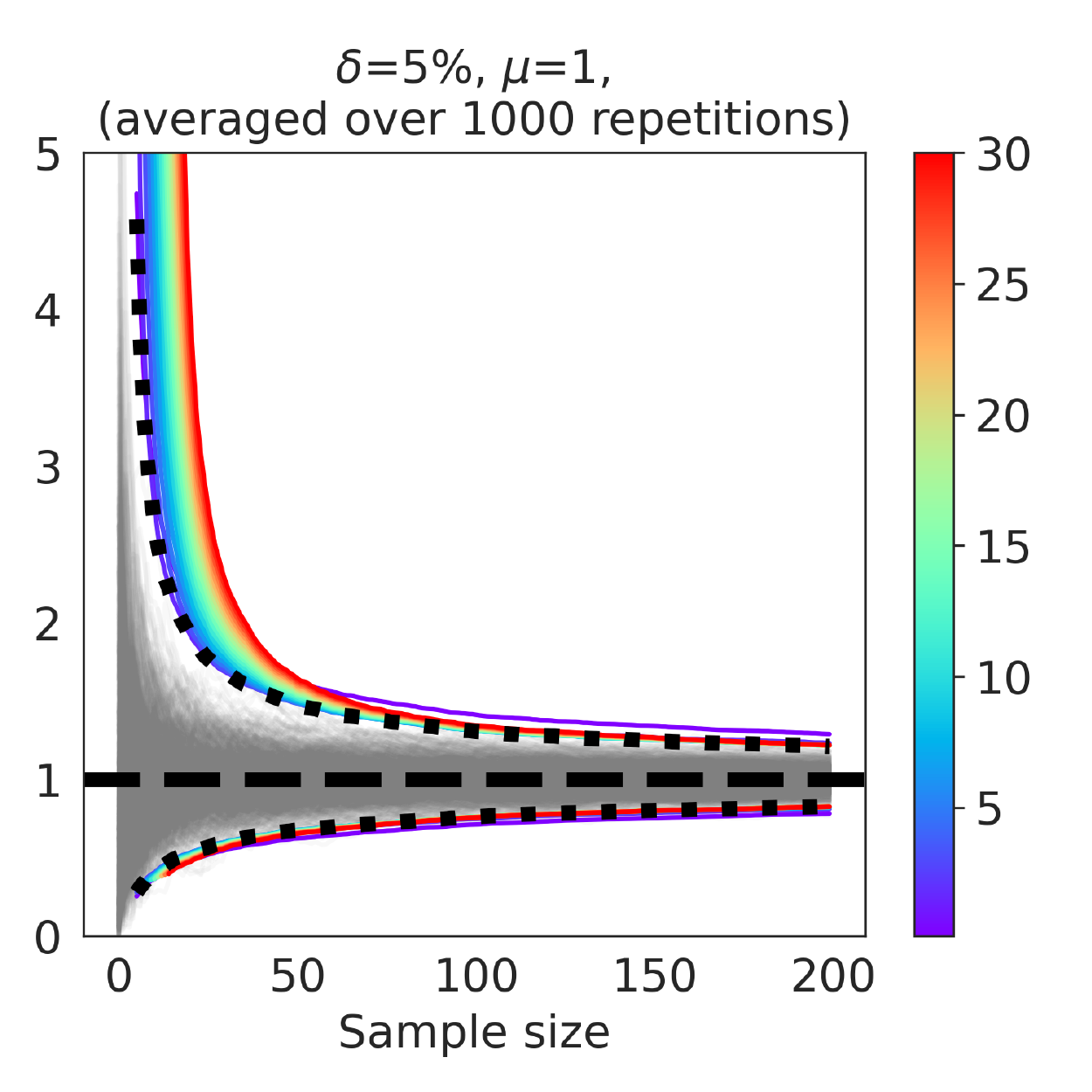}
    	\caption{Exponential}
    \end{subfigure}
    \begin{subfigure}[t]{.45\linewidth}
        \includegraphics[width=\linewidth]{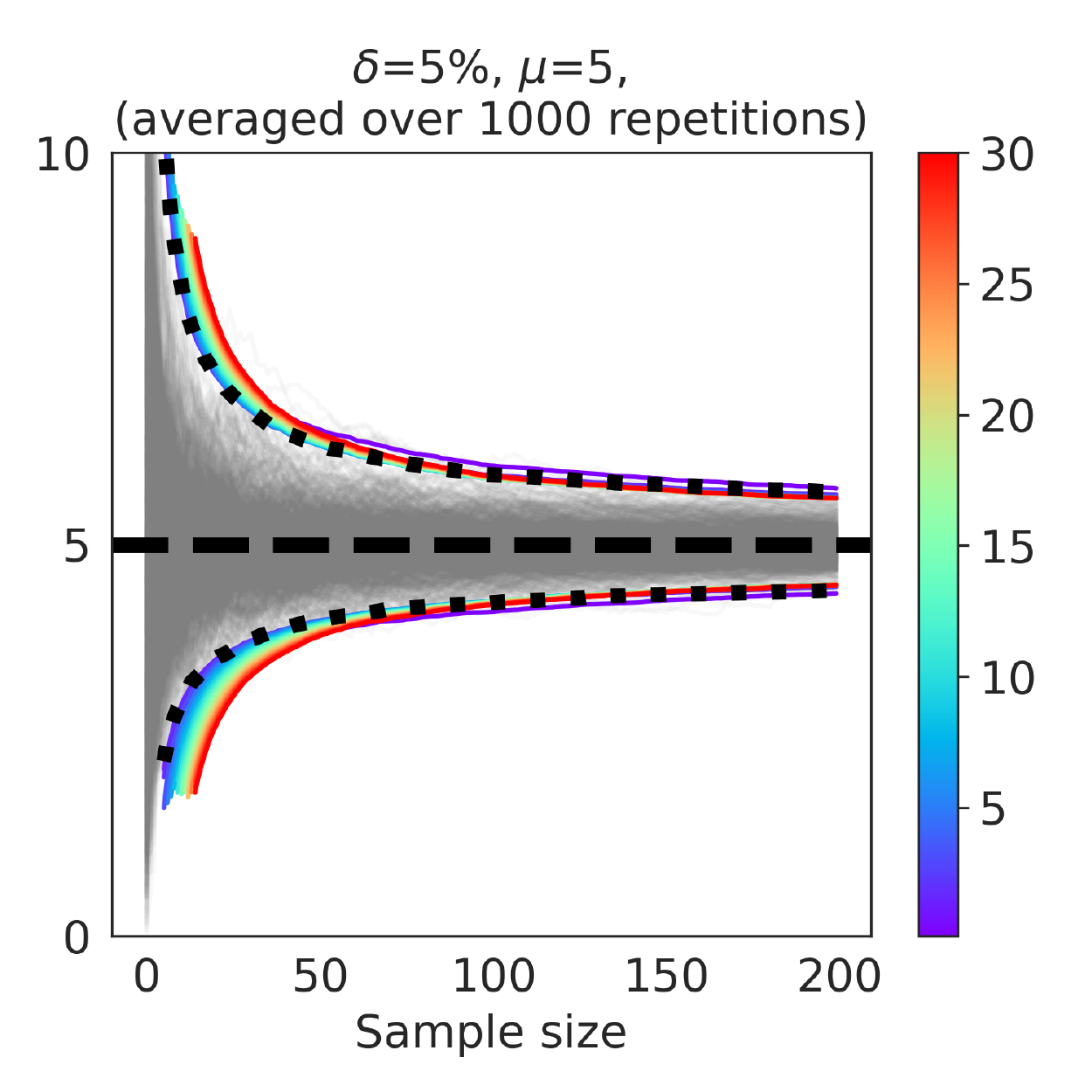}
        \caption{Chi-square}
	\end{subfigure}
    \caption{Confidence envelopes for varying $c\in[0.1, 30]$ for $\cB(0.5)$ (top left), $\cN(0, 1)$ (top right), $\cE(1)$ (bottom left) and $\chi^2(5)$ (bottom right). The dotted black line corresponds to the heuristics $c_n\approx 0.12 n$.}
	\label{fig:c_sensi-app}
\end{figure}

\begin{figure}[H]
	\centering
	\includegraphics[width=0.49\linewidth]{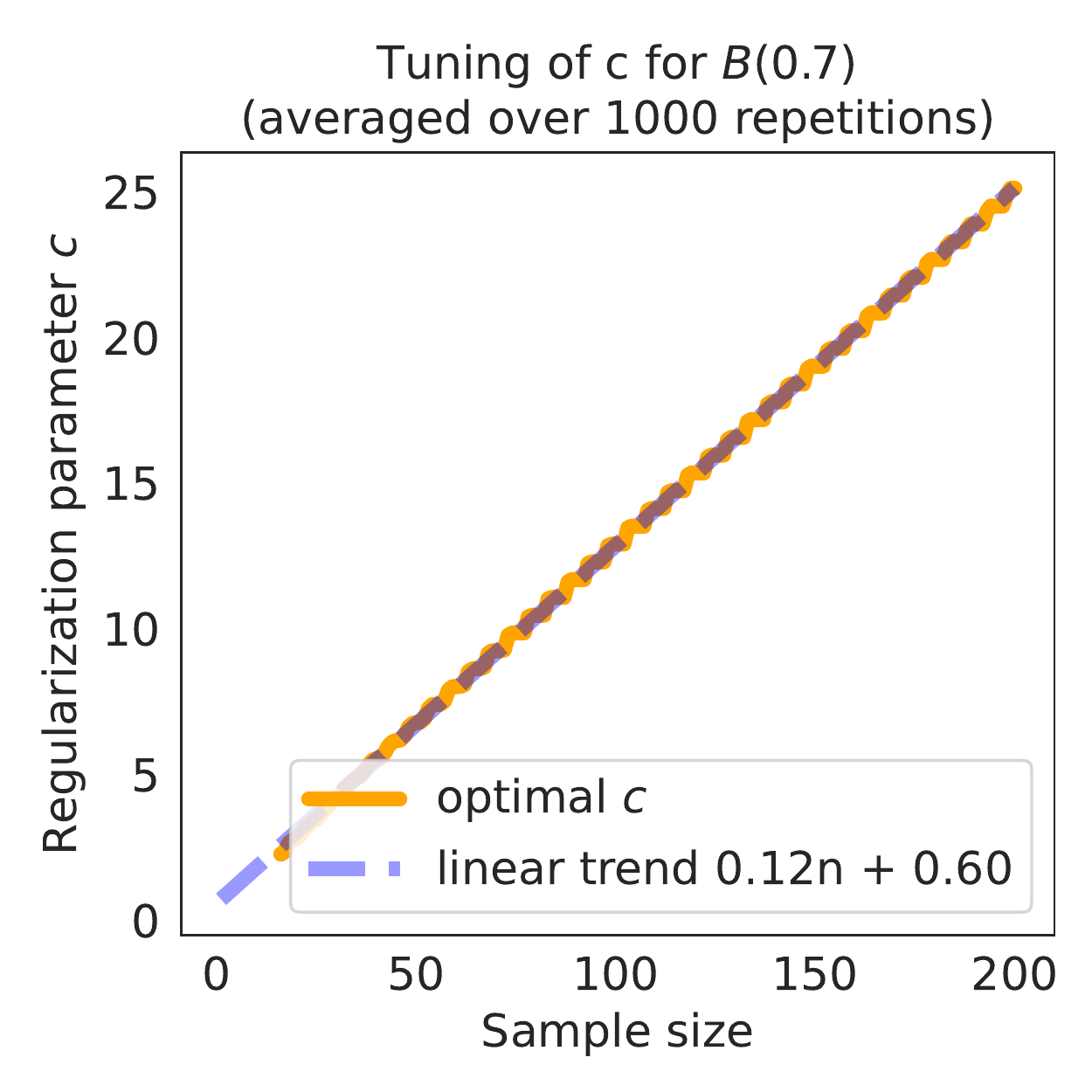}
	\includegraphics[width=0.49\linewidth]{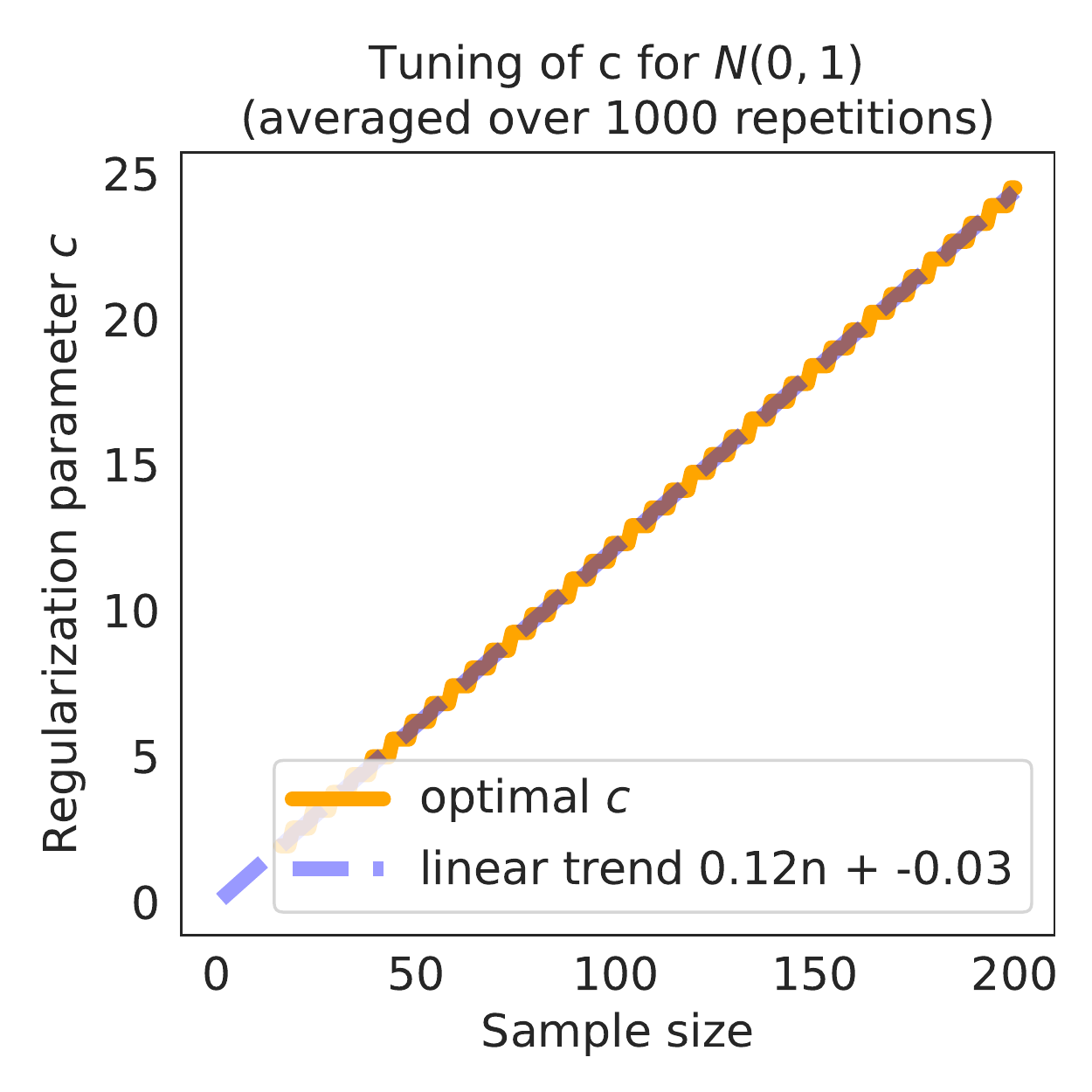}\\
	\includegraphics[width=0.49\linewidth]{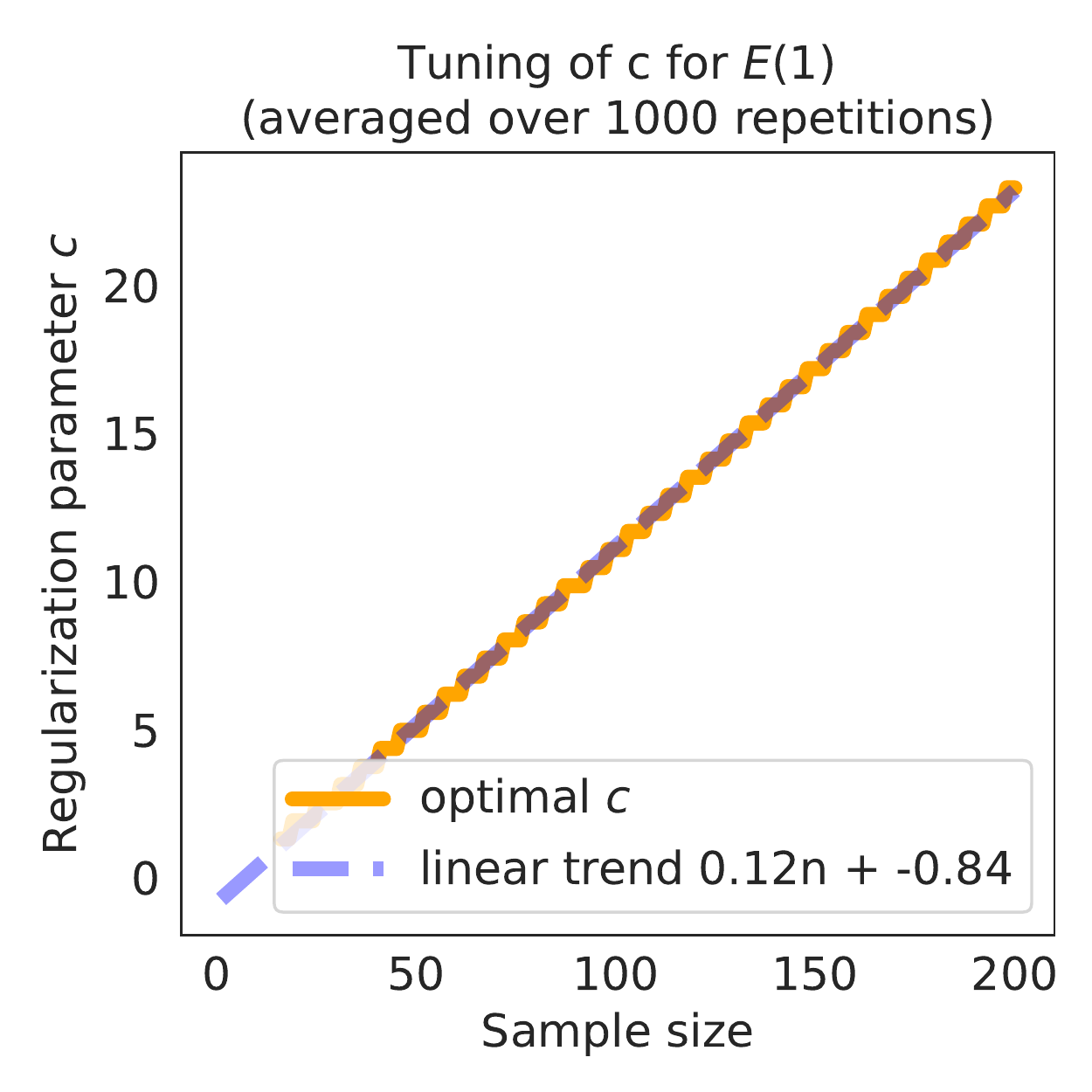}
	\includegraphics[width=0.49\linewidth]{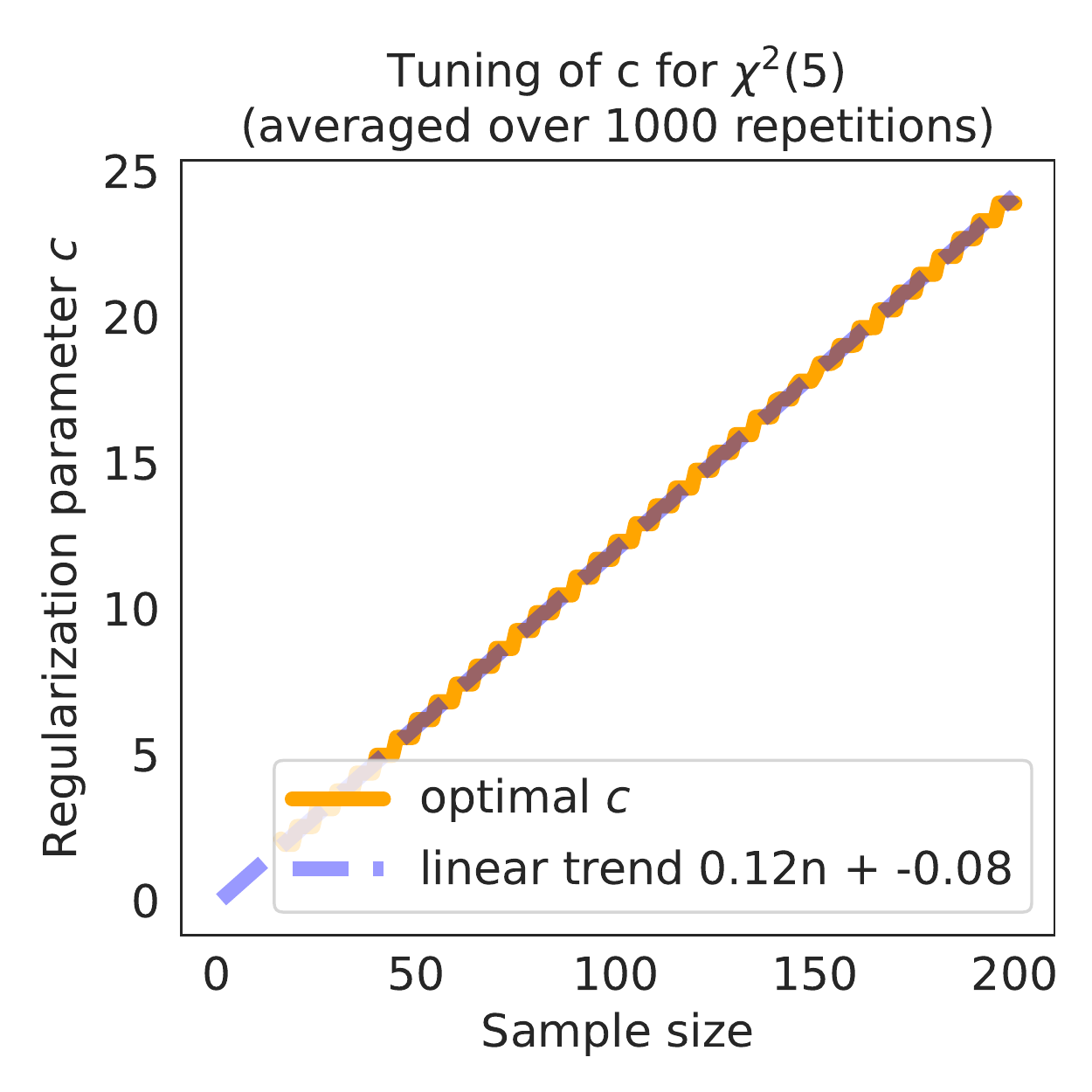}
	\caption{Optimal (smallest confidence interval diameter) local regularization parameter $c^*_n=\argmin_{c} \lvert \Theta_{n, c}\left(\delta\right)\rvert$ for $\delta=5\%$ on Bernoulli (top left), Gaussian (top right), exponential (bottom left) and Chi-square (bottom right). Note the apparent common linear trend $c^*_n\approx 0.12 n$. Results are averaged over $1000$ independent simulations.}
	\label{fig:tuning_c}
\end{figure}

In this section, we provide additional experiments regarding the  local regularization parameter $c$. We study the sensitivity of the bounds to the values of $c$, reported in Figure~\ref{fig:c_sensi-app}. Then, we report in Figure~\ref{fig:tuning_c} plots showing a tuning of parameter $c$ for each fixed $n=n_0$ illustrated for Bernoulli,  Gaussian, exponential and Chi-square distributions. Specifically, $c$ is chosen to minimize the width of the confidence set at $n=n_0$, i.e., $c_{n_0} = \min_{c>0} \lvert \Theta_{n_0, c}\left(\delta\right)\rvert$. Let us recall that such a tuning is only valid for a single value. Hence, while using $c=c_{n_0}$ for a fixed $n_0$ is allowed by the theory to produce time-uniform bounds for all $n$, using $c=c_n$, that is a different value for different $n$, is not supported by the theory as it would break the martingale property necessary for the mixture construction to hold, which is manifested by a contradiction with the law of iterated logarithm (see Section~\ref{sub:main-result}).
 
\section{Application: Generalized Likelihood Ratio Test in Exponential Families}
\label{app:GLR}
In this section, we apply our result to revisit GLR tests in exponential families. 
The Generalized Likelihood Ratio (GLR) in the exponential family model $\cE$ writes a follows
\beqan
G^\cE_{1:s:t} &=& \log \bigg(\frac{\sup_{\theta_1,\theta_2} \prod_{t'=1}^s p_{\theta_1}(X_{t'})\prod_{t'=s+1}^t p_{\theta_2}(X_{t'})  }{\sup_{\theta}  \prod_{t'=1}^t p_\theta(X_{t'})} \bigg)\\
&=&
\sup_{\theta_1,\theta_2} \sum_{t'=1}^s \log p_{\theta_1}(X_{t'}) + \sum_{t'=s+1}^t \log p_{\theta_2}(X_{t'}) 
-\sup_\theta \sum_{t'=1}^t \log p_\theta(X_{t'})\\
&=&s \cL^\star( F_{1:s}) + (t-s)\cL^\star( F_{s+1:t})   - t \cL^\star( F_{1:t})\\
 &=&s  (\cL^\star( F_{1:s}) -\cL^\star( F_{1:t})) + (t-s)(\cL^\star( F_{s+1:t})   -\cL^\star( F_{1:t}))\,,
\eeqan
where $F_{s+1:t}= \frac{1}{t-s}\sum_{t'=s+1}^t F(X_{t'})$, and it makes sense to introduce $\theta_{s+1:t}$ such that $\nabla \cL(\theta_{s+1:t}) = F_{s+1:t}$. 
The GLR test is then defined, for a threshold $\alpha$ as 
\beqan
\tau(\alpha;\cE) &=& \min\{ t \in \Nat : \max_{s\in[0,t)} G^\cE_{1:s:t}\geq \alpha \}
\eeqan

An alternative formulation of the GLR shows that
\begin{align*}
G^\cE_{1:s:t} &= \inf_\theta\sup_{\theta_1,\theta_2} \log \bigg(\frac{ \prod_{t'=1}^s p_{\theta_1}(X_{t'})\prod_{t'=s+1}^t p_{\theta_2}(X_{t'})  }{ \prod_{t'=1}^t p_\theta(X_{t'})} \bigg)\\
&= \inf_\theta\sup_{\theta_1,\theta_2} s \Big(\langle \theta_1, F_{1:s}\rangle - \cL(\theta_1) -\langle \theta, F_{1:t}\rangle + \cL(\theta)\Big) \\
&\quad + (t-s) \Big(\langle \theta_2, F_{s+1:t}\rangle - \cL(\theta_2) -\langle \theta, F_{1:t}\rangle + \cL(\theta)\Big)\,.
\end{align*}
The first supremum is obtained for $\theta_1=[\nabla \cL]^{-1}(F_{1:s})$, with value equal to 
$\cB_\cL(\theta,\theta_1) + \langle \theta-\theta_1,\nabla \cL(\theta_1)\rangle + \langle \theta_1,F_{1:s}\rangle + \langle \theta_1,F_{1_t} \rangle$, that is $\cB_\cL(\theta,\theta_{1:s})
+ \langle\theta, F_{1:s}-F_{1:t}\rangle$.
Likewise, the the second supremum is obtained for 
$\theta_2=[\nabla \cL]^{-1}(F_{s+1:t})$, with value equal to 
$\cB_\cL(\theta,\theta_{s+1:t})
+ \langle\theta, F_{s+1:t}-F_{1:t}\rangle$.
Combining these two results, and remarking that $s (F_{1:s}-F_{1:t}) + (t-s)(F_{s+1:t}-F_{1:t}) =0$, we obtain that
\beqa\label{eqn:GLRinf}
G^\cE_{1:s:t} &=& \inf_\theta  s \cB_\cL(\theta,\theta_{1:s}) + (t-s)\cB_\cL(\theta,\theta_{s+1:t})
\eeqa
We further note that a solution to this optimization problem 
is obtained for $\theta$ such that 
$s(\nabla L(\theta) - \nabla L(\theta_{1:s}))+
(t-s)(\nabla L(\theta) - \nabla L(\theta_{s+1:t}))=0$. Reorganizing the terms, this entails that 
$\nabla L(\theta) = \frac{1}{t}(s F_{1:s}+ (t-s) F_{s+1:t})=F_{1:t}$,
and thus (without surprise) $\theta = \theta_{1:t}$.

Written in the form \eqref{eqn:GLRinf}, the GLR satisfies for any $\theta'$,
\beqan
G^\cE_{1:s:t} &\leq & s \cB_\cL(\theta',\theta_{1:s}) + (t-s)\cB_\cL(\theta',\theta_{s+1:t})\,.
\eeqan
In particular, the false alarm rate of the GLR test, when all observations are generated from parameter $\theta$, can be bounded using
\beqan
\Pr_{\theta}(\tau(\alpha;\cE)<\infty) &=& \Pr_{\theta}\big( \exists (t,s)\in \Nat^2, s<t :  G^\cE_{1:s:t} \geq \alpha\big)\\
&\leq& \Pr_{\theta}\Big( \exists (t,s)\in \Nat^2, s<t :  s \cB_\cL(\theta,\theta_{1:s}) + (t-s)\cB_\cL(\theta,\theta_{s+1:t}) \geq \alpha\Big)\,.
\eeqan
At this point, provided that for appropriate terms $\alpha_1,\alpha_2$,
\begin{align*}
&\Pr_{\theta}\Big( \exists s\in \Nat:  s \cB_\cL(\theta,\theta_{1:s}) \geq \alpha_2\Big)\leq \frac{\delta}{2}\,,\\
&\Pr_{\theta}\Big( \exists (t,s)\in \Nat^2, s<t :  (t-s)\cB_\cL(\theta,\theta_{s+1:t}) \geq \alpha_1\Big)\leq \frac{\delta}{2}\,.
\end{align*}
 we deduce by a simple union bound argument, and for $\alpha=\alpha_1+\alpha_2$ that  $\Pr_{\theta}(\tau(\alpha:\cE)<\infty) \leq \delta$.
 
 \paragraph{A regularized GLR test}
 Our concentration result shows that such a controlled can be obtained for a penalized version of the maximum likelihood parameter estimates $\theta_{1:s},\theta_{s+1,t},\theta_{1,t}$. Namely, let us consider the set 
 \beqan
 \Theta_{s, c}(\delta) &=& \bigg\{ \theta \in \Theta : \exists s\in \Nat:  s \cB_\cL(\theta,\theta_{1:s,c}(\theta)) \geq \alpha_1\bigg\}\,,\\
 \Theta_{s+1:t, c}(\delta) &=& \bigg\{ \theta \in \Theta : \exists (t,s)\in \Nat^2, s<t :  (t-s)\cB_\cL(\theta,\theta_{s+1:t,c}(\theta)) \geq \alpha_2\bigg\}\,.
 \eeqan
Provided that $\alpha_1$ and $\alpha_2$ are defined to ensure that these are high-probability confidence sets for the parameter $\theta_0$, then they should have non-empty intersections under $\bP_{\theta}$. This motivates the following definition of the regularized GLR-test:
\beqan
\tau^{\delta}_{c}(\cE) &=& \min\left\lbrace t \in \Nat : \exists s \!<\! t, \Theta_{s, c}(\delta/2) \cap \Theta_{s+1:t, c}(\delta/2) = \emptyset \right\rbrace\,.
\eeqan
The appropriate quantity for $\alpha_1$ is immediately obtained by Theorem~\ref{thm:genericmom} as $\frac{s}{s+c}(\log(1/\delta) + \gamma_{1:s,c}(\theta))$.
Now for $\alpha_2$, we need to study doubly-time uniform concentration inequalities over both $s$ and $t$.

\subsection{Doubly-time uniform concentration inequalities for exponential families}

\thmthree*

\begin{myproof}{of Theorem~\ref{thm:Doubly-uniform}}
We consider that all the $(X_{t'})_{t'\in\Nat}$ come from a distribution with same parameter $\theta$.
Let $t\in\Nat$ and $s\in\{0,\dots,t-1\}$.
We introduce the scan-mean $\mu_{s+1:t} = \frac{1}{t-s}\sum_{t'=s+1}^{t} F(X_{t'})$ and its mean $\mu = \nabla \cL(\theta)$. We define as in the proof of Theorem~\ref{thm:genericmom},
for each $\lambda$ the martingale 
	\beqan
	M_{s+1:t}^\lambda &=& \exp\bigg(\langle \lambda, (t-s)(\mu_{s+1:t}-\mu)\rangle - (t-s) \cB_{\cL,\theta}(\lambda)\bigg)\,.
	\eeqan
Then, applying the method of mixture and replacing each term with its scan version from $s+1$ to $t$, we introduce the quantity 
\beqan
M_{s+1,t}			&=&
			\exp\big( (t\!-\!s\!+\!c)\cB_{\cL,\theta}^\star(x) \big)\frac{G(\theta,c)}{G(\theta_{s+1:t,c}(\theta_0),t-s+c)}
\eeqan
where  $x = \frac{t\!-\!s}{t\!-\!s\!+\!c}(\mu_{s+1:t} - \nabla \cL(\theta))$ and $\theta_{s+1:t,c}(\theta)$ is the regularized estimate from the scan samples
\beqan
\theta_{s+1:t,c}(\theta) = (\nabla \cL)^{-1}\bigg(\frac{ \sum_{t'=s+1}^t F(X_{t'}) + c \nabla \cL(\theta)}{t\!-\!s\!+\!c}\bigg)\,.
\eeqan

Remarking that $\Esp[M_{t,t}]\leq \Esp[M_{t+1,t}]=1$ and that $(M_{s+1,t})_{s}$ is a nonnegative supermartingale, we can now control its doubly-time uniform deviations following the proof of \citet{maillard19HDR}[Th.3.2, p.58].
More precisely, for any non-decreasing function $g$, it can be shown that
\beqan
\Pr_\theta\big[\exists t, \exists s<t: M_{s+1,t} \geq \frac{g(t)}{\delta}\big] \leq \delta \Esp_\theta\Big[\max_t \max_{s<t} \frac{M_{s+1:t}}{g(t)}\Big] \leq \delta \sum_{t=1}^\infty\frac{1}{g(t)}\,.
\eeqan
Rewriting the terms thanks to the duality formulas, yields
\beqan
\Pr_{\theta}\bigg[\exists t, \exists s<t: (t\!-\!s\!-\!c)\cB_\cL(\theta,\theta_{s+1:t,c}(\theta)) \geq \log\Big(\frac{g(t)}{\delta}\Big) + \gamma_{s+1:t,c}(\theta)\bigg] \leq \delta \sum_{t=1}^\infty\frac{1}{g(t)}\,.
\eeqan

This suggests to set $\alpha_2 = \frac{t\!-\!s}{t\!-\!s\!-\!c}\left(\log\left(\frac{g(t)}{\delta}\right) + \gamma_{s+1:t,c}(\theta)\right)$ in the definition of the regularized GLR test. Putting all terms together, this leads to the definition of the following regularized GLR test, for a given false-detection error probability $\delta\in[0,1]$ and regularization parameter $c$,
\begin{align*}\label{eqn:GLR_reg}
&\Theta_{s, c}(\delta) = \left\lbrace \theta \in\Theta: (s\!+\!c) \cB_\cL(\theta,\theta_{1:s,c}(\theta)) \leq \log(1/\delta)+ \gamma_{1:s,c}(\theta) \right\rbrace\,,\\
&\Theta_{s+1:t, c}(\delta) = \left\lbrace \theta \in\Theta: (t\!-\!s+\!c)\cB_\cL(\theta,\theta_{s+1:t,c}(\theta)) \leq \log(g(t)/\delta) + \gamma_{s+1:t,c}(\theta) \right\rbrace\,,\\
&\tau^{\delta}_{c}(\cE) = \min\left\lbrace t \in \Nat : \exists s \!<\! t, \Theta_{s, c}(\delta/2) \cap \Theta_{s+1:t, c}(\delta/2) = \emptyset \right\rbrace \,.\end{align*}
The factor $g(t)$ that inflates the width of the confidence set can be tuned to satisfy $\sum_{t=1}^{+\infty}1/g(t)\leq 1$, thus controlling the above deviation probability by at most $\delta$ (ideally, $\sum_{t=1}^{+\infty}1/g(t)$ should be close to $1$ to avoid overinflating the confidence width). A natural choice for this is $g(t)=\kappa (1\!+\!t)\log^2(1\!+\!t)$ where $\kappa = 2.10974 \geq \sum_{t=1}^{\infty} \frac{1}{(1\!+\!t)\log^2(1\!+\!t)}$ (for completeness, we derive in the next section an elementary way to computing suitable values for $\kappa$). By construction, this test is guaranteed to have a false-detection probability controlled by $\delta$ (that is, $\Pr_{\theta}(\tau^{\delta}_{c}(\cE)<\infty)\leq \delta$).

\paragraph{Doubly time-uniform control of supermartingale sequences}
For completeness, we  reproduce below the derivation from \citet{maillard19HDR}[Th.3.2, p.58], applied to our setup.
First, let us note that
\beqan
\Pr\bigg( \exists t\!\in\!\Nat_\star,\exists s\!\in\![0,t-1],\,\,  M_{s+1:t}\geq \frac{g(t)}{\delta}
\bigg)
&=&\Pr\bigg( \max_{t\in\Nat_\star}\max_{s\in[0,t-1]}
\frac{M_{s+1:t}}{g(t)}\geq \frac{1}{\delta}\bigg)\\
&\leq &\delta \Esp\bigg[\max_{t\in\Nat_\star}\max_{s\in[0,t-1]}
\frac{M_{s+1:t}}{g(t)}\bigg]\,.
\eeqan	
Let us also denote $\tau$ the random stopping time corresponding to the first occurrence $t$ of the event $\max_{s\in[0,t-1]}
\frac{M_{s+1:t}}{g(t)}\geq \frac{1}{\delta}$.
It is convenient to introduce the quantity
	$\overline{M}_t = \frac{\sum_{s\in\{0,\dots,t-1\}} M_{s+1,t}}{g(t)}$ for each $t\in\Nat_\star$. Since each $M_{s+1,t}$ and $g(t)$ is nonnegative, we first get that for every random stopping time $\tau\in\Nat_\star$, the following inequality holds:
	\beqan
	\Esp\bigg[ \frac{\max_{s<\tau} M_{s+1,\tau}}{g(\tau)}\bigg] &\leq &\Esp\bigg[ \overline{M}_\tau \bigg]= \Esp\bigg[ \overline{M}_1 + \sum_{t=1}^\infty (\overline{M}_{t+1}- \overline{M}_t) \indic{\tau>t}\bigg]\,.
	\eeqan
	Furthermore, we note that, conveniently 
	\beqan
	\overline{M}_{t+1}- \overline{M}_t &=& \frac{M_{t+1,t+1}}{g(t+1)} + \sum_{s=0}^{t-1} \bigg(\frac{M_{s+1,t+1}}{g(t+1)}-\frac{M_{s+1,t}}{g(t)}\bigg)\,.
	\eeqan
	Next, by assumption, we note that $\displaystyle{\Esp[M_{s+1,t+1}|\cF_t] \leq M_{s+1,t}}$.
	Thus, since  $\indic{\tau>t}\in\cF_t$, we deduce that
	\begin{align*}
	&\Esp\bigg[ \frac{\max_{s<\tau} M_{s+1,\tau}}{g(\tau)}\bigg]\\
    &\quad\leq \Esp\big[ \overline{M}_1\big] + \sum_{t=1}^\infty  \frac{\Esp[M_{t+1,t+1}]}{g(t+1)}  +\sum_{t=1}^\infty \sum_{s<t} \Esp\bigg[ \bigg(\frac{1}{g(t+1)}-\frac{1}{g(t)}\bigg)M_{s+1,t}\indic{\tau>t}\bigg]\\
	&\quad= \Esp\big[ \overline{M}_1\big] + \sum_{t=1}^\infty  \frac{\Esp[M_{t+1,t+1}]}{g(t+1)} + \sum_{t=1}^\infty \sum_{s<t}  \bigg(\frac{1}{g(t+1)}-\frac{1}{g(t)}\bigg)\underbrace{\Esp\bigg[M_{s+1,t}\indic{\tau>t}\bigg]}_{\geq0}\,.
	\end{align*}
	Hence, the assumption that $g$ is non-decreasing  ensures that the last sum is upper bounded by $0$. Since on the other hand $\Esp[M_{t+1,t+1}]\leq 1$ holds for all $t$ (and thus $\Esp\big[ \overline{M}_1\big]\leq1/g(1)$), we deduce that the following inequality holds:
	\beqan
	\Esp\bigg[ \frac{\max_{s<\tau} M_{s+1,\tau}}{g(\tau)}\bigg] &\leq& \frac{1}{g(1)}+ \sum_{t=1}^\infty \frac{1}{g(t+1)}  = \sum_{t=1}^\infty \frac{1}{g(t)} \leq 1\,.
	\eeqan
\end{myproof}

\subsection{Computing the inflating factor $g(t)$}

The right-hand side of the doubly time-uniform confidence set of Theorem~\ref{thm:Doubly-uniform} involves a $\log g(t)/\delta$ term instead of $\log 1/\delta$ as in Theorem~\ref{thm:genericmom}, which is a by-product of the union-like argument used in the proof. Ideally, $g(t)$ should grow as slow as possible with $t$ to avoid unnecessary looseness in the confidence bound. That is however limited by the constraint $\sum_{t=1}^{\infty}1/g(t) \leq 1$, which prohibits the use of a linearly growing $g(t)$ (since $\sum_{t=1}^{\infty} 1 / t$ diverges). The choice $g(t)=\kappa (1\!+\!t)\log^{1\!+\!\eta}(1\!+\!t)$ for some $\eta>0$ and $\kappa>0$ guarantees that the series of inverses converges, although its limit is not available in closed-form. However, it is sufficient to set $\kappa$ to an upper bound on $\sum_{t=1}^{\infty} \frac{1}{(1\!+\!t)\log^{1\!+\!\eta}(1\!+\!t)}$. The following elementary lemma explains how to compute a tight value for $\kappa$.

\begin{lemma}\label{lem:kappa} Let $f\colon t\in\bN \mapsto (1\!+\!t)\log^{1\!+\!\eta}(1\!+\!t)$ and define $S_p = \sum_{t=1}^{p} 1/f(t)$ for $p\in\bN\cup\lbrace +\infty\rbrace$. Then $S_{\infty} \leq S_p + \frac{1}{\eta\log(1+p)^\eta}$.
\end{lemma}

\begin{myproof}{of Lemma~\ref{lem:kappa}}
The function $f$ is nondecreasing and positive, therefore for $t\in\bN$ and $x\in[t, t+1)$, it holds that $1/f(t+1)\leq 1/f(x)$. Integrating both terms as functions of $x$ over the interval $[t, t+1)$ yields $1/f(t+1)\leq \int_{t}^{t+1}1/f(x) dx$. Summing starting at $t=p$, we further get the following sum-integral comparison:
\[S_{\infty} - S_{p} \leq \int_p^{\infty} \frac{dx}{f(x)} \,.\]
Finally, note that $\frac{d}{dx} \frac{1}{\eta\log^\eta (1+x)} = -\frac{1}{f(x)}$, and thus $\int_p^{\infty} \frac{dx}{f(x)} = \frac{1}{\eta\log(1+p)^\eta}$.
\end{myproof}

This lemma shows that $\kappa=S_p + 1/\log(1\!+\!p)$ for some $p\geq 1$ ($\eta=1$) is a valid choice for the definition of the inflating factor $g(t)$, and is straightforward to compute numerically. For $p=100$, we obtain $\kappa\lesssim 2.10974$ and increasing $p$ only changes further digits. 

\subsection{Experiments on change point detection for Gaussian with unknown variance}\label{app:xp_changepoint}

We illustrate the change point GLR test above in a numerical experiment as follows. We consider the one-dimensional exponential family of centered Gaussian distributions with unknown variance $\cE=\left\lbrace \cN(0, \sigma^2), \sigma>0\right\rbrace$. A sequence of independent random variables $(X_t)_{t\in\bN}$ is drawn from $X_t\sim\cN(0, \sigma_0^2)$ if $t\leq t^*$ and from $X_t\sim\cN(0, \sigma_1^2)$ if $t>t^*$, where $t^*=50$ and $\sigma_0=1$ and $\sigma_1\in[1, 2, 3, 4]$. Of note, this setting corresponds to an open question in \citet{Maillard2018GLR}, which studies GLR tests in sub-Gaussian families, which are adapted to the detection of changing means but not variances. Detection times are reported as $\min(\tau^{\delta}_{c}(\cE), T)$ with $T=100$, and $\tau^{\delta}_{c}(\cE)>T$ is interpreted as no change being detected.  For the doubly time-uniform confidence set in the definition of the regularized GLR test, we use the factor $g(t)= \kappa (1\!+\!t) \log^2(1\!+\!t)$ with $\kappa=2.10974$. 

A practical motivation for this setting is, for instance, the design of maintenance models for equipment: with time, a component of a physical system (mechanical component, measuring instrument...) starts to wear off and while it is still functioning (same output on average), it is more imprecise or less stable (higher output variance); the goal of the maintenance agent is to detect as early as possible such changes of regime in order to replace the failing component before it breaks completely. 

\begin{figure}[H]
    \centering
    \begin{subfigure}[t]{.49\linewidth}
		\includegraphics[width=1\linewidth,height=1\linewidth]{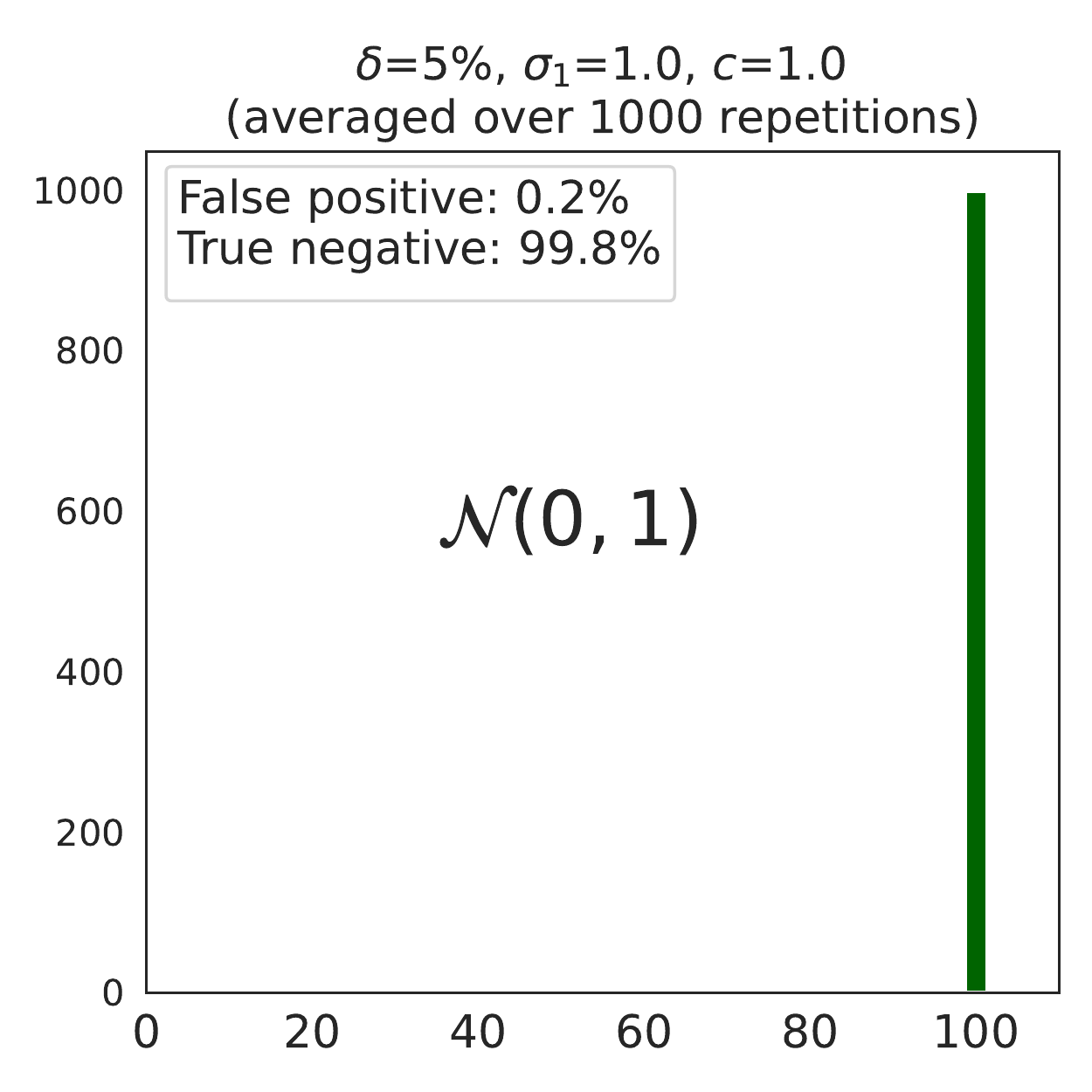}
		\vskip -2mm
		\caption{\label{fig:changepoint_no_change}$\sigma_0=\sigma_1=1$ (no change)}
	\end{subfigure}
	\begin{subfigure}[t]{.49\linewidth}
		\includegraphics[width=1\linewidth,height=1\linewidth]{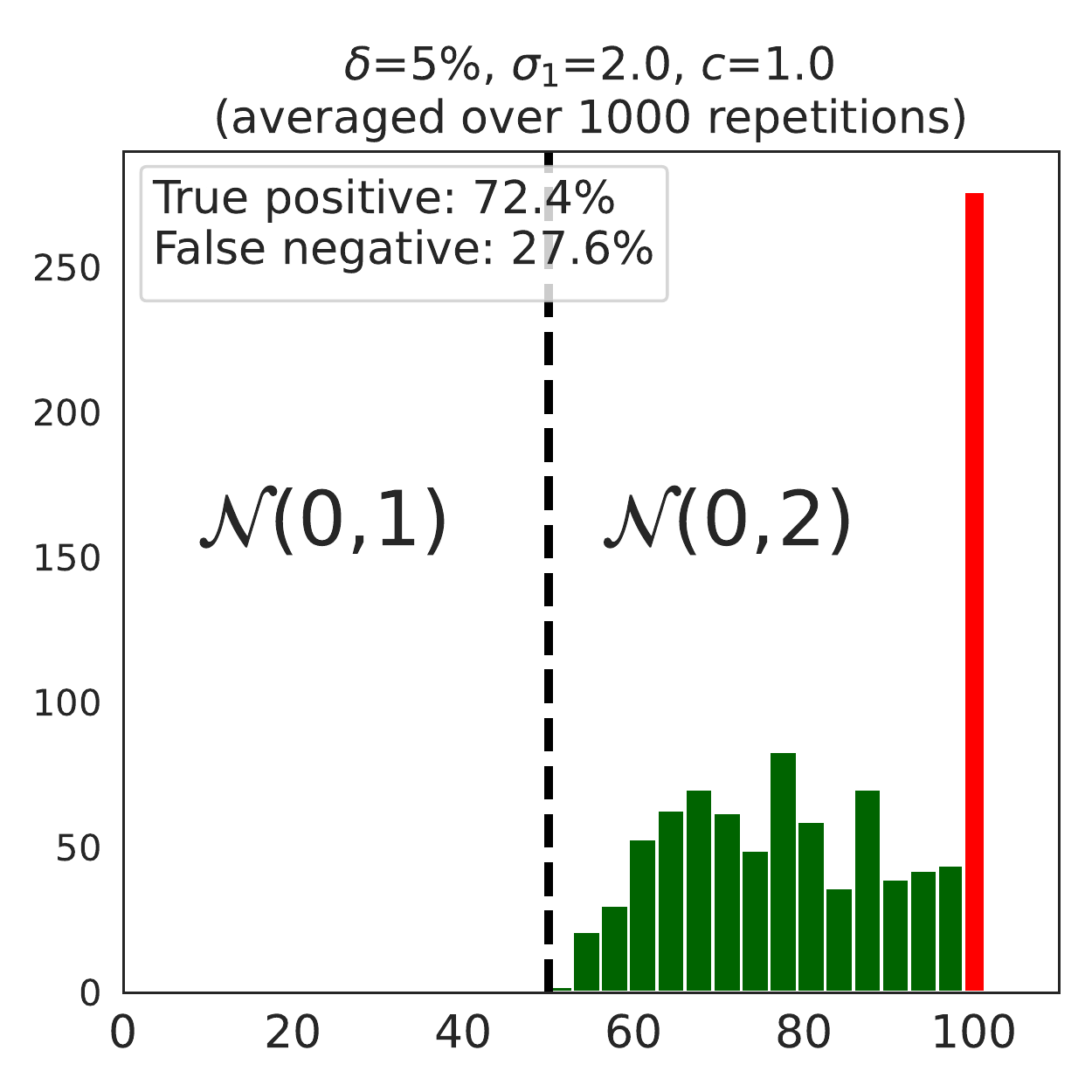}
		\caption{\label{fig:changepoint_small_change}$\sigma_0=1, \sigma_1=2$}
	\end{subfigure}\\
    \begin{subfigure}[t]{.49\linewidth}
		\includegraphics[width=1\linewidth,height=1\linewidth]{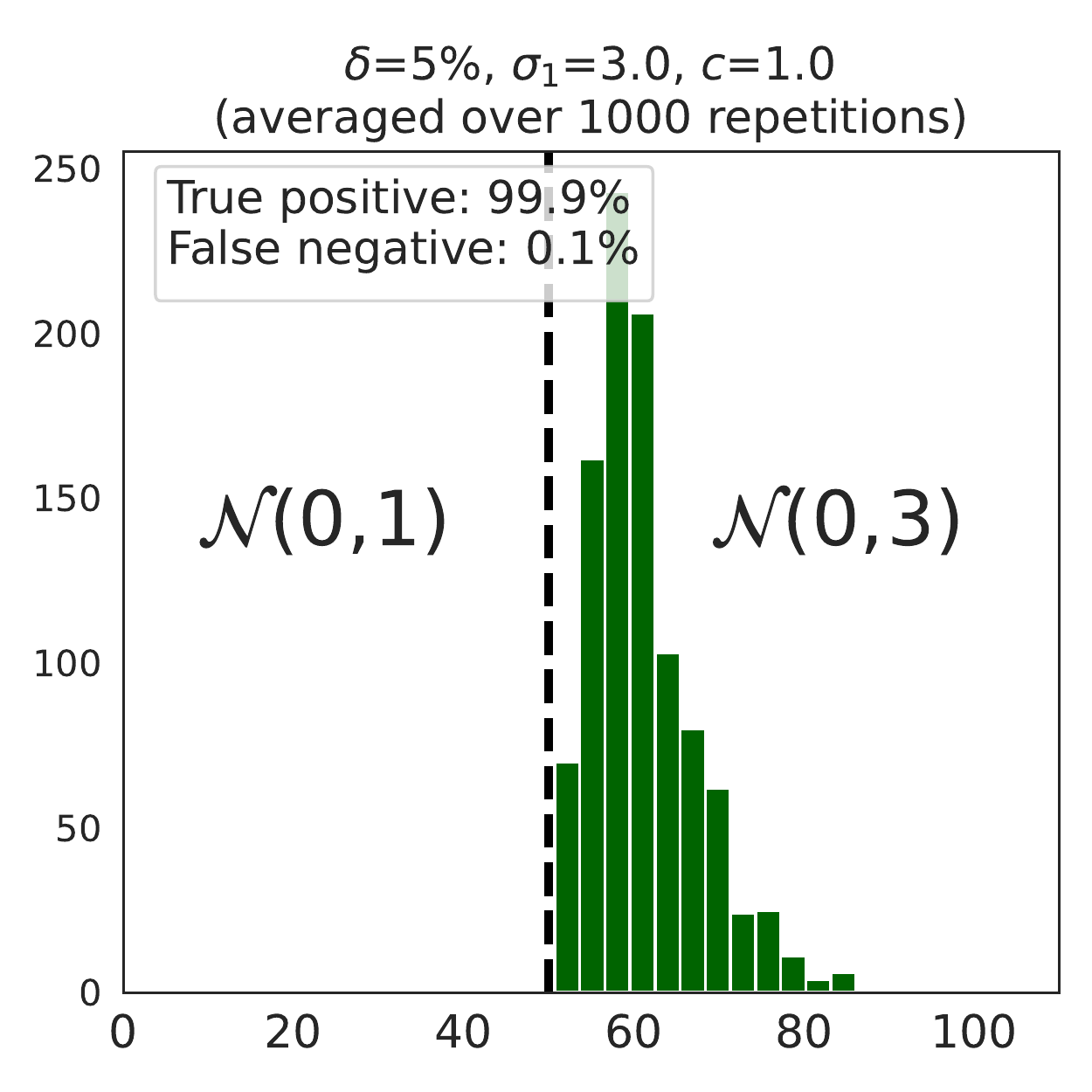}
		\vskip -2mm
		\caption{$\sigma_0=\sigma_1=3$}
	\end{subfigure}
	\begin{subfigure}[t]{.49\linewidth}
		\includegraphics[width=1\linewidth,height=1\linewidth]{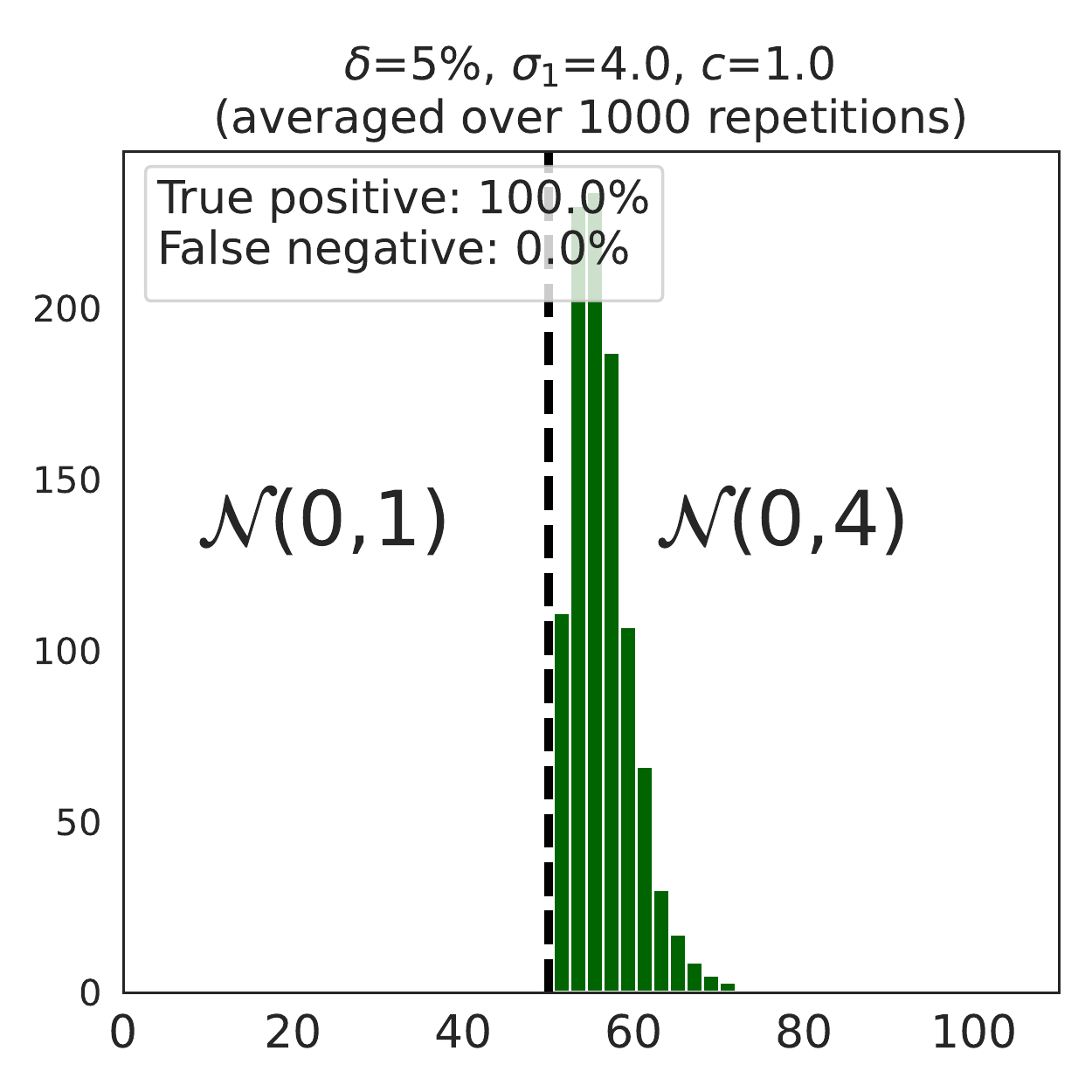}
		\caption{$\sigma_0=1, \sigma_1=4$}
	\end{subfigure}
    \caption{Histograms of detection times $\tau^{\delta}_{c}(\cE)$ in the Gaussian setting described in Appendix~\ref{app:xp_changepoint}. Green: correct detection (negative in (a), positive in (b), (c) and (d)). Red: incorrect detection (positive in (a), negative in (b), (c) and (d)). Black dashed line: change point $t^*$.}
	\label{fig:changepoint}
\end{figure}

\begin{figure}[H]
    \centering
	\includegraphics[width=.6\linewidth]{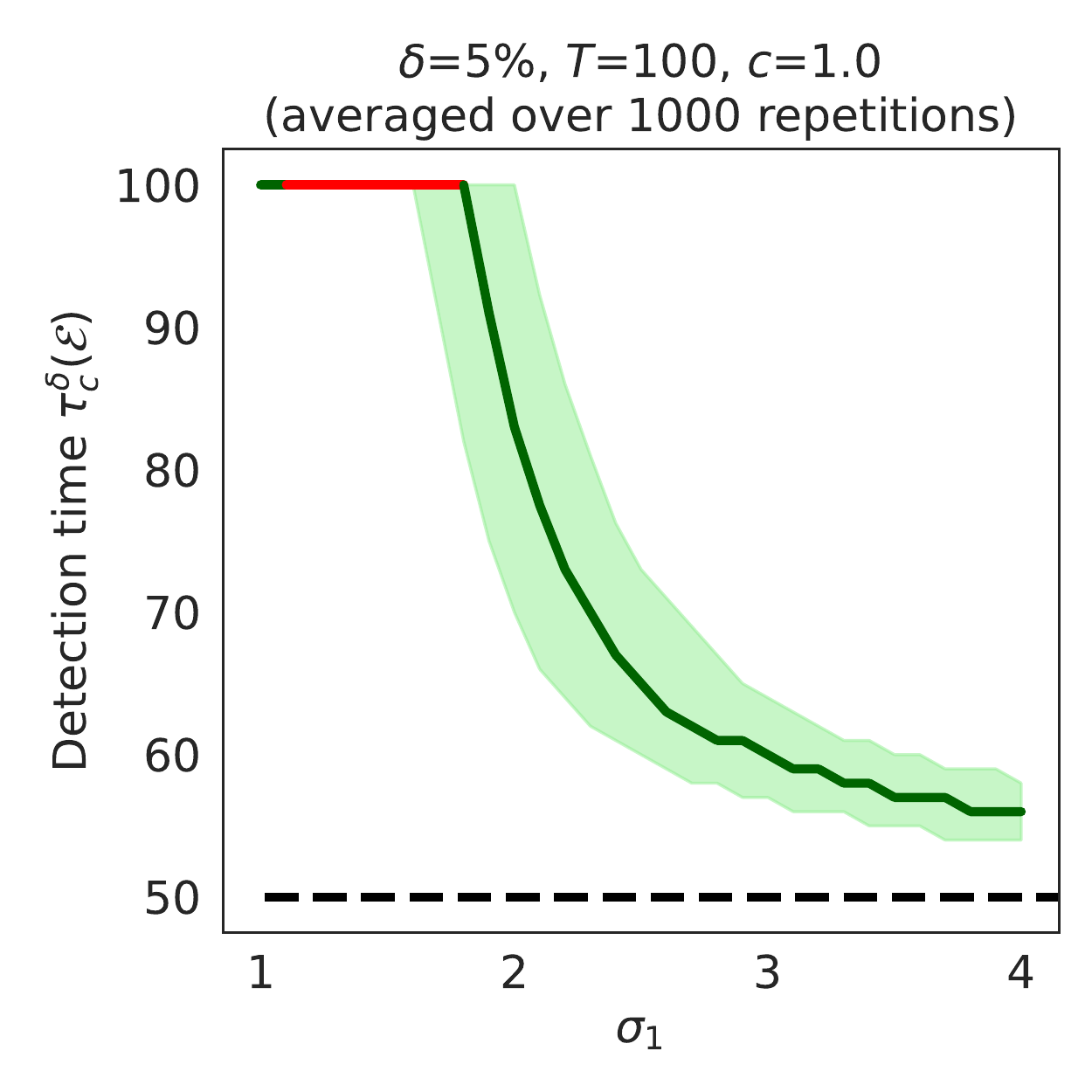}
    \caption{Detection time $\tau^{\delta}_{c}(\cE)$ in the Gaussian setting described in Appendix~\ref{app:xp_changepoint}. Solid line: median detection time. Shaded area: interquartile range of detection times (25th and 75th percentiles). Black dashed line: change point $t^*$.}
	\label{fig:changepoint_sigmas}
\end{figure}

In Figure~\ref{fig:changepoint}, we report the histograms of detection times across $1000$ independent simulations for increasingly abrupt changes of variance ($\sigma_1\in\lbrace 1, 2, 3, 4\rbrace$). As expected, the distribution of detection times shifts closer to the actual change point $t^*$ when $\sigma_1$ increases, i.e., more obvious changes are detected earlier. This is confirmed in Figure~\ref{fig:changepoint_sigmas}, where we report the median and interquartile range of detection times for $\sigma_1\in[1, 4]$.

We empirically validate in Figure~\ref{fig:changepoint_no_change} the control of the false positive probability by at most $\delta=5\%$. However, the actual false positive rate appears much lower ($0.2\%$), which is a consequence of the looseness of the union bound involving the factor $g(t)$ in the doubly time-uniform confidence set. Following \citet{Maillard2018GLR}, a sharper approach would involve a direct concentration result on the pair $\left(\theta_{s,c},\theta_{s+1:t, c}\right)$ (i.e., in the terminology of \cite{Maillard2018GLR}, a joint bound, as opposed to the current disjoint one). This is a nontrivial result which we leave for future work. Finally, note that the high false negative rate of Figure~\ref{fig:changepoint_small_change} (27.6\%) is an artifact of thresholding the detection time at $T=100$; increasing $T$ would enable later detections, thus reducing the number of observed false negatives, at the cost of increasing the average detection delay (as well as the computational burden of the experiment).

\section{Application: Linear Contextual Bandits}\label{app:bandits}

In this section, we apply Theorem~\ref{thm:momEFsequence} to build confidence sets in the well-known linear bandit setting.
We consider the setting of linear contextual bandits \citep{Auer02,abbasi2011improved}, but with possibly arm-dependent noise variance (heteroscedastic noise). An algorithm for this problem chooses, at each round $t \in \Nat$, an action or arm $A_t\!\in\! \cA$, and subsequently observes a reward $X_t=\theta^\top \phi(A_t) + \epsilon_t$, where $\phi\!:\cA \to \Real^d$ is a (fixed) map from actions to their context vectors, $\theta \in \Real^d$ is a vector of weights unknown to the algorithm and given the action $A_t$, $\epsilon_t \sim \cN(0,\sigma_{A_t}^2 I)$ is a Gaussian noise with known variance $\sigma^2_{A_t}$. The goal is to select actions to maximize the expected cumulative reward $\sum_t \theta^\top \phi(A_t)$. Any rational agent would choose the action $A_t$ causally depending upon the history of arms and reward sequences available before the start of round $t$. Naturally, for each $n \!\in\! \Nat$, it boils down to controlling the deviation between the unknown parameter $\theta$ and a suitable estimate $\theta_n$ built from $n$ observations $(A_t,X_t)_{t \leq n}$. 

Under this observation model, the feature function and the log-partition function are given by $F_t(x)=\frac{x}{\sigma_{A_t}^2}\phi(A_t)$ and $\cL_t(\theta)=\frac{1}{2\sigma_{A_t}^2}\norm{\theta}^2_{\phi(A_t)\phi(A_t)^\top}$. Let us define the matrix 
\begin{align*}
V_n=\sum_{t=1}^{n}\frac{1}{\sigma_{A_t}^2}\phi(A_t)\phi(A_t)^\top\,,
\end{align*}
and introduce the Legendre function $\cL_0(\theta)=\frac{1}{2}\norm{\theta}^2_{V_0}$, where $V_0$ is some fixed positive definite matrix. Then, the parameter estimate and the Bregman information gain take the form
\begin{align*}
\theta_{n,\cL_0} = \left(V_n + V_0\right)^{-1}\sum_{t=1}^{n}\frac{X_t}{\sigma_{A_t}^2}\phi(A_t)~,\quad
\gamma_{n,\cL_0} = \log\frac{\det(V_0+V_n)^{1/2}}{\det(V_0)^{1/2}}~.
\end{align*}

Now, defining $\cH_{t-1}:=\lbrace A_1,X_1,\ldots,A_{t-1},X_{t-1},A_t\rbrace$ to be the set of all information available before observing $X_t$, one can see that the confidence set $\Theta_{n,\cL_0}(\delta)$ given by Theorem~\ref{thm:momEFsequence} is the set of all $\theta \in \Theta$ satisfying
\begin{align*}
  \norm{\theta-\theta_{n,\cL_0}}^2_{V_n+V_0}
  \leq \norm{\theta}^2_{V_0} + 2\log\frac{\det(V_0+V_n)^{1/2}}{\delta\det(V_0)^{1/2}}.
\end{align*}


\paragraph{Comparison with Existing Results}
Under the assumption that the noise $\epsilon_t$ is $\sigma_{A_t}$-sub-Gaussian conditioned on $\cH_{t-1}$, \citet{abbasi2011improved,kirschner2018information} obtain a high-probability confidence set $\widetilde\Theta_{n,\cL_0}(\delta)$, which is the set of all $\theta \in \Theta$ satisfying
\beqan
  \norm{\theta - \theta_{n, \cL_0}}_{V_n + V_0}
  \leq \norm{\theta}_{V_0} + \sqrt{2\log\frac{\det(V_0+V_n)^{1/2}}{\delta\det(V_0)^{1/2}}}.  
  \eeqan
  Interestingly, since $\sqrt{a + b} \leq \sqrt{a}+\sqrt{b}$ for positive $a,b$, it is then clear that $\Theta_{n,\cL_0}(\delta) \subset \widetilde\Theta_{n,\cL_0}(\delta)$. In particular, as a side result of our approach, we obtain a tighter confidence set compared to them.

\end{appendix}

\end{document}